%% file: iclr2026_conference.tex
\documentclass{article} % For LaTeX2e
\usepackage{iclr2026_conference,times}

\input{math_commands.tex}

\input{sections/preamble/authors}

\usepackage[utf8]{inputenc} % allow utf-8 input
\usepackage[T1]{fontenc}    % use 8-bit T1 fonts
\usepackage{hyperref}       % hyperlinks
\usepackage{url}            % simple URL typesetting
\usepackage{booktabs}       % professional-quality tables
\usepackage{amsfonts}       % blackboard math symbols
\usepackage{nicefrac}       % compact symbols for 1/2, etc.
\usepackage{microtype}      % microtypography
\usepackage[table]{xcolor}  % colors
\usepackage{subcaption}     % in preamble
\usepackage{graphicx}
\usepackage{float}

\usepackage{listings}
\usepackage{xkcdcolors}
\usepackage[dvipsnames]{xcolor}
\usepackage[most]{tcolorbox}
\usepackage{wrapfig}
\usepackage[capitalise]{cleveref}
\usepackage{amsmath}
\usepackage{amssymb}
\usepackage{tikz}
\usepackage{caption}
\usepackage{adjustbox}
\usepackage{enumitem}
\usepackage{xparse}
\usepackage{multirow}
\hypersetup{
    colorlinks=true,
    linkcolor=xkcdMerlot,
    urlcolor=xkcdBluish,
    linktoc=all,
    citecolor=xkcdPrussianBlue,
}% hyperlinks

\crefname{figure}{Figure}{Figures}
\Crefname{figure}{Figure}{Figures}

\usetikzlibrary{arrows.meta, positioning, shapes.geometric, calc, matrix}
\usepackage{mathtools}

\newcommand{\highlightnodetikz}[2]{%
  \raisebox{0pt}[1.5em][0pt]{%
    $\underbracket[0.5pt][1pt]{\vphantom{pg}\smash{\texttt{#1}}}_{\textcolor{xkcdPinkyRed}{\textstyle #2}}$%
  }%
}

\usepackage{algorithm}
\usepackage{algpseudocode}

\newcommand{\pointer}[1]{\textcolor{xkcdTeal}{\texttt{<P#1>}}}
\newcommand{\variable}[1]{\textcolor{xkcdLightOrange}{\texttt{#1}}}
\newcommand{\role}[1]{\textcolor{xkcdPrussianBlue}{\texttt{#1}}}
\newcommand{\concept}[1]{\texttt{#1}}
\newcommand{\stopwordamr}[1]{\textcolor{xkcdDarkOrange}{\texttt{#1}}}

\NewDocumentCommand{\drawcolorvector}{O{0}O{0}O{1}m}{%
  \foreach [count=\i from 0] \col in {#4} {
    \pgfmathsetmacro{\x}{#1 + \i * #3}
    \pgfmathsetmacro{\s}{#3}
    \fill[\col] (\x,#2) rectangle ++(\s,\s);
    \draw (\x,#2) rectangle ++(\s,\s);
  }
}
\NewDocumentCommand{\pcref}{m o}{%
  \IfNoValueTF{#2}
    {(%
      \cref{#1}%
    )}%
    {(%
      #2~\cref{#1}%
    )}%
}

\newcommand{\modelname}[0]{SAFT} % {S}tructure-{a}ware {f}ine-{t}uning

\newcommand{\rebuttal}[1]{\textcolor{black}{#1}}

\definecolor{tumblue}{RGB}{0,101,189}
\definecolor{darkgreen}{RGB}{0, 140, 64}  % readable, darker than default

\title{\modelname{}: Structure-Aware Fine-Tuning of LLMs for AMR-to-Text Generation}
\iclrfinalcopy % Uncomment for camera-ready version, but NOT for submission.
\begin{document}
\maketitle

\input{sections/0-abstract}
\input{sections/1-introduction}
\input{sections/2-background}
\input{sections/3-method}
\input{sections/4-experiments}
\input{sections/5-related-works}
\input{sections/6-conclusion}
\input{sections/99-reproducibility-statement}

\bibliographystyle{unsrtnat}
\bibliography{references}
%%%%%%%%%%%%%%%%%%%%%%%%%%%%%%%%%%%%%%%%%%%%%%%%%%%%%%%%%%%%
\appendix
\section*{Appendix}
\input{sections/appendix/A-amr}
\input{sections/appendix/B-implementation_details}
\input{sections/appendix/C-additional_experiments}
\input{sections/appendix/D-limitations}

\input{sections/appendix/E-assets_and_licences}
\input{sections/appendix/F-related_work}

\input{sections/appendix/G-examples}

\end{document}

%% file: math_commands.tex
\usepackage{amsmath,amsfonts,bm}

\def\eqref#1{equation~\ref{#1}}
\def\1{\bm{1}}

\def\vt{{\bm{t}}}

\def\mA{{\bm{A}}}

\def\mD{{\bm{D}}}

\def\mH{{\bm{H}}}
\def\mI{{\bm{I}}}

\def\mL{{\bm{L}}}

\def\mU{{\bm{U}}}

\def\mX{{\bm{X}}}

\def\mLambda{{\bm{\Lambda}}}

\DeclareMathAlphabet{\mathsfit}{\encodingdefault}{\sfdefault}{m}{sl}
\SetMathAlphabet{\mathsfit}{bold}{\encodingdefault}{\sfdefault}{bx}{n}

\def\gA{{\mathcal{A}}}

\def\gD{{\mathcal{D}}}
\def\gE{{\mathcal{E}}}

\def\gG{{\mathcal{G}}}

\def\gL{{\mathcal{L}}}

\def\gR{{\mathcal{R}}}
\def\gS{{\mathcal{S}}}
\def\gT{{\mathcal{T}}}
\def\gU{{\mathcal{U}}}
\def\gV{{\mathcal{V}}}

\newcommand{\R}{\mathbb{R}}
\newcommand{\C}{\mathbb{C}}

%% file: sections/preamble/authors.tex
\author{%
  Rafiq Kamel\(^{1,}\)\thanks{Equal contribution.} \\
  \And
  Filippo Guerranti\(^{1, 2, 3, \ast}\)\\
  \And
  Simon Geisler\(^{1, 2,}\)\thanks{Now at Google.}\\
  \And
  Stephan G{\"u}nnemann\(^{1, 2, 3}\)\\
  \AND
  \vspace{-2em}\\
  {\small
    \(^1\)School of Computation, Information and Technology, Technical University of Munich, Germany
  } \\
  {\small 
    \(^2\)Munich Data Science Institute (MDSI), Germany} \\
  {\small 
    \(^3\)Munich Center for Machine Learning (MCML), Germany} \\
  {\small 
    \texttt{\{r.kamel,f.guerranti,s.geisler,s.guennemann\}@tum.de}
  }\\
}

%% file: sections/0-abstract.tex
\begin{abstract}
Large Language Models (LLMs) are increasingly applied to tasks involving structured inputs such as Abstract Meaning Representations (AMRs). However, common approaches either linearize graphs, discarding crucial structural cues, or rely on specialized architectures that are incompatible with standard pretrained LLMs. We present \modelname{}, a structure-aware fine-tuning method that augments LLMs with graph-sensitive positional encodings derived from the magnetic Laplacian of AMRs. These encodings are projected into the LLM embedding space, introducing relational inductive bias without modifying the model architecture. Designed to be applicable across tasks involving graph-structured inputs, we demonstrate its effectiveness on AMR-to-text generation, where \modelname{} establishes a new state of the art on AMR 3.0 with a +3.5 BLEU improvement over prior baselines. Performance gains grow with graph complexity, highlighting the value of structure-aware representations in enhancing LLM performance.\footnote{Accepted at the KDD2025 Workshop on Structured Knowledge for LLMs.}

% OLD VERSION (NeurIPS)
% Large Language Models (LLMs) are increasingly applied to tasks involving structured inputs such as graphs. Abstract Meaning Representations (AMRs), which encode rich semantics as directed graphs, offer a rigorous testbed for evaluating LLMs on text generation from such structures. Yet, current methods often arbitrarily linearize AMRs, discarding key structural cues, or rely on architectures incompatible with standard LLMs. We introduce \modelname{}, a structure-aware fine-tuning approach that injects graph topology into pretrained LLMs without architectural changes. We compute direction-sensitive positional encodings from the magnetic Laplacian of transformed AMRs and project them into the embedding space of the LLM. While possibly applicable to any graph-structured inputs, we focus on AMR-to-text generation as a representative and challenging benchmark. \modelname{} sets a new state-of-the-art on AMR 3.0 with a 3.5 BLEU improvement over baselines. Gains scale with graph complexity, highlighting the value of structure-aware representations in enhancing LLM performance. \modelname{} offers a general and effective pathway for bridging structured data and language models.

\end{abstract}

%% file: sections/1-introduction.tex
\section{Introduction}

Large Language Models (LLMs) have become the dominant paradigm for natural language processing, demonstrating strong generalization across a wide range of sequential tasks. Increasingly, researchers are exploring how to extend the capabilities of LLMs to structured data domains such as graphs \citep{jin2024llmgraphsurvey, jiang2023structgpt, fatemi2024talk, zhang2022greaselm, tang2024graphgpt}, driven by  growing interest in extending the reasoning and representation capabilities of LLMs beyond sequential data to more expressive, structured modalities. However, existing approaches that adapt LLMs to graphs often require architectural modifications, or auxiliary components. These strategies compromise the scalability and flexibility of LLMs as pretrained, general-purpose sequence models.

A particularly well-defined and linguistically grounded graph representation is the Abstract Meaning Representation (AMR) \citep{banarescu2013amr}: a rooted, directed acyclic graph that encodes predicate-argument structure and core semantic relations. We focus on the AMR-to-text generation task: producing a natural language sentence that accurately expresses the meaning of an AMR graph. This task represents a strong benchmark for evaluating the ability of LLMs to interface with structured semantic representations, as it demands sensitivity to graph topology and semantic content while preserving fluency and coherence in the generated output.

Despite its importance, AMR-to-text generation remains challenging due to the inherent relational and semantic structure of AMRs. Sequence-to-sequence models \citep{bevilacqua2021spring, cheng2022bibl} linearize AMRs, discarding structural information crucial for semantic fidelity. Graph-to-sequence methods \citep{song2018graph2seq, zhu2019modeling, ribeiro2021structadapt} preserve structure through Graph Neural Networks (GNNs), but their reliance on specialized encoders breaks compatibility with pretrained LLMs. More recent work attempts to repurpose LLMs for this task via prompting or fine-tuning on linearized AMRs \citep{mager2020gpt, yao2024semantic, raut2025can}, but these methods still overlook the underlying graph structure, critical to meaning preservation.
This fragmentation reveals a critical gap:  
\begin{quote}
\centering
\textit{How can graph-structured information be integrated into LLMs in a lightweight, architecture-agnostic way to enable structure-aware generation?}
\end{quote}

\input{figures/pipeline}

We address this question with \textbf{\modelname{}}, a structure-aware fine-tuning method that augments LLM inputs with positional encodings derived from graph topology. Specifically, we compute direction-sensitive graph positional encodings from the magnetic Laplacian \citep{furutani2020magnetic, geisler2023transformers} of an AMR-derived graph and inject them into the embeddings of the graph linearization tokens via a lightweight projection network. This design ensures compatibility with any decoder-only LLM and avoids architectural changes to the model, as illustrated in \cref{fig:pipeline}.

% While our approach is grounded in AMR-to-text generation, its design is conceptually applicable to other tasks involving graph-structured inputs, such as drug design \citep{zheng2024llmsdrug}, code representation learning \citep{allamanis2017learning}, and scene graph-to-text generation \citep{yang2019auto}. 
\rebuttal{We propose a conceptually general approach for injecting structural inductive bias into LLMs via graph positional encodings. In this work, we specifically evaluate our idea in the context of AMR-to-text generation} as it provides a linguistically motivated, semantically rich benchmark that allows for precise evaluation of structural understanding in language generation. 
% Their formalism enables controlled experimentation with topology and meaning, making them an ideal foundation for this line of work. 
\modelname{} provides a concrete step toward aligning graph structures with pretrained LLMs, focusing on AMRs as a high-value benchmark for studying structure-aware generation, \rebuttal{while we recognize that the field of structure‑to‑text tasks extends beyond AMRs}.
Our contributions include:
\begin{itemize}[leftmargin=*, itemsep=2pt, topsep=2pt]
\item A \textbf{structure-aware fine-tuning framework for LLMs} that incorporates graph positional encodings into token embeddings, enabling relational inductive bias without modifying the model architecture.
\item A novel formulation of \textbf{AMR-specific positional encodings} derived from the eigenvectors of the magnetic Laplacian, effectively capturing \textit{directionality} and \textit{structural information} in AMRs.
\item Comprehensive experiments showing that \modelname{} achieves \textbf{state-of-the-art performance} on AMR 3.0, with a +3.5 BLEU improvement over baselines, and increased gains on graphs with higher structural complexity, such as document-level AMRs.
\end{itemize}

%% file: figures/pipeline.tex
\begin{figure*}[!t]
    \centering
    \includegraphics[width=0.95\textwidth]{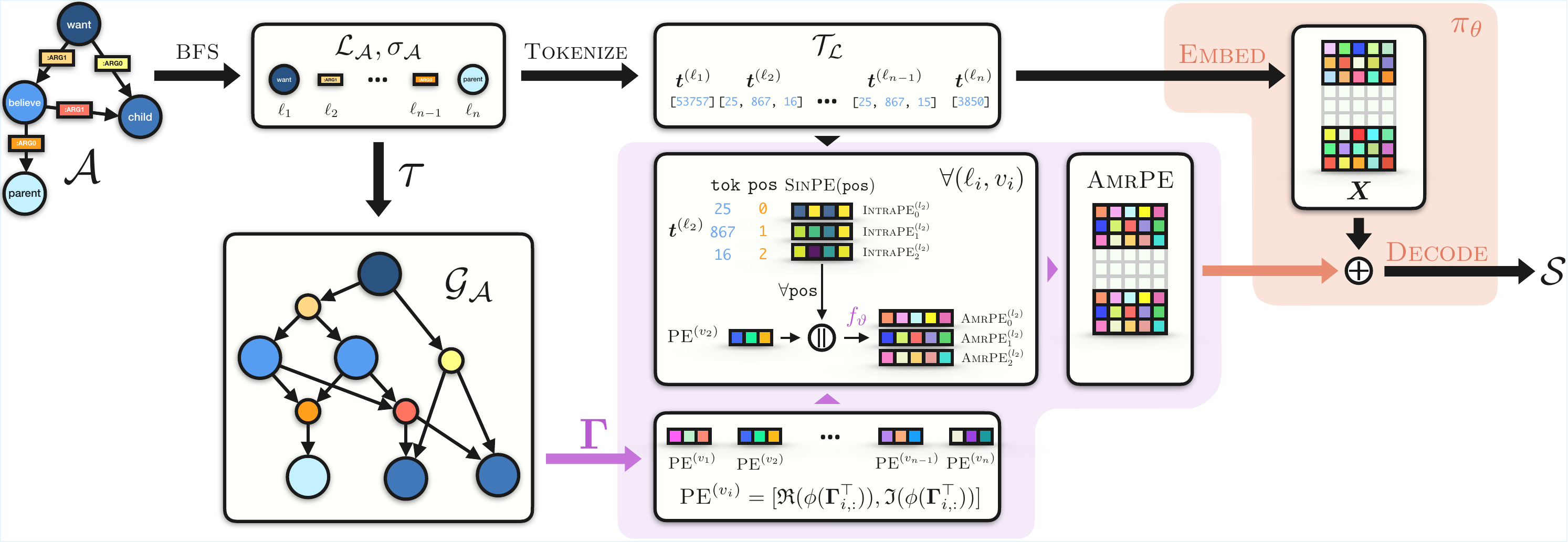}
    \caption{\textbf{Overview of SAFT.} An AMR graph $\gA$ is first linearized into a token sequence $\gL_\gA$. We then construct a graph transformation $\gG_\gA$ and compute structure-aware positional encodings from its magnetic Laplacian. These encodings are combined with standard token positions to form AMR-specific embeddings (\textsc{AmrPE}). A simple MLP \( f_\vartheta \) aligns them with the embedding space of the LLM \(\pi_\theta\), after which they are injected into the token embeddings $X$. The model is fine-tuned to generate text $\gS$, enabling structure-aware AMR-to-text generation without altering the LLM architecture.}
    \label{fig:pipeline}
    % \caption{
    %     \textbf{Overview of SAFT.} Given the AMR linearization \( \gL_{\gA} \), we construct its transformation \( \gG_\gA=\tau(\gA) \), compute graph PEs from the magnetic Laplacian \( \smash{\mL^{(q)}} \) of \( \gG_\gA \), and combine them with intra-node token PEs. The result is projected via \( f_\vartheta \) and injected into the token embeddings \( \mX \) of the tokenized linearization, enabling structure-aware generation without modifying the LLM architecture.
    % }
        % \textbf{Overview of SAFT.} We linearize an AMR graph into \(\gL_\gA\), transform it into \(\gG_\gA\), and compute intra-node tokens \(\gT_\gL\) (\cref{sec:method.amr_transformation}). We then compute graph PEs using the magnetic Laplacian \(\smash{\mL^{(q)}}\) of \(\gG_\gA\), and intra-node PEs from \(\gT_\gL\) (\cref{sec:method.amr_pe}), and then integrated the projection (\(f_\theta\)) of their concatenation in the LLM embedding space during fine-tuning (\Cref{sec:method.fine_tuning}).
\end{figure*}

%% file: sections/2-background.tex
\section{Background}
\label{sec:bg}

We introduce the concepts of graph representations and graph positional encodings, Abstract Meaning Representations (AMRs), and the application of Large Language Models (LLMs) to structured input.

\subsection{Graph Representations and Graph Positional Encodings}
\label{sec:bg.graphs}

\textbf{Edge-labeled directed graphs.} AMRs are a prime example of edge-labeled directed graphs. We formally define these as tuples \( \gA = (\gV_\gA, \gE_\gA, \gR_\gA) \in \mathbb{A} \), where \( \gV_\gA \) is the set of \(n_\gA = |\gV_\gA|\) nodes, \( \gE_\gA \subseteq \gV_\gA \times \gV_\gA \) is the set of \(m_\gA = |\gE_\gA|\) directed edges, and \( \gR_\gA \) is a finite set of relation types (edge labels), such that each edge \( (u, v) \in \gE_\gA \) is associated with a label \( r_{u,v} \in \gR_\gA \). The space of these graphs is denoted by \(\mathbb{A}\).

\textbf{Edge-unlabeled directed graphs.} To understand positional encodings on graphs, it is useful to first consider edge-unlabeled directed graphs, defined as tuples \( \gG = (\gV_\gG, \gE_\gG) \in \mathbb{G} \), where \( \gV_\gG \) is the set of \(n_\gG = |\gV_\gG|\) nodes, and \( \gE_\gG \subseteq \gV_\gG \times \gV_\gG \) is the set of \(m_\gG = |\gE_\gG|\) directed, binary, and unlabeled edges, with \( e_{u,v} = 1 \) if there is a directed edge from node \( u \) to node \( v \), and \( 0 \) otherwise. The space of such graphs is denoted by \(\mathbb{G}\). The relational structure is encoded in the adjacency matrix \( \mA \in \{0,1\}^{n_\gG \times n_\gG} \), where \( \mA_{u,v} = e_{u,v} \). We define the out-degree matrix \( \mD \) as a diagonal matrix with \( \smash{\mD_{u,u} = \sum_v \mA_{u,v}} \). Symmetrizing the adjacency matrix as \( \smash{\mA_S = \mA \lor \mA^\top} \) (element-wise logical OR) yields an undirected representation of the graph, with a corresponding symmetrized degree matrix \( \mD_S \). \rebuttal{While this is not \textit{always} necessary, it is required in our approach, as it} allows for the computation of the symmetric normalized Laplacian \( \smash{\mL_S = \mI - \mD_S^{-1/2} \mA_S \mD_S^{-1/2}} \), which has a real eigendecomposition \( \smash{\mL_S = \mU \mLambda \mU^\top} \), where \( \mU \in \R^{n_\gG \times n_\gG} \) contains orthonormal eigenvectors and \( \mLambda = \text{diag}(\lambda_1, \dots, \lambda_{n_\gG}) \in \R^{n_\gG \times n_\gG} \) is a diagonal matrix of real eigenvalues. However, this symmetrization inherently loses the directional information present in the original directed graph.

\textbf{Magnetic Laplacian.} To address the loss of directionality, we can employ the magnetic Laplacian \citep{furutani2020magnetic}, which introduces directional information via complex-valued phase shifts. For \( q \in \R_{\geq 0} \), the magnetic Laplacian \( \smash{\mL^{(q)} \in \C^{n_\gG \times n_\gG}} \) is defined as:
\begin{equation}
    \label{eq:magnetic_laplacian}
    \mL^{(q)} := \mD_S - \mA_S \odot \exp \left(i \boldsymbol{\Theta}^{(q)}\right),
\end{equation}
where \( i = \sqrt{-1} \), \( \odot \) denotes the Hadamard product (element-wise multiplication), and the phase matrix \( \boldsymbol{\Theta}^{(q)} \in \R^{n_\gG \times n_\gG} \) is given by \( (\boldsymbol{\Theta}^{(q)})_{u,v} = 2\pi q ((\mA)_{u,v} - (\mA)_{v,u}) \). The magnetic Laplacian \( \smash{\mL^{(q)}} \) is Hermitian, guaranteeing a complete set of complex eigenvectors \( \bm{\Gamma} \in \C^{n_\gG \times n_\gG} \). 

\textbf{Graph Positional Encodings (GPEs)} assign each node a notion of position withing the graph. Most common approaches leverage the spectral properties of the graph Laplacian to encode the structural position of nodes. In particular, the eigenvectors of the Laplacian provide an orthonormal basis that captures the graph's structure at varying frequencies. 

Following \citep{belkin2003laplacian, dwivedi2021generalization}, we can define the Laplacian-based positional encoding \( \phi(v_i) \in \R^k \) for a node \( v_i \in \gV_\gG \) as the \(i\)-th row of the first \(k\) eigenvectors in \( \mU \):
\begin{equation}
    \label{eq:graph_laplacian_pe}
    \phi(v_i) = \left[\mU_{i,1}, \dots, \mU_{i,k}\right]^\top,
\end{equation}
where \( k \ll n_\gG \) is a chosen dimensionality. These embeddings inherently capture coarse-to-fine structural patterns and are invariant to node permutations.

\subsection{Abstract Meaning Representation}
\label{sec:bg.amr}

Abstract Meaning Representation (AMR) \citep{langkilde1998generation, banarescu2013amr, mansouri2025amrsurvey} is a semantic formalism that represents the meaning of a sentence as a rooted, directed acyclic graph \( \gA = (\gV_\gA, \gE_\gA, \gR_\gA) \). The nodes \( \gV_\gA \) represent concepts, which are typically predicates, entities, or abstract ideas. The directed edges \( \gE_\gA \subseteq \gV_\gA \times \gV_\gA \) capture the semantic relationships between these concepts. Each edge \( (u, v) \in \gE_\gA \) is associated with a label \( r_{u,v} \in \gR_\gA \), where \( \gR_\gA \) is a finite set of predefined semantic roles. Common relation labels include \texttt{:ARG0} (agent), \texttt{:ARG1} (patient/theme), and \texttt{:mod} (modifier). A key characteristic of AMR is its abstraction from surface syntax, ensuring that sentences with equivalent semantics are mapped to isomorphic AMR graphs. For instance, the sentences ``\textit{The child wants the parent to believe them}'' and ``\textit{What the child wanted is for the parent to believe them}'' share the same underlying AMR structure.

\textbf{AMR Linearizations.} To enable the processing of AMR graphs by sequence-to-sequence models, such as LLMs, it is necessary to linearize the graph structure into a sequential format. This process, termed \textit{linearization}, transforms an AMR graph \( \gA \) into a sequence of labels \( \gL_\gA = (\ell_1, \ell_2, \dots, \ell_L) \in \Sigma^\ast \), where each \( \ell_i \) is a label from a predefined vocabulary \( \Sigma \). A common serialization is the Penman notation \citep{kasper1989flexible, bateman1990upper, goodman2020penman}, a parenthetical representation that encodes the graph's concepts and relations in a compact textual form.

More recently, methods like breadth-first search (BFS) and depth-first search (DFS) based linearizations have been extended to AMRs \citep{bevilacqua2021spring}. For example, BFS linearization traverses the graph level by level, employing special tokens to denote relation types and reentrancies, resulting in a structured sequence that aims at preserving the graph's information for autoregressive learning \citep{konstas2017neuralamr}. Additional details and visualizations are provided in \cref{app:amr}. 

\subsection{Blueprint of Large Language Models}
\label{sec:bg.llm}

Large Language Models (LLMs) \citep{vaswani2017attention, devlin2019bert, brown2020language, touvron2023llama} are parameterized functions \( \smash{\pi_\theta : \Lambda^\ast \to \Delta^{|\Lambda|^m-1}} \) mapping an input token sequence \(x \in \Lambda^\ast\) to a probability distribution over output sequences \(y \in \Lambda^m\) of length \(m\). Here, \(\theta\) denotes the model parameters, \(\Lambda\) the token vocabulary, and \(\smash{\Delta^{|\Lambda|^m-1}}\) the probability simplex over \(\smash{\R^{|\Lambda|^m}}\). Outputs are generated autoregressively via  \(\smash{\pi_\theta(y \mid x) = \prod_{t=1}^m \pi_\theta(y_t \mid y_{<t}, x)}\), with \(y_{<t} = (y_1, \dots, y_{t-1})\).

LLMs are typically pretrained on massive text corpora by predicting the next token in a sequence. This enables them to learn intricate linguistic and semantic patterns, resulting in strong generalization capabilities across various natural language processing tasks, often without task-specific supervision. To adapt a pretrained LLM to a specific downstream task, its parameters \( \theta \) are fine-tuned on a task-specific dataset \( \smash{\gD = \{(x^{(i)}, y^{(i)})\}_{i=1}^N} \), where \( \smash{x^{(i)} \in \Lambda^\ast} \) and \( \smash{y^{(i)} \in \Lambda^m} \).

%% file: sections/3-method.tex
\section{Structure-Aware Fine-Tuning for AMR-to-Text Generation}
\label{sec:method}

We present \modelname{}, a lightweight method for fine-tuning pretrained LLMs on AMR-to-text generation by incorporating structural information from the input graph. The key idea is to inject graph positional encodings, derived from the magnetic Laplacian of the AMR graph, into the token embeddings during fine-tuning. This guides the model to better capture graph topology and long-range dependencies.

\textbf{Task Definition.}
Given an AMR graph \( \gA \in \mathbb{A} \), the goal is to generate a natural language sentence \( \gS \in \Sigma^\ast \) such that \( \gS = \psi(\gA) \) is fluent and semantically faithful to the input.

\textbf{Approach.}
\modelname{} enhances LLM decoding by conditioning on structure-aware graph representations. We first apply a semantics-preserving transformation to the AMR graph \pcref{sec:method.amr_transformation}, compute positional encodings from its magnetic Laplacian \pcref{sec:method.amr_pe}, and inject them into the LLM's embedding space during fine-tuning \pcref{sec:method.fine_tuning}. We report a pseudo-code description of our approach in \cref{alg:amr2text_streamlined} and a visualization in \cref{fig:pipeline}.

%%%%%%%%%%%%%%%%%%%%%%%%%%%%%%%%%%%%%%%%%%%%%%%%%%%%%%%%%%%%%%%%%%%%%%%%%%%%%%%%%%%
\subsection{Semantically-Preserving Transformation of AMR Graphs}
\label{sec:method.amr_transformation}

Edge-labeled graphs, such as AMRs, pose a challenge for computing eigenvectors of the graph Laplacian, a standard step in deriving graph positional encodings. Applying the Laplacian directly would necessitate ignoring the crucial semantic information encoded in their edge labels. To overcome this, we introduce a transformation \( \tau \) that converts a linearized AMR into a directed, edge-unlabeled \textit{semantic-preserving graph} (SPG). 
The SPG retains the core semantics of the original AMR structure while enabling the application of Laplacian-based spectral methods for positional encoding.
\rebuttal{This is similar in spirit to the reification used in Semantic Role Labeling or Resource Description Framework graphs \citep{marcheggiani2017encoding, marcheggiani2018deep}, but our transformation is tailored to AMRs and magnetic-Laplacian PEs.}

\textbf{BFS Linearization.}  
We begin by applying a breadth-first search (BFS) traversal of the AMR graph \( \gA \), yielding a linearized sequence of labels
\(\gL_\gA = (\ell_1, \dots, \ell_L) \in \Sigma^\ast\),
where each \( \ell_i \) represents a concept or role label.  Each label corresponds to a node in the SPG \( \gG_\gA = (\gV_\gG, \gE_\gG) \), through an injective function \( \sigma_\gA : \mathbb{Z} \rightarrowtail \gV_\gG \) that aligns labels to their corresponding node in the graphs, i.e, \( \sigma_\gA(i) = v_i \). This implies \( |\gV_\gG| = L \).

\textbf{SPG Transformation.}  
The transformation \( \tau : \Sigma^\ast \times (\Sigma \rightarrowtail \gV_\gG) \to \mathbb{G} \) constructs the SPG \( \gG_\gA = \tau(\gL_\gA, \sigma_\gA) \) from the label sequence and alignment mapping. Labeled edges in the original AMR are represented as role nodes in the SPG, preserving role semantics via directed edges to their source and target concept nodes. We unite co-referring nodes (e.g., marked with \texttt{<P1>}), and merge their connectivity. The resulting SPG is semantically equivalent to the original AMR but uses unlabeled edges for spectral compatibility, and explicits re-entrancies and coreferences.
Additional details and visualization of the semantically-preserving transformation are available in \cref{app:implementation.amr_transformation}.
% \cref{fig:amr_to_spg} provides a visual example. 

\textbf{Tokenization of Node Labels.}  
Each textual label \(\ell_i \in \gL_\gA\) is associated with a node \( v_i \in \gV_\gG \), \( v_i = \sigma_\gA(i) \). We tokenize each node label into a sequence of tokens \(\vt^{(\ell_i)} \in \Lambda^{p_i}\)
\begin{equation}
    \label{eq:node_tokenization}
    \vt^{(\ell_i)} = \textsc{Tokenize}(\ell_i) = (t^{(\ell_i)}_1, \dots, t^{(\ell_i)}_{p_i}),
\end{equation}
where \( \Lambda \) denotes the tokenizer's vocabulary and \( p_i \) is the number of tokens produced from the label \( \ell_i \).  When \( p_i > 1 \), we refer to \( v_i = \sigma_\gA(i) \) as a \textit{multi-token node}. 
%
% We define the linearization token sequence \( \gT_{\gL} \), obtained by concatenating the token sequences across all nodes in their linearized order:
% \begin{equation}
%     \label{eq:linearization_tokenization}
%     \gT_{\gL} = \vt^{(v_1)} \mathbin{\|} \vt^{(v_2)} \mathbin{\|} \dots \mathbin{\|} \vt^{(v_L)} = (t_1, \dots, t_p) \in \Lambda^{p}
% \end{equation}
% where \(p = \sum_{i=1}^{L}p_i\) is the total number of tokens.
% This flattened sequence forms the actual input to the language model. 

%%%%%%%%%%%%%%%%%%%%%%%%%%%%%%%%%%%%%%%%%%%%%%%%%%%%%%%%%%%%%%%%%%%%%%%%%%%%%%%%%%%
\subsection{AMR-Specific Positional Encodings}
\label{sec:method.amr_pe}

In the previous section we defined the transformation from an AMR graph \(\gA\) to its semantically-preserving representation \(\gG_\gA\) that allows us to apply spectral graph theory and compute graph positional encodings.
Here, we present our AMR-specific graph positional encodings, which we compute from the magnetic Laplacian \pcref{sec:bg.graphs} of the semantic-preserving graph \(\gG_\gA\). These encodings are meant to capture the topology of the AMR structure and its directionality.

\textbf{Node-level PEs.}
The magnetic Laplacian, defined in \cref{eq:magnetic_laplacian}, encodes directionality through complex phase shifts. We compute the magnetic Laplacian \( \mL^{(q)} \) of \( \gG_\gA \) and extract the eigenvectors corresponding to the lowest \(k\) eigenvalues, forming a complex matrix \( \bm{\Gamma} \in \mathbb{C}^{n_\gG \times k} \). Each node \( v_i \in \gV_\gG \) is assigned a complex-valued \(k\)-dimensional embedding \(\phi(v_i) \in \mathbb{C}^k\), \(\smash{\phi(v_i) = \left[\bm{\Gamma} _{i,1}, \dots, \bm{\Gamma} _{i,k}\right]^\top}\).
% \begin{equation}
%     \label{eq:graph_mag_laplacian_pe}
%     \phi(v_i) = \left[\bm{\Gamma} _{i,1}, \dots, \bm{\Gamma} _{i,k}\right]^\top.
% \end{equation}
We convert \( \phi(v_i) \) to a real-valued vector \(\smash{\textsc{PE}^{(v_i)} \in \R^{2k}}\) by concatenating the real and imaginary parts:
\begin{equation}
    \label{eq:node_pe}
    \textsc{PE}^{(v_i)} = [\Re(\phi(v_i)) \,\, \Im(\phi(v_i))].
\end{equation}
These positional encodings provide a spectral representation that reflects both local and global graph structure. Nodes with similar structural roles in the AMR, such as arguments or modifiers, will have similar embeddings, even if distant in the graph.

\textbf{Intra-node Token Positional Encodings.}
For each token \( \smash{t^{(\ell_i)}_j} \) in \( v_i =\sigma_\gA(i) \) we apply sinusoidal positional encodings \citep{vaswani2017attention} to preserve their intra-node ordering:
\begin{equation}
    \label{eq:intra_pe}
    \textsc{IntraPE}^{(\ell_i)}_j = \textsc{SinPE}(j), \quad \text{for } j = 1, \dots, p_i,
\end{equation}
where \( \textsc{IntraPE}^{(\ell_i)}_j\in \R^d \).
For single-token nodes (\( p_i = 1 \)), we use \( \textsc{IntraPE}^{(\ell_i)}_1 = \textsc{SinPE}(0) \).

\textbf{AMR Positional Encodings.}
We combine the node-level and intra-node positional encodings to obtain the final AMR-specific positional encoding for each token \( \smash{t^{(\ell_i)}_j} \):
\begin{equation}
    \label{eq:amr_pe}
    \textsc{AmrPE}^{(\ell_i)}_j = f_\vartheta\left(\textsc{PE}^{(v_i)} \,\big\Vert\ \textsc{IntraPE}^{(\ell_i)}_j\right),
\end{equation}
where \( \smash{\textsc{AmrPE}^{(\ell_i)}_j\in \R^{d_\text{emb}}} \), \( f_\vartheta: \R^{2k + d} \rightarrow \R^{d_\text{emb}} \) is a two-layer MLP with GeLU activation function \citep{hendrycks2016gaussian}, and \( \big\Vert \) denotes vector concatenation. This projection defines token-level positional encodings that captures (\textit{i}) structural knowledge from the node-level positional encodings, and (\textit{ii}) label-level sequential information from the intra-node token positional encodings. The embedding is mapped into the LLM embedding space (\(d_\text{emb}\), see \cref{sec:method.fine_tuning}), allowing seamless integration during fine-tuning.
Concatenating the positional encodings across all nodes/labels in their linearized order, as defined by \(\sigma_\gA\), determines the final AMR-specific positional encodings matrix:
\begin{equation}
    \label{eq:amr_pe_all}
    \textsc{AmrPE} = \left( \textsc{AmrPE}^{(\ell_1)}_1 \, \dots \, \textsc{AmrPE}^{(\ell_1)}_{p_1} \, \dots \, \textsc{AmrPE}^{(\ell_2)}_1 \, \dots \, \textsc{AmrPE}^{(\ell_L)}_{p_L} \right)^\top,
\end{equation}
where \(\textsc{AmrPE} \in \R^{p \times d_\text{emb}}\) and \( p = \sum_{i=1}^L p_i \) is the total number of tokens in the linearization. \textsc{AmrPE} is a representation of each token in the linearization that considers both the position of the token within its node-label and the global position in the graph.

% \input{assets/algorithms/amr2text_full}
% \input{assets/algorithms/amr2text}
%%%%%%%%%%%%%%%%%%%%%%%%%%%%%%%%%%%%%%%%%%%%%%%%%%%%%%%%%%%%%%%%%%%%%%%%%%%%%%%%%%%

\subsection{LLM Fine-Tuning with AMR-Specific Positional Encodings}
\label{sec:method.fine_tuning}

For ease of exposition, we represent the pretrained LLM decoder model as a composition:
\[
    \pi_\theta = \textsc{Decode} \circ \textsc{Embed}
\]
where \(\textsc{Embed}: \Lambda^p \to \R^{p \times d_\text{emb}}\) maps a sequence of tokens into the LLM's embedding space, and \(\textsc{Decode}: \R^{p \times d_\text{emb}} \to \Sigma^\ast\) generates the output text sequence.
% \footnote{\textsc{Decode} is an abstraction over decoder components, including the transformer layers, head, and tokenizer.}.
%
Given an AMR graph \(\gA\), we obtain a linearized sequence of node labels \(\smash{\gL_\gA = (\ell_1, \dots, \ell_L)}\) and their corresponding tokenized forms \(\smash{\vt^{(\ell_i)} = (t^{(i)}_1, \dots, t^{(i)}_{p_i})}\). The overall token sequence \( \gT_{\gL} \in \Lambda^p \) is:
\begin{equation}
    \label{eq:linearization_tokenization}
    \gT_{\gL} = \vt^{(\ell_1)} \mathbin{\|} \vt^{(\ell_2)} \mathbin{\|} \dots \mathbin{\|} \vt^{(\ell_L)} = (t_1, \dots, t_p) 
\end{equation}
where \(p = \sum_{i=1}^L p_i\) is the total number of tokens in the linearized graph.
The sequence \(\gT_{\gL}\) is mapped to the LLM embedding space as:
\begin{equation}
    \mX = \textsc{Embed}(\gT_{\gL}) \in \R^{p \times d_\text{emb}}.
\end{equation}

\textbf{Integrating AMR positional encodings.}
We then integrate structure-aware positional encodings to the embedded representation of the linearized sequence of tokens \rebuttal{through the additive representation}
% The final structure-aware embeddings used for decoding are:
\begin{equation}
    \label{eq:integrate_embedding}
    \mH = \mX + \textsc{AmrPE} \in \R^{p \times d_\text{emb}},
\end{equation}
which preserves the dimensionality of the LLM embedding space.
\rebuttal{This modification affects only the input embeddings; the base model's positional encoding mechanism (e.g., RoPE \citep{su2024rope}) remains unchanged and continues to operate inside the attention layers.}
Finally, \(\mH\) is then fed to the LLM decoder components, including the transformer layers, head, and tokenizer, to return the generated output sequence \(\gS \in \Sigma^\ast\), as \(\gS = \textsc{Decode}(\mH)\).

\textbf{Prompt Handling.}
For clarity and modularity, we exclude the prompt segment from the structure-aware positional encoding process. The prompt is tokenized independently from the AMR linearization to avoid disrupting the alignment between graph nodes and tokens. Its tokens are embedded using the standard learned embeddings without any additional positional encodings beyond those already handled by the pretrained model. This design simplifies the architecture and ensures that the inductive bias introduced by our positional encodings applies exclusively to the AMR portion of the input. Additional details are provided in \cref{app:implementation}.

% \textbf{Dimensional Consistency.}
% We ensure that the AMR-specific positional encoding \( \mathrm{AMR\text{-}PE}_j \) matches the embedding dimension \( d_\mathrm{emb} \) of the LLM, via the projection MLP \( f_\vartheta \) (cf. \cref{eq:amr_pe}), so the summation in the embedding space is well-defined and numerically stable.

% \input{figures/pipeline}

%% file: sections/4-experiments.tex
\section{Experiments}
\label{sec:experiments}

We evaluate our structure-aware fine-tuning approach on the AMR 3.0 benchmark. 
First, we show that \modelname{} achieves a new state of the art on sentence-level AMR-to-text generation (\cref{sec:exp.main_results}). We then demonstrate that it \rebuttal{typically} improves over conventional fine-tuning, with gains that become more pronounced at increased structural complexity (\cref{sec:exp.complexity_stratified}). We further extend this analysis to document-level AMRs, where \modelname{} yields even larger improvements (\cref{sec:exp.doc_amr_results}). \rebuttal{To separate the effect of fine-tuning from potential pretraining exposure, we also evaluate state-of-the-art closed models in zero-shot and few-shot settings, showing that strong general-purpose LLMs still struggle with AMR-to-text generation without task-specific supervision (\cref{sec:exp.eval_sota_llms}). Finally, we analyze how injecting \textsc{AmrPE}s affects the geometry of token representations in the model’s latent space (\cref{sec:exp.geometric_analysis}) and report the associated computational overhead (\cref{app:runtime}).}

\subsection{Experimental Setup}
\label{sec:exp.setup}
We use AMR 3.0 (LDC2020T02) \citep{knight2020amr30} and DocAMR, which incorporates discourse structure and inter-sentence dependencies; split details are provided in \cref{app:assets_and_licences}. We fine-tune pretrained LLMs, including LLaMA 3.2 (1B, 3B) \citep{touvron2023llama}, Qwen 2.5 (0.5B, 1.5B, 3B) \citep{bai2023qwen}, and Gemma (2B) \citep{team2024gemma}, with Low-Rank Adaptation (LoRA)\footnote{SAFT is compatible with any parameter-efficient tuning method.}~\citep{hu2022lora}, both with our graph-based positional encoding module (\modelname{}) and without it (FT). Training and inference settings are identical across conditions. All native components of the base models are preserved; for example, rotary positional embeddings (RoPE) remain active\footnote{\rebuttal{\textsc{AmrPE}s do not alter or replace the LLM's own positional encoding.}}. We report BLEU \citep{papineni2002bleu} and chrF \citep{popovic2015chrf} scores using greedy decoding, and additionally BERTScore \citep{zhang2020bertscoreevaluatingtextgeneration} and METEOR \citep{lavie2007meteor} in \cref{app:extra_metrics}. All experiments are implemented in \href{https://github.com/Lightning-AI/litgpt}{\texttt{LitGPT}}; further details are in \cref{app:training}.

\subsection{AMR 3.0 results}
\label{sec:exp.main_results}

We evaluate \modelname{} on AMR 3.0 against widely used AMR-to-text baselines: SPRING \citep{bevilacqua2021spring}, StructAdapt \citep{ribeiro2021structadapt}, and BiBL \citep{cheng2022bibl}. Unless explicitly noted, all models are trained on the same AMR 3.0 training split and evaluated on the official held-out test set. SPRING and BiBL additionally report variants trained with extra heuristically labeled data, which we gray out in \cref{tab:main_results}. Both models rely on linearized AMR inputs and sequence-to-sequence training. \rebuttal{StructAdapt, in contrast, introduces a dedicated graph encoder within an encoder–decoder architecture. Our approach differs in that it operates on modern decoder-only LLMs and injects structural information through graph positional encodings without introducing a separate encoder component.}
\input{assets/tables/results_main_wrap}
The results in \cref{tab:main_results} present a comparative evaluation of prior approaches\footnote{Reported scores are taken directly from the original publications and not reproduced.} \citep{bevilacqua2021spring,cheng2022bibl,ribeiro2021structadapt}, alongside our fine-tuned \rebuttal{decoder-only} LLMs, both with and without the proposed graph positional encoding module (\modelname{}). \rebuttal{While these architectures are not directly matched, this comparison provides useful context for situating decoder-only models within existing AMR-to-text work.} For baseline models, we include results for versions trained with and without extra heuristically labeled data where available. Notably, we find that fine-tuning several LLMs using LoRA \citep{hu2022lora} already yields improvements over earlier models, including those that use extra training data. Our proposed approach, \modelname{}, further boosts performance, highlighting the benefit of incorporating structural information in the form of graph positional encodings during LLM fine-tuning. As shown in \cref{tab:main_results}, \modelname{} \rebuttal{typically} outperforms FT, with aggregate BLEU gains of +0.8 over the FT variant across the different models. However, such aggregate scores can obscure important variation across inputs of different structural complexity. To more accurately characterize when and how \modelname{} helps, we turn to a stratified analysis.

\subsection{Complexity-stratified results}
\label{sec:exp.complexity_stratified}

To evaluate performance variation with input complexity, we stratify the evaluation by AMR graph depth $\delta(\mathcal{A})$, measured on the original AMR $\mathcal{A}$ prior to preprocessing, by grouping examples where \(\delta(\gA) \geq z\) for varying thresholds $z$. We then calculate the BLEU score on these stratified subsets to quantify how our structure-aware fine-tuning approach improves performance at increasing \(\delta(\mathcal{A})\) with respect to conventional fine-tuning. We define the following metric: 
{\small %
\begin{equation} %
    \Delta_{\mathrm{BLEU}}(z) = \text{BLEU}_{\text{\modelname{}}}^{z} - \text{BLEU}_{\text{FT}}^{z},    
\end{equation}
}%
where \(\smash{\text{BLEU}_{\text{\modelname{}}}^{z}}\) and \(\smash{\text{BLEU}_{\text{FT}}^{z}}\) represent the BLEU scores achieved by \modelname{} and conventional fine-tuning (FT), respectively, on instances with AMR depth $\delta(\mathcal{A}) \geq z$.
\cref{fig:stratified_eval.absolute} shows the extent to which \modelname{} improves the performance of LLMs at increasing levels of semantic complexity. On semantically complex AMRs (depth \(\delta(\gA) \geq 8\)), \modelname{} surpasses FT by +1.1 to +4.4 BLEU.
% \cref{fig:stratified_eval.absolute}.

The divergence of the curves at greater depths and \rebuttal{an overall upward trend} across all models highlight the increasing importance of modeling structure explicitly as semantic complexity grows. While Gemma 2B showed little overall improvement across the full dataset \pcref{tab:main_results}[see], the benefit of \modelname{} becomes more pronounced on examples with greater semantic complexity.

\sloppy To further assess how the benefit varies with respect to depth-one AMRs (i.e., \(\delta(\gA) = 1\)), we define a second-order delta which measures the change in improvement relative to these simple structures:
\begin{equation}
    \smash{\Delta_{\mathrm{BLEU}}^{1}(z) = \Delta_{\mathrm{BLEU}}(z) - \Delta_{\mathrm{BLEU}}(1)}.
\end{equation}
\cref{fig:stratified_eval.relative} further validates the finding that \modelname{} delivers increasing gains on more complex AMRs. Specifically, it shows the relative improvement of each model compared to its own performance on depth-one AMRs. The positive upward trends across all models indicate that the advantage of incorporating graph structure grows with semantic complexity. 
This consistent behavior across model families and sizes reinforces the scalability and applicability of our method across different architectures and parameter scales.

\input{figures/stratified_eval_horiz}
\subsection{\textsc{DocAMR} results}
\label{sec:exp.doc_amr_results}
To understand even further how \modelname{} impacts performance on highly complex AMRs, we evaluate both \modelname{} and standard fine-tuning (FT) on \textsc{DocAMR}, a supplementary subset of AMR 3.0 \citep{knight2020amr30} whose test set consists of sentence-level AMRs from the standard AMR test split, merged into documents with annotated coreference edges spanning sentences. Notably, we evaluate models that were conventionally fine-tuned (FT) and structurally-aware fine-tuned (\modelname{}) on sentence-level AMRs (AMR 3.0), testing their \rebuttal{cross-dataset generalization} capabilities on document-level AMRs (\textsc{DocAMR}). This setup allows us to assess the models’ ability to generalize to cross-sentence semantics without explicit document-level supervision.

As shown in \cref{fig:docamr_results}, \modelname{} consistently improves performance across all levels of complexity which is measured by $\#_\mathrm{AMR}$, the number of AMR graphs contained within a document-level AMR, indicating the overall size and structural density of the input. While the downward trend shows that all models struggle with deep topologies, the consistent, and occasionally increasing, improvement shows that using \modelname{} improves performance on more structurally-dense inputs. The performance gap between \modelname{} and conventional fine-tuning (FT) widens with increasing AMR complexity, suggesting that structural inductive bias becomes increasingly crucial in complex document-level generation.
 This reinforces our central claim that structure-aware fine-tuning is especially beneficial when models must reason over longer contexts and inter-sentential relations. We excluded the Gemma model from this experiment due to its limited context window (4096 tokens), which could not accommodate DocAMR inputs.
\input{figures/doc_amr_eval}

\textbf{Results summary.} Conventional fine-tuning (FT) of LLMs already surpasses prior non-LLM baselines on AMR-to-text generation. \modelname{}, our structure-aware fine-tuning (\modelname{}) method, further improves performance across model families and scales. The improvements are especially consistent on semantically complex and document-level inputs, confirming that graph-based positional encodings enhance the model’s ability to reason over AMR structure.
\input{assets/tables/gpt_results}

\vspace{-1em}

\rebuttal{\subsection{Evaluation under zero-shot and few-shot prompting on SoTA LLMs}}
\label{sec:exp.eval_sota_llms}

\rebuttal{
    We additionally evaluate two closed-source large language models, GPT-4o-mini and GPT-4o, in both zero-shot and few-shot settings. We use a fixed task prompt with seven AMR–text examples; the full template is provided in \cref{app:prompts}.
    As shown in \cref{tab:zeroshot_ft_saft}, moving from zero-shot to few-shot leads to only marginal improvements for both models. Despite their scale, their performance remains substantially below that of a much smaller 3B model fine-tuned on AMR. In fact, standard fine-tuning (FT) of Qwen 2.5 (3B) already more than doubles the BLEU score of GPT-4o-mini, and \modelname{} further improves on FT using the same backbone.
    This suggests that AMR-to-text generation is not easily solved through prompting alone under our prompting setup. In-context examples provide limited signal about the underlying graph structure, whereas fine-tuning exposes the model to the full supervision of the training set and allows \modelname{} to inject explicit structural information directly.
}%

% \vspace{-2.5em}
\rebuttal{%
    \subsection{Geometric analysis of \textsc{AmrPE}s}  
    \label{sec:exp.geometric_analysis}
    We investigate how the integration of structural information via our \textsc{AmrPE}s (\cref{eq:amr_pe_all}) modifies the token embeddings \(\mX\), generating the representations \(\mH=\mX+\textsc{AmrPE}\) (\cref{eq:integrate_embedding}).  \cref{fig:embedding_dynamics} summarizes our results.
    \textbf{(a) Norm in the same order of magnitude.} \textsc{AmrPE}s have a larger norm than token embeddings, with a median ratio of \(\|\textsc{AmrPE}\|_2 / \|\mX\|_2 \approx 3\), but remain within the same order of magnitude. This implies that the structural signal is not obscured by the text token embeddings, nor does it obscure them in turn.
    \textbf{(b) Orthogonal direction.} The cosine between \(\textsc{AmrPE}\) and \(\mX\) is concentrated around zero, indicating that structural embeddings add information about directions that are generally orthogonal to the textual subspace. This suggests an enrichment rather than an overwriting of semantic content.
    \textbf{(c) Variance concentrated on a few axes.} Projecting \textsc{AmrPE}s in the PCA basis of \(\mX\) reveals that most of its variance is concentrated on a small number of principal directions, with a single dominant component carrying most of the information. The structural signal is therefore highly directional rather than diffuse.
    \textbf{(d) Consistent shift.} Projecting \(\mX\) and \(\mH\) into the PCA space of \(\mX\), we observe that \textsc{AmrPE}s induce a consistent and directed shift of \(\mX\) along a few dominant directions, with the residual variance distributed over multiple weaker components.
    \textbf{(e) Greater intrinsic dimensionality.} The spectrum of explained variance of \(\mH\) is flatter than that of \(\mX\), requiring many more components to achieve the same variance thresholds. This reflects a clear increase in intrinsic dimensionality.
    Overall, this geometric separation suggests a potential mechanism by which SAFT may avoid interfering with semantic representations, though direct evaluation of broader capabilities is left for future work.
}%

\input{figures/embedding_dynamics}

%% file: assets/tables/results_main_wrap.tex
\begin{wraptable}[30]{!T}{0.55\textwidth}
    % \small
    \centering
    % \vspace{-.2em}
    \vspace{-1.3em}
    \caption{
    \textbf{\modelname{} achieves state-of-the-art results on AMR-to-text generation.}
    We report BLEU/CHRF on AMR 3.0, comparing: prior work, and our LLMs (FT vs. \modelname{}). All models use identical training data unless marked `+'. Best results per metric (excluding those using additional data) are highlighted in bold. \rebuttal{We report SAFT (ours) relative improvement over FT in brackets.}}
    \label{tab:main_results}
    \vspace{-0.5em}
    \resizebox{\linewidth}{!}{%
    \begin{tabular}{llll}
        \toprule
        \textbf{Model} & \textbf{Variant} &  \textbf{BLEU} $\uparrow$ & \textbf{CHRF} $\uparrow$ \\
        \midrule
        \multicolumn{4}{l}{\textit{Previous Work}} \\
        \midrule
        \multirow{2}{*}{BiBL}
            & w/o Extra data &  47.4 & 74.5 \\
            & \textcolor{gray!50}{+ Extra data} &  \textcolor{gray!50}{50.7} & \textcolor{gray!50}{76.7} \\
        \multirow{2}{*}{SPRING}
            & w/o Extra data &  44.9 & 72.9 \\
            & \textcolor{gray!50}{+ Extra data}  & \textcolor{gray!50}{46.5} & \textcolor{gray!50}{73.9} \\
        StructAdapt & w/o Extra data &  48.0 & 73.2 \\
        \midrule
        \multicolumn{4}{l}{\textit{Our Finetuned LLMs: trained without extra data}} \\
        \midrule
        \multirow{2}{*}{LLaMA 3.2 (3B)} 
        & FT &  53.5 & 75.5 \\
        & \cellcolor{tumblue!15}\modelname{} &  \cellcolor{tumblue!15}\textbf{54.2} \scriptsize{(+1.3\%)} & \cellcolor{tumblue!15}\textbf{76.0} \scriptsize{(+0.7\%)} \\
        \midrule
        \multirow{2}{*}{LLaMA 3.2 (1B)} 
        & FT &  45.5 & 70.9 \\
        & \cellcolor{tumblue!15}\modelname{} &  \cellcolor{tumblue!15}47.8 \scriptsize{\textbf{(+5.1\%)}} & \cellcolor{tumblue!15}71.9 \scriptsize{(+1.4\%)} \\
        \midrule
        \multirow{2}{*}{Qwen 2.5 (3B)} 
        & FT &  51.6 & 72.1 \\
        & \cellcolor{tumblue!15}\modelname{} &  \cellcolor{tumblue!15}51.9 \scriptsize{(+0.6\%)} & \cellcolor{tumblue!15}74.8 \scriptsize{\textbf{(+3.7\%)}} \\
        \midrule
        \multirow{2}{*}{Qwen 2.5 (1.5B)} 
        & FT &  50.5 & 73.7 \\
        & \cellcolor{tumblue!15}\modelname{} &  \cellcolor{tumblue!15}51.7 \scriptsize{(+2.4\%)} & \cellcolor{tumblue!15}74.5 \scriptsize{(+1.1\%)} \\
        \midrule
        \multirow{2}{*}{Qwen 2.5 (0.5B)} 
        & FT &  42.7 & 69.0 \\
        & \cellcolor{tumblue!15}\modelname{} &  \cellcolor{tumblue!15}42.9 \scriptsize{(+0.5\%)} & \cellcolor{tumblue!15}69.3 \scriptsize{(+0.4\%)} \\
        \midrule
        \multirow{2}{*}{Gemma (2B)} 
        & FT &  52.9 & 73.5 \\
        & \cellcolor{tumblue!15}\modelname{} &  \cellcolor{tumblue!15}52.9 \scriptsize{(+0.1\%)\(^\ast\)} & \cellcolor{tumblue!15}73.6 \scriptsize{(+0.1\%)} \\
        \bottomrule
    \end{tabular}
    }
    \flushleft{\scriptsize{\(^\ast\)Not rounded scores: 52.87 (FT) vs. 52.91 (SAFT)}.}
\end{wraptable}

%% file: figures/stratified_eval_horiz.tex
\begin{figure}[!t]
    \centering
    \begin{subfigure}[t]{0.49\textwidth}
        \centering
        \includegraphics[width=\linewidth]{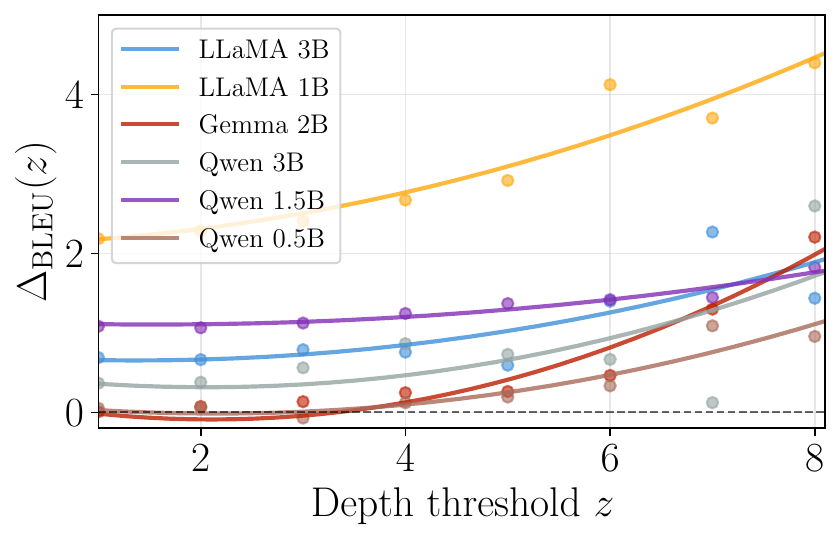}
        \caption{Absolute improvement (\(\Delta_{\mathrm{BLEU}}(z)\)).}
        \label{fig:stratified_eval.absolute}
    \end{subfigure}
    \hfill
    \begin{subfigure}[t]{0.49\textwidth}
        \centering
        \includegraphics[width=\linewidth]{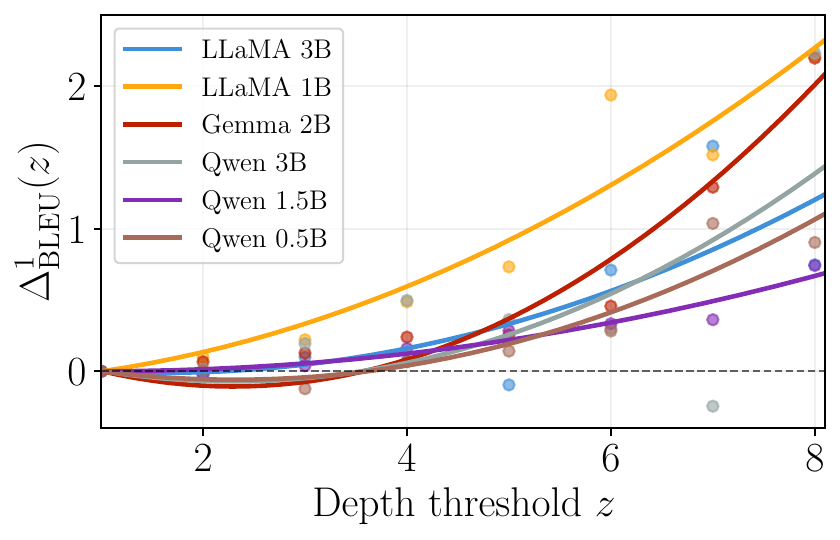}
        \caption{Relative improvement (\(\smash{\Delta_{\mathrm{BLEU}}^{1}}(z)\)).}
        \label{fig:stratified_eval.relative}
    \end{subfigure}
    
    \caption{
    \textbf{BLEU score improvements} of structurally-aware fine-tuned (\modelname{}) models over conventionally fine-tuned (FT) counterparts, on AMRs of depth $\delta(\mathcal{A}) \geq z$. 
    (a)~\textbf{Absolute improvement} (\(\Delta_\text{BLEU}\)): differences in BLEU between \modelname{} and FT models across graph depths and model families.
    % are consistently positive across depths and model families, with only minor deviations.  
    (b)~\textbf{Relative improvement} (\(\smash{\Delta^{1}_\text{BLEU}}\)): differences in BLEU between \modelname{} and FT models across graph depths and model families normalized by performance at depth-1 graphs.  
    Both plots reveal an increasing advantage of \modelname{} as structural complexity grows, demonstrating its effectiveness in leveraging graph topology for improved generation. Lines are 2nd-degree polynomial fits.
    }
    \label{fig:stratified_eval}
\end{figure}

%% file: figures/doc_amr_eval.tex
\begin{figure}[!t]
    \centering
    \begin{subfigure}[t]{0.328\textwidth}
        \includegraphics[width=\linewidth]{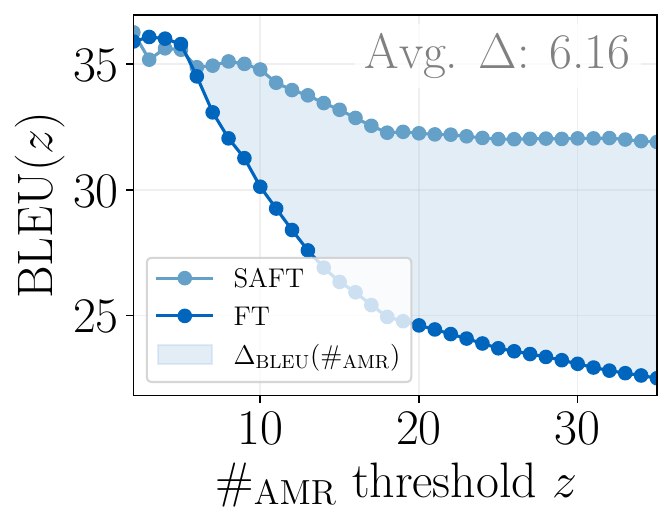}
        \caption{Qwen 2.5 3B}
        \label{fig:qwen1.5b}
    \end{subfigure}
    \hfill
    \begin{subfigure}[t]{0.328\textwidth}
        \includegraphics[width=\linewidth]{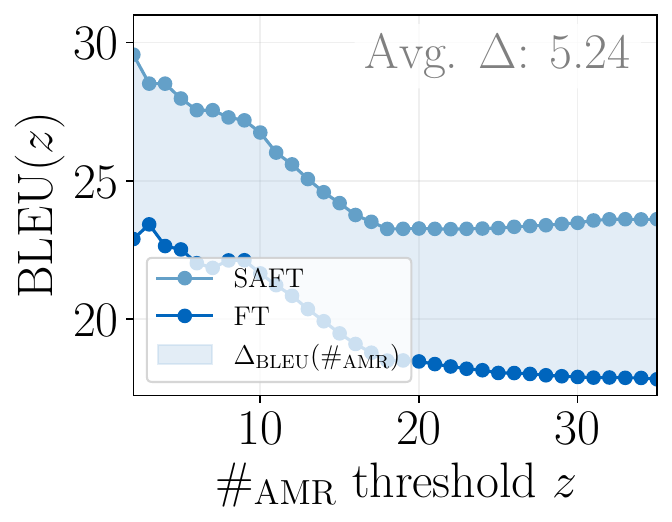}
        \caption{LLaMA 3.2 1B}
        \label{fig:qwen3b}
    \end{subfigure}
    \hfill
    \begin{subfigure}[t]{0.328\textwidth}
        \includegraphics[width=\linewidth]{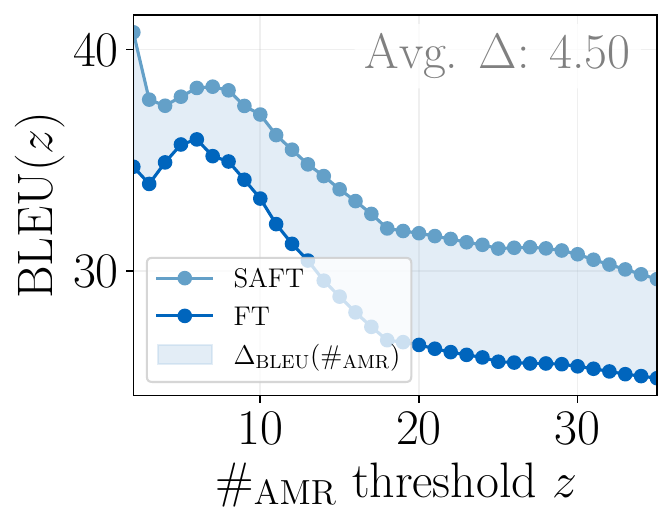}
        \caption{LLaMA 3.2 3B}
        \label{fig:llama3b}
    \end{subfigure}
\caption{
\textbf{\modelname{} demonstrates increasing gains over standard fine-tuning (FT) as document complexity increases.} Performance on the DocAMR test set: Each plot shows the BLEU score improvement of \modelname{} over FT models, evaluated cumulatively on document-level AMRs with \(\#_{\mathrm{AMR}} \leq z\), where \(\#_{\mathrm{AMR}}\) denotes the number of AMR graphs contained in a document. This bottom-up stratified evaluation reveals how \modelname{} performs on increasingly complex document structures. On average across document sizes, \modelname{} outperforms FT models by +6.16, +5.24, and +4.50 BLEU for Qwen 2.5 3B, LLaMA 3.2 1B, and LLaMA 3.2 3B, respectively.
}

\label{fig:docamr_results}
\end{figure}

%% file: assets/tables/gpt_results.tex
\begin{table}[!t]
    % \small
    \centering
    \vspace{-1.2em}
    \caption{
    \textbf{Comparison of zero-shot, \rebuttal{few-shot}, and finetuned models.} 
    We report BLEU, CHRF, BERTScore, and METEOR. Best results per metric are bolded.
    }
    \label{tab:zeroshot_ft_saft}
    \resizebox{0.8\linewidth}{!}{%
    \begin{tabular}{l l c c c c}
        \toprule
        \textbf{Model} & \textbf{Setting} & \textbf{BLEU} $\uparrow$ & \textbf{CHRF} $\uparrow$ & \textbf{METEOR} $\uparrow$ & \textbf{BERTScore} $\uparrow$ \\
        \midrule
        GPT-4o-mini & Zero-shot & 16.3 & 44.8 & 38.4 & 74.8 \\
                    & \rebuttal{Few-shot} & \rebuttal{17.0} & \rebuttal{45.2} & \rebuttal{39.9} & \rebuttal{75.7} \\
        \midrule
        \rebuttal{GPT-4o} & \rebuttal{Zero-shot}     & \rebuttal{28.0} & \rebuttal{59.8} & \rebuttal{52.1} & \rebuttal{81.1} \\
                    & \rebuttal{Few-shot}      & \rebuttal{29.6} & \rebuttal{60.1} & \rebuttal{52.4} & \rebuttal{81.8} \\
        \midrule
        \multirow{2}{*}{Qwen 2.5 (3B)} 
        & FT & 51.6 & 72.1 & 59.4 & 82.9 \\
        & \cellcolor{tumblue!30}SAFT (ours) 
        & \cellcolor{tumblue!30}\textbf{51.9} 
        & \cellcolor{tumblue!30}\textbf{74.8} 
        & \cellcolor{tumblue!30}\textbf{60.1}
        & \cellcolor{tumblue!30}\textbf{83.7}  \\
        \bottomrule
    \end{tabular}
    }
\end{table}

% | Model          | Setting   | BLEU | METEOR | CHRF  | BERTScore |
% |----------------|-----------|------|--------|-------|-----------|
% | GPT‑4o-mini    | zero‑shot | 16.3 | 38.4   | 44.8  | 74.8      |
% | GPT‑4o-mini    | few‑shot  | 17.0 | 39.9   | 45.2  | 75.7      |
% | GPT‑4o         | zero‑shot | 28.0 | 52.1   | 59.8  | 81.1      |
% | GPT‑4o         | few‑shot  | 29.6 | 52.4   | 60.1  | 81.8      |
% | Qwen 2.5 (3B)  | FT        | 51.6 | 59.4   | 72.1  | 82.9      |
% | Qwen 2.5 (3B)  | SAFT  | **51.9** | **60.1** | **74.8** | **83.7** |

%% file: figures/embedding_dynamics.tex
\begin{figure}[t]
    \centering
    \includegraphics[width=\linewidth]{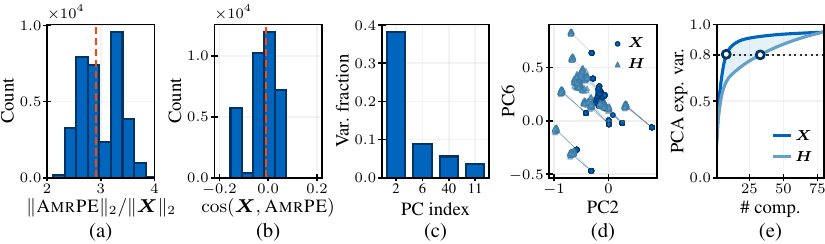}
    \caption{
        \rebuttal{
            \textbf{Effect of AMR positional embeddings (\textsc{AmrPE}) on token representation geometry.}
            (a) \textsc{AmrPE} has a larger but same-order magnitude \(\mX\).
            (b) It is injected along directions largely orthogonal to \(\mX\).
            (c) Its variance concentrates on a small number of PCA directions of \(\mX\).
            (d) In this PCA space, \(\mH=\mX+\textsc{AmrPE}\) shifts coherently along the dominant axis.
            (e) \(\mH\) exhibits a flatter explained-variance spectrum, indicating increased intrinsic dimensionality.
        }
    }
    \label{fig:embedding_dynamics}
\end{figure}

%% file: sections/5-related-works.tex
\vspace{1em}

\section{Related Work}
\label{sec:related_work}

To the best of our knowledge, \modelname{} is the first work to inject graph positional encodings into LLMs for structure-aware fine-tuning without modifying the model's architecture. \modelname{} is architecture-agnostic: it introduces relational inductive bias through precomputed, parameter-free encodings (from the magnetic Laplacian spectrum), injected via a lightweight projection into the LLM's embedding space. This eliminates the need for graph-specific training while allowing direct fine-tuning of the LLM. Prior work on graph-to-text generation, particularly AMR-to-text, can be grouped into three families: \textbf{linearization-based}, \textbf{adapter-based}, and \textbf{graph-tuning-based}. We discuss here some of the main approaches in each family and report additional details in \cref{app:related_work}.

\textbf{Linearization-based.} Graphs are serialized into sequences for seq2seq models such as BART or T5. Examples include SPRING \citep{bevilacqua2021spring}, AMR-BART \citep{bai2022graph}, and BiBL \citep{cheng2022bibl}, which differ in traversal strategies and auxiliary tasks. LLMs have also been adapted via fine-tuning \citep{raut2025can, mager2020gpt} or prompting \citep{yao2024semantic, jin2024analyzing}. Similar ideas extend beyond AMR to molecular graphs \citep{zheng2024llmsdrug}, tables \citep{fang2024llmstabular}, and 3D meshes \citep{wang2024llamamesh}.

\textbf{Adapter-based.} These approaches introduce graph-native components, such as GNN-based adapters or modified attention layers, to inject relational information. StructAdapt \citep{ribeiro2021structadapt} uses graph-aware adapters within pretrained transformers, while others directly encode AMR graphs via graph-to-sequence models \citep{zhu2019modeling, song2018graph2seq, wang2020amr}.

\textbf{Graph-tuning-based.} Recent methods integrate learned graph modules with LLMs, for example by training GNN-based adapters \citep{huang2024graphadapter}, adding graph transformers at each layer \citep{chai2023graphllm}, or prepending GNN-derived embeddings into the prompt \citep{tang2024graphgpt}. While effective, these strategies require additional trainable modules, architectural changes, or costly pretraining. 

%% file: sections/6-conclusion.tex
\section{Conclusion}

% We present \modelname{}, a structure-aware fine-tuning strategy that enriches LLMs with relational information through graph positional encodings. We specifically apply our approach to AMR-to-text, a challenging task in NLP. Our method improves generation quality over standard approaches and shows better potential than conventional fine-tuning. We empirically show that structural inductive bias can enhance LLMs' ability to reason over graphs on this particular task. Gains are amplified as AMR complexity increases, suggesting that structure-aware tuning helps capture long-range semantic dependencies. We validate this across both sentence- and document-level AMR benchmarks (AMR 3.0 and DocAMR). Our findings offer a path forward for incorporating graph positional information into LLMs for a wide range of graph-to-text and graph-centric tasks.

We introduce \modelname{}, a structure-aware fine-tuning strategy that injects relational inductive bias into LLMs using graph positional encodings derived from AMR structures. Applied to AMR-to-text generation—a challenging task requiring deep semantic understanding—our approach consistently improves generation quality over conventional fine-tuning and non-LLM-based baselines. We find that performance gains are most pronounced as AMR complexity increases, indicating that structural guidance is particularly valuable for modeling long-range dependencies and rich graph semantics. These results hold across both sentence-level (AMR 3.0) and document-level (DocAMR) benchmarks. Our findings demonstrate that integrating structural signals into LLMs can enhance their reasoning over graph-structured inputs, and we believe this opens the door to broader applications of graph-aware fine-tuning across graph-to-text and other graph-centric tasks.

% - introduce SAFT: structure aware fine-tuning of LLMs with graph-positional encodings derived from AMRs. We show that already fine-tuning LLMs on this tasks shows potential vs common baselines, and adding structural information through our model further improves vs normal fine-tuning.
% - we further show that when the complexity of the amr increases (meaning more semantic information and bigger graphs, i.e., larger receptive fields) adding structural information to the fine-tuning of the model improves performance across the board. we validate these results on AMR 3.0 (sentence-level) and DocAMR (document-level)
% - we hope this work will pave the way to study the integreation of graph poesitional encodings in LLMs not only to AMR-to-text but other graph-based tasks

%% file: sections/99-reproducibility-statement.tex
\section*{Reproducibility Statement}
We have taken deliberate steps to ensure the reproducibility of our results. The implementation of SAFT builds directly on the open-source \href{https://github.com/Lightning-AI/litgpt}{\texttt{LitGPT}} framework, with all modifications described in \cref{app:training} and detailed pseudo-code in \cref{app:algorithm}. Hyperparameter configurations, training schedules, and model-specific settings are reported in \cref{app:training,tab:hyperparams,tab:hyperparams_FT}. Datasets (AMR~3.0 and DocAMR) are described in \cref{app:datasets}, including licensing details. Our evaluation uses standard metrics (BLEU, ChrF, BERTScore, METEOR), with references provided in \cref{sec:experiments} and \cref{app:extra_metrics}. Computational overhead analyses are included in \cref{app:runtime}, and zero-shot baselines are reported in \cref{sec:exp.eval_sota_llms}.

\section*{Use of Large Language Models}
Large language models (LLMs) are a central component of our proposed method: 
we fine-tune pretrained LLMs as the backbone of our approach. 
In addition, LLMs were used as a writing aid to polish the text of the paper, 
improving clarity, grammar, and style. 
LLMs were not used to generate research ideas, design experiments, analyze results, 
or draw conclusions. All conceptual and scientific contributions are the responsibility of the authors.

%% file: sections/appendix/A-amr.tex
\section{Abstract Meaning Representation}
\label[appendix]{app:amr}

\subsection{Reference sentence}
\label[appendix]{app:amr.reference_sentence}
We report here the representations of the sentence used in the main body (\cref{sec:bg.amr}) as reference:
\begin{quote}
    \textit{The child wants the parent to believe them.}
\end{quote}
\cref{fig:amr.graph} is the graph representation of the sentence, while \cref{fig:amr.penman,fig:amr.bfs} are the Penman and BFS linearizations, respectively.
\input{figures/amr_representations_horiz}

\subsection{Visualizing other AMR examples}
\label[appendix]{app:amr.other_sentences}
We visualize the linearization and preprocessing steps, as detailed in \cref{app:implementation.amr_transformation} for additional sentences. \cref{tab:amr_examples_with_sentences} summarizes the examples and points to the corresponding figures showing the original AMR graphs and the transformation pipeline.

\input{assets/tables/amr_examples}

%% file: figures/amr_representations_horiz.tex
\begin{figure}[!ht]
\centering
\begin{subfigure}{0.32\linewidth}
    \centering
    \input{assets/tikz/amr_graph}
    \vspace*{0.45em}
    \caption{Graph representation}
    \label{fig:amr.graph}
\end{subfigure}
\hfill
\begin{subfigure}{0.28\linewidth}
    \centering
    \input{assets/tikz/amr_penman}
    \vspace*{-0.6em}
    \caption{Penman notation}
    \label{fig:amr.penman}
\end{subfigure}
\hfill
\begin{subfigure}{0.28\linewidth}
    \centering
    \input{assets/tikz/amr_bfs}
    \vspace*{-0.6em}
    \caption{BFS linearization}
    \label{fig:amr.bfs}
\end{subfigure}

\caption{Three aligned representations of the sentence \textit{``The child wants the parent to believe them.''}: (a) a graph-based AMR structure, (b) its corresponding Penman notation, and (c) a BFS linearization used for sequence-based processing.}
\label{fig:amr}
\end{figure}

%% file: assets/tikz/amr_graph.tex
\begin{tikzpicture}[
    ->, 
    >=Stealth, 
    font=\scriptsize, 
    node distance=1cm and 1.5cm,
    every node/.style={font=\scriptsize},
    concept/.style={
        % draw, 
        shape=rectangle,
        minimum height=4mm, 
        minimum width=5mm, 
        align=center,
        inner sep=2pt
    }
  ]
  % Nodes
  \node[concept] (w) at (0,0) {\textbf{want-01}};
  \node[concept] (b) at (-1.3,-1.1) {\textbf{believe-01}};
  \node[concept] (c) at (1.3,-1.1) {\textbf{child}};
  \node[concept] (p) at (0,-2.2) {\textbf{parent}};

  % Edges
  \draw[->] (w) -- (b) node[midway, fill=white, inner sep=1pt] {\role{:ARG1}};
  \draw[->] (w) -- (c) node[midway, fill=white, inner sep=1pt] {\role{:ARG0}};
  \draw[->] (b) -- (c) node[midway, fill=white, inner sep=1pt] {\role{:ARG1}};
  \draw[->] (b) -- (p) node[midway, fill=white, inner sep=1pt] {\role{:ARG0}};
\end{tikzpicture}

%% file: assets/tikz/amr_penman.tex
\begin{tikzpicture}[every node/.style={font=\ttfamily\scriptsize}]
    \node[text width=\linewidth] (amr_penman) {
        (\variable{w} / \concept{want-01} \\
        \hspace*{1em}\role{:ARG0} (\variable{c} / \concept{child}) \\
        \hspace*{1em}\role{:ARG1} (\variable{b} / \concept{believe-01} \\
        \hspace*{2em}\role{:ARG0} (\variable{p} / \concept{parent}) \\
        \hspace*{2em}\role{:ARG1} \variable{c}))
    };
\end{tikzpicture}

%% file: assets/tikz/amr_bfs.tex
\begin{tikzpicture}[every node/.style={font=\ttfamily\scriptsize}, node distance=0.2cm and 0.5cm]
    % Nodes
    \node[anchor=west, text width=\linewidth] (amr_bfs) {
        \pointer{0} \concept{want-01} \role{:ARG0} \pointer{1} \concept{child} \role{:ARG1} 
        \pointer{2} \concept{believe-01} \stopwordamr{<stop>} 
        \pointer{2} \role{:ARG0} \pointer{3} \concept{parent} \role{:ARG1} \pointer{1} \stopwordamr{<stop>}
    };
\end{tikzpicture}

%% file: assets/tables/amr_examples.tex
\begin{table}[!ht]
\centering
\caption{Examples with sentences, AMR graphs, and preprocessing steps}
\resizebox{\linewidth}{!}{%
\begin{tabular}{l p{7.2cm} l l}
\toprule
\textbf{Ex.} & \textbf{Sentence} & \textbf{Original AMR Graph} & \textbf{Preprocessing Steps} \\
\midrule
I   & I used to play tennis. & ~\cref{fig:example1_amr} & \cref{fig:preprocessing_example_1} \\
II  & This is really eye-opening. & ~\cref{fig:example2_amr} & \cref{fig:preprocessing_example_2} \\
III & The key is to be as objective as possible. & ~\cref{fig:example3_amr} & \cref{fig:preprocessing_example_3} \\
IV  & Speeding and accidents have surged as well. & ~\cref{fig:example4_amr} & \cref{fig:preprocessing_example_4} \\
\bottomrule
\end{tabular}
}
\label{tab:amr_examples_with_sentences}
\end{table}

%% file: sections/appendix/B-implementation_details.tex
\section{Implementation details}
\label[appendix]{app:implementation}

\subsection{Semantically-preserving AMR transformation}
\label[appendix]{app:implementation.amr_transformation}

\input{figures/amr_to_spg_steps}
We report detailed information of our proposed semantically-preserving transformation \( \tau : \Sigma^\ast \times (\Sigma \rightarrowtail \gV_\gG) \to \mathbb{G} \) of AMR graph (\cref{sec:method.amr_transformation}).

Given a linearization (label sequence) \(\gL_\gA\) and alignment \(\sigma_\gA\)---as discussed in \cref{sec:method.amr_transformation} and shown in \cref{fig:amr_to_spg_steps.linearization}---the transformation \(\tau\) constructs the SPG \( \gG_\gA = \tau(\gL_\gA, \sigma_\gA) \) through the following steps:
\begin{enumerate}
    \item \textbf{Substructure Construction} (\(\textsc{ToSubgraph}\), \cref{fig:amr_to_spg_steps.substructures}): 
    \(\gL_\gA\) is segmented at each \texttt{<stop>} token. Each segment defines a local subgraph rooted at a head concept and includes its outgoing role-labeled edges (e.g., \texttt{:ARG0}) and target nodes.
    \[
        \{\bar{\gA}_i\}_i = \textsc{ToSubgraph}(\gL_\gA).
    \]
    
    \item \textbf{Edge-to-Node Conversion} (\(\textsc{RoleExpand}\), \cref{fig:amr_to_spg_steps.role_stop_order}):  
    Each labeled edge \((u \xrightarrow{r} v)\) is expanded into a role node \(r\), creating two unlabeled edges: 
    \((u \to r)\) and \((r \to v)\). This yields a directed graph with no edge labels.
    \[
        \bar{\gG}_i = \textsc{RoleExpand}(\bar{\gA}_i).
    \]
    
    \item \textbf{Stop Node Re-insertion} (\(\textsc{AddStopNodes}\), \cref{fig:amr_to_spg_steps.role_stop_order}):  
    The \texttt{<stop>} labels are inserted in each subgraph as a special terminal node. These nodes mark the end of node expansions and, alongside \(\sigma_\gA\), support alignment between graph nodes and tokens in \(\gL_\gA\).
    \[
        \hat{\gG}_i = \textsc{AddStopNodes}(\bar{\gG}_i).
    \]
    
    \item \textbf{Node Ordering Assignment} (\(\sigma_\gA^{-1}\), \cref{fig:amr_to_spg_steps.role_stop_order}):  
    Assign to each node an index inherited from the BFS order to preserve alignment between token positions in \(\gL_\gA\) and graph nodes in \(\gG\).
    \[
        i_v = \sigma^{-1}_\gA(v), \quad \forall v \in \hat{\gG}_i.
    \]
    
    \item \textbf{Pointer-Based Merging} (\(\textsc{Merge}\), \cref{fig:amr_to_spg_steps.merge}):  
    For each pointer index \(j\) (e.g., \texttt{<P2>}), identify co-referring nodes \(\gU_j = \{u_1, \dots, u_k\}\) such that all \(u_i\) share pointer \(j\). Then:
    \begin{enumerate}
        \item Merge \textit{incoming} edges:  
        \(\gE^{\text{in}}_{\gU_j} = \bigcup_{i=1}^{k} \gE^{\text{in}}(u_i)\),  
        with \(\gE^{\text{in}}(u_i) = \{ (v, u_i): (v, u_i) \in \gE_\gA \}\).  
        \item Merge \textit{outgoing} edges: 
        \(\gE^{\text{out}}_{\gU_j} = \bigcup_{i=1}^{k} \gE^{\text{out}}(u_i)\), 
        with \(\gE^{\text{out}}(u_i) = \{ (u_i, v): (u_i, v) \in \gE_\gA \}\)
    
       \item Update the connectivity for each \(u_i \in \gU_j\):  
        \(\gE^{\text{in}}(u_i) := \gE^{\text{in}}_{\gU_j},\quad \gE^{\text{out}}(u_i) := \gE^{\text{out}}_{\gU_j}\)
    \end{enumerate}
    \[
        \gG_\gA = \textsc{Merge}(\{\bar{\gG}_i\}_i).
    \]
    
\end{enumerate}

\subsection{Algorithm}
\label[appendix]{app:algorithm}

\input{assets/algorithms/amr2text_full}

% Labeled edges in the original AMR are represented as role nodes in the SPG, preserving role semantics via directed edges to their source and target concept nodes. We unite co-referring nodes (e.g., marked with \texttt{<P1>}), and merge their connectivity. The resulting SPG is semantically equivalent to the original AMR but uses unlabeled edges for spectral compatibility, and explicits re-entrancies and coreferences. 

% Processing edge-labeled graphs, such as AMRs, pose a challenge for standard spectral graph theory \pcref{sec:bg.graphs}. To bridge this gap, we introduce a semantically-preserving transformation that converts AMRs into equivalent directed graphs without edge labels. 

% Given an input AMR graph \(\gA = (\gV_\gA, \gE_\gA, \gR_\gA) \in \mathbb{A}\), we construct a transformed directed graph \(\gG = (\gV_\gG, \gE_\gG) \in \mathbb{G}\), referred to as the \textit{Role-Expanded Graph} (REG). This representation preserves the full semantic content of the original AMR while making its structure more amenable to computing graph-based positional encodings. We define the overall transformation as a function \(\textsc{Preprocess}: \mathbb{A} \to \mathbb{G}\) from AMR graphs to REGs:
% \begin{equation}
%     \gG = \textsc{Preprocess}(\gA),
% \end{equation}
% where \(\gG\) is obtained through a sequence of graph-structuring steps detailed below.

\subsection{Training details}
\label[appendix]{app:training}

\paragraph{Training setup.}
We build on the open-source \texttt{LitGPT}\footnote{\url{https://github.com/Lightning-AI/litgpt}} framework and extend it to incorporate our structure-aware representations. In particular, we (i) generate graph-based positional encodings (\textsc{AmrPE}), (ii) align AMR nodes to token positions via node-aware tokenization, (iii) introduce a lightweight projection layer $f_\theta$ to map these encodings into the LLM’s embedding space, and (iv) add task-specific prompting to support AMR-to-text generation.

\rebuttal{As in standard sequence-to-sequence fine-tuning, we optimize a cross-entropy objective. The pretrained model weights are updated only through LoRA adapters, while the parameters of $f_\theta$ are trained from scratch. All other LLM parameters remain frozen.}

\paragraph{Tokenizer.}
We experimented with adding AMR role labels (e.g., \texttt{:ARG0}, \texttt{:ARG1}, \texttt{:mod}) to the tokenizer and extending the model's vocabulary accordingly. However, we found that the default tokenizer yielded more stable performance, suggesting that extending the vocabulary with role labels did not offer additional benefits. Therefore, we retain the original tokenizer throughout all experiments.

\paragraph{Hardware setup.} 
All models were trained on a single GPU node with 64\,GB of RAM. Models with 2 billion parameters or more were trained on an NVIDIA H100 GPU, while smaller models (< 2B parameters) were trained on an A100 GPU.

\subsubsection{Hyperparameters}
The hyperparameter choice for each model can be found in \cref{tab:hyperparams} and \cref{tab:hyperparams_FT}.

\input{assets/tables/HP_table}
\paragraph{LoRA Hyperparameters.} 
Given the high computational cost of fine-tuning, we adopted a practical manual hyperparameter search strategy focused on LoRA configurations. We used all LoRA layer types (query, key, value, projection, and head) by default, with a rank ($r$) and scaling factor ($\alpha$) chosen from \{4, 8, 16, 32\}. Dropout rates were selected from \{0.05, 0.1, 0.15\}. In cases where overfitting was observed, we first adjusted the dropout rate to improve generalization. If overfitting persisted, we disabled the LoRA head component, which we found to be the least critical for performance in preliminary runs. This strategy allowed us to balance empirical effectiveness with computational feasibility.

\paragraph{Epochs and training time.} 
All models were trained for 10 epochs with checkpoints saved at the end of each epoch, and the one with the best validation BLEU was chosen (the number of epochs reported is the one with the best BLEU). Training time varied from 9 hours for the smallest models to 16 hours for the larger ones.

\paragraph{Leaning rate.} We use a learning rate schedule with linear warmup for the first 100 optimizer steps, followed by cosine annealing until the end of training. 

\paragraph{Custom hyperparameters.} 
There are five hyperparameters that are specific to our approach: 
\begin{itemize}
    \item \textbf{Number of eigenvectors} (\(k\)): the number of eigenvectors used as positional encodings; we select the \textit{k} eigenvectors corresponding to the smallest \textit{k} eigenvalues. We found that the performance is most stable in the range of 20 to 40 eigenvectors and therefore we chose from \{20,25,30,35,40\}.
    \item \textbf{MLP learning rate multiplier} (\(\mu\)): to improve training stability, we scale the learning rate of the MLP projecting positional encodings by a constant factor \(\mu\), applied on top of the scheduled learning rate; that is, \(\text{LR}_{f_\theta}(t) = \mu \cdot\text{LR}(t)\), where \(\text{LR}(t)\) is the base learning rate at step \(t\).
    \item \textbf{Magnetic parameter} (\(q\)): controls the strength of the complex rotation in the magnetic Laplacian, modulating the influence of edge directionality. After experimenting with values between \(10^{-3}\) and \(0.5\), we found \(q = 0.25\) yielded the most stable results and fixed it for most experiments.
    \item \textbf{Sinusoidal PE frequency base} (\(q_{\text{sin}}\)): the base used in the frequency scaling of sinusoidal positional encodings, analogous to that in Transformer models. Since inter-node sequences are relatively short in our setting, we use \(q_{\text{sin}} = 1000\).
    \item \textbf{Sinusoidal PE dimension} (\(d\)): defines the number of features in the sinusoidal positional encodings concatenated with the eigenvector-based encodings. We set this to 8.
\end{itemize}

\paragraph{Models.} % FT vs SAFT
We used Low-Rank Adaptation (LoRA) \citep{hu2022lora} to fine-tune the following pretrained LLMs: LLaMA 3.2 (3B and 1B) \citep{touvron2023llama}, Qwen 2.5 (3B, 1.5B, and 0.5B) \citep{bai2023qwen}, Gemma 2B \citep {team2024gemma}.
We also attempted to fine-tune Gemma 7B, but encountered frequent out-of-memory (OOM) issues when training on longer AMR sequences, which limited its usability in our experiments.
For each model, we compare two variants: one fine-tuned with our positional encodings (PEs) integrated during training, and one without. For both variants, we report results using the best-performing checkpoint found during development. During evaluation, the PEs are activated consistently based on the corresponding training configuration.

\subsubsection{Prompting Format}
\label[appendix]{app:prompts}

\paragraph{Fine-tuning prompt.}
To enable AMR-to-text generation with large language models, we adopt a structured prompting format implemented via the \texttt{AMR2Text} prompt style. Each prompt consists of three components:

\begin{itemize}
  \item A \textbf{starting token}, which includes task metadata and generation instructions:
  \begin{quote}
  \small{
      \texttt{<AMR-to-Text>}\\
      \texttt{[Task: AMR-to-Text]}\\
      \texttt{[Instruction] Convert the following AMR into natural language text.}\\
      \texttt{[Input: AMR]}
  }
  \end{quote}

  \item The \textbf{linearized AMR graph} \(\gL_\gA\), inserted directly after the input header. This is a token sequence derived from the input AMR graph \(\gA\) (see \cref{sec:method.amr_transformation}).

  \item An \textbf{ending token}, marking the beginning of the generation segment:
  \begin{quote}
  \texttt{[Output: Text]}
  \end{quote}
\end{itemize}

The full prompt passed to the model is thus structured as:
\begin{quote}
  \small{
    \texttt{<AMR-to-Text>}\\
    \texttt{[Task: AMR-to-Text]}\\
    \texttt{[Instruction] Convert the following AMR into natural language text.}\\
    \texttt{[Input: AMR]}\\
    \(\gL_\gA\)\\
    \texttt{[Output: Text]}
  }
\end{quote}

\rebuttal{\paragraph{Few-shot prompt for GPT models.}
The full prompt passed to the model is structured as:
\begin{quote}
\small{
\texttt{You are an assistant that converts AMRs to fluent English sentences.}\\
\texttt{Your job: For each item, convert its `amr` into one fluent English sentence.}\\
\texttt{Do not include explanations; only output JSON.}\\[2mm]
\texttt{Output format (STRICT):}\\
\texttt{\{}\\
\texttt{~~"predictions": [ \{ "id": "...", "text": "..." \} ]}\\
\texttt{\}}\\[2mm]
\texttt{\#\#\# FEW-SHOT EXAMPLES}\\[1mm]
\texttt{Example 1 (input): [{"id":"1","amr":"(u/understand-01 :ARG0 (i/i)}\\
\texttt{:ARG1 (t/thing :ARG1-of (s/say-01 :ARG0 (p/person :name}\\
\texttt{(n/name :op1 "Ron" :op2 "Paul"))))"}]}\\
\texttt{Example 1 (output): \{"predictions":[\{"id":"1","text":"I get what Ron Paul is saying."\}]\}}\\[1mm]
\texttt{Example 2 (input): [{"id":"2","amr":"(s/say-01 :ARG0 (l/libertarian}\\
\texttt{:mod (h/hardcore)) :ARG1 (t/that))"}]}\\
\texttt{Example 2 (output): \{"predictions":[\{"id":"2","text":"Thats what a Hardcore}\\
\texttt{Libertarian would say."\}]\}}\\[1mm]
\texttt{Example 3 (input): [{"id":"3","amr":"(h/hard-02 :degree (k/kind-of) :ARG1}\\
\texttt{(i/import-01 :ARG1 (f/food)) :ARG1-of (c/cause-01 :ARG0 (c2/consider-01}\\
\texttt{:ARG1 (p/probable :domain (w/wipe-out-02 :ARG1 (s/store) :mod (t/too))))))"}]}\\
\texttt{Example 3 (output): \{"predictions":[\{"id":"3","text":"Kind of hard to import food}\\
\texttt{considering that the stores are probably wiped out too."\}]\}}\\[1mm]
\texttt{...}\\[2mm]
\texttt{\#\#\# NOW PROCESS THE REAL BATCH}\\
\texttt{[ \{"id":"0","amr":"(d/date-entity :day 21 :month 8 :year 2007)"\} ]}\\
}
\end{quote}%
}

%% file: figures/amr_to_spg_steps.tex
\begin{figure}[!ht]
    \centering
    \begin{subfigure}[b]{\textwidth}
        \centering
        \input{assets/tikz/amr_graph}
        % \vspace{-1.5em}
        \caption{Input AMR graph \(\gA\)}
        \label{fig:amr_to_spg_steps.input}
    \end{subfigure}
    \vspace{1em}
    \begin{subfigure}[b]{\textwidth}
        \centering
        \vspace{1.4em}
        \input{assets/tikz/amr_bfs_ordering}
        \vspace*{1.3em}
        \caption{\(\gL_\gA = (\ell_1, \dots, \ell_L) = \textsc{BFS}(\gA)\), \(\sigma_\gA: \mathbb{Z} \rightarrowtail \gV_\gG\)}
        \label{fig:amr_to_spg_steps.linearization}
    \end{subfigure}
    \vspace{1em}
    \begin{subfigure}[b]{\textwidth}
        \centering
        \vspace{1em}
        \input{assets/tikz/amr_preprocessing_substructures}
        \caption{\textsc{ToSubgraph}}
        \label{fig:amr_to_spg_steps.substructures}
    \end{subfigure}
    \vspace{1em}
    \begin{subfigure}[b]{\textwidth}
        \centering
        \input{assets/tikz/amr_preprocessing_roleexpand_stop_ordering}
        \caption{\textsc{RoleExpand}, \textsc{AddStopNodes}, and \(\sigma_\gA^{-1}\)}
        \label{fig:amr_to_spg_steps.role_stop_order}
    \end{subfigure}
    \vspace{1em}
    \begin{subfigure}[b]{\textwidth}
        \centering
        \input{assets/tikz/amr_preprocessing_merging}
        \caption{\textsc{Merge}}
        \label{fig:amr_to_spg_steps.merge}
    \end{subfigure}
    \caption{Transformation steps from an AMR graph \(\gA\) to the Role-Expanded Graph \(\gG\).}
    \label{fig:amr_to_spg_steps}
\end{figure}

%% file: assets/tikz/amr_bfs_ordering.tex
\begin{tikzpicture}[baseline, every node/.style={font=\ttfamily\scriptsize, inner sep=0pt}]
    \matrix[column sep=1mm, row sep=0mm] {
        \node {\highlightnodetikz{\pointer{0} \concept{want-01}}{0}}; &
        \node {\highlightnodetikz{\role{:ARG0}}{1}}; &
        \node {\highlightnodetikz{\pointer{1} \concept{child}}{2}}; &
        \node {\highlightnodetikz{\role{:ARG1}}{3}}; &
        \node {\highlightnodetikz{\pointer{2} \concept{believe-01}}{4}}; &
        \node {\highlightnodetikz{\stopwordamr{<stop>}}{5}}; &
        \node {\highlightnodetikz{\pointer{2}}{6}}; &
        \node {\highlightnodetikz{\role{:ARG0}}{7}}; &
        \node {\highlightnodetikz{\pointer{3} \concept{parent}}{8}}; &
        \node {\highlightnodetikz{\role{:ARG1}}{9}}; &
        \node {\highlightnodetikz{\pointer{1}}{10}}; &
        \node {\highlightnodetikz{\stopwordamr{<stop>}}{11}}; \\
    };
\end{tikzpicture}%

%% file: assets/tikz/amr_preprocessing_substructures.tex
\begin{tikzpicture}[
    ->, 
    >=Stealth, 
    font=\scriptsize, 
    node distance=1cm and 1.5cm,
    every node/.style={font=\scriptsize},
    concept/.style={
        % draw, 
        shape=rectangle,
        minimum height=3.5mm,
        minimum width=6mm, 
        align=center,
        inner sep=0pt
    }
  ]
  % Top graph
  \node[concept] (w) {\pointer{0} \textbf{want-01}};
  \node[concept, below=0.4 of w, xshift=-2cm] (c1) {\pointer{1} \textbf{child}};
  \node[concept, below=0.4 of w, xshift=2cm] (b1) {\pointer{2} \textbf{believe-01}};

  \draw[->] (w) -- (b1) node[midway, fill=white, inner sep=1pt] {\role{:ARG1}};
  \draw[->] (w) -- (c1) node[midway, fill=white, inner sep=1pt] {\role{:ARG0}};
  
  % Anchor point shifted to the right of w
  \coordinate[right=5.5cm of w] (anchor);

  % Bottom graph (relative to anchor, shifted right)
  \node[concept, at=(anchor)] (b2) {\pointer{2} \textbf{believe-01}};
  \node[concept, below=0.4 of b2, xshift=2cm] (c2) {\pointer{1} \textbf{child}};
  \node[concept, below=0.4 of b2, xshift=-2cm] (p)  {\pointer{3} \textbf{parent}};

  \draw[->] (b2) -- (c2) node[midway, fill=white, inner sep=1pt] {\role{:ARG1}};
  \draw[->] (b2) -- (p)  node[midway, fill=white, inner sep=1pt] {\role{:ARG0}};
\end{tikzpicture}

%% file: assets/tikz/amr_preprocessing_roleexpand_stop_ordering.tex
\begin{tikzpicture}[
    ->, 
    >=Stealth, 
    font=\scriptsize, 
    node distance=1cm and 1.5cm,
    every node/.style={font=\scriptsize},
    concept/.style={
        % draw, 
        shape=rectangle,
        minimum height=3.5mm,
        minimum width=6mm, 
        align=center,
        inner sep=2pt
    },
    role/.style={
        % draw,
        shape=rectangle,
        minimum height=3.5mm,
        minimum width=6mm,
        align=center,
        inner sep=2pt,
        % fill=xkcdPrussianBlue!15
    },
    stop/.style={
        % draw,
        shape=rectangle,
        minimum height=3.5mm,
        minimum width=6mm,
        inner sep=2pt,
        align=center,
        % fill=xkcdOrange!15
    }
  ]
    % Top graph
    \node[concept] (w) {\textcolor{xkcdPinkyRed}{0}, \pointer{0} \textbf{want-01}};
    \node[concept, below=0.7 of w, xshift=2cm] (b1) {\textcolor{xkcdPinkyRed}{4}, \pointer{2} \textbf{believe-01}};
    \node[concept, below=0.7 of w, xshift=-2cm] (c1) {\textcolor{xkcdPinkyRed}{2}, \pointer{1} \textbf{child}};
    
    % Intermediate nodes
    \node[role, below=0.2 of w, xshift=1cm] (r1) {\textcolor{xkcdPinkyRed}{3}, \role{:ARG1}};
    \node[role, below=0.2 of w, xshift=-1cm]  (r2) {\textcolor{xkcdPinkyRed}{1}, \role{:ARG0}};
    
    % Stop nodes for top graph
    \node[stop, right=0.6 of w, yshift=-0.4cm] (s1) {\textcolor{xkcdPinkyRed}{5}, \stopwordamr{<stop>}};
    
    % Edges
    \draw[->] (w) -- (r1);
    \draw[->] (r1) -- (b1);
    \draw[->] (w) -- (r2);
    \draw[->] (r2) -- (c1);
    \draw[->] (w) -- (s1);
    
    % Anchor point shifted to the right of w
    \coordinate[right=5.5cm of w] (anchor);
    
    % Bottom graph
    \node[concept, at=(anchor)] (b2) {\textcolor{xkcdPinkyRed}{6}, \pointer{2} \textbf{believe-01}};
    \node[concept, below=0.7 of b2, xshift=2cm] (c2) {\textcolor{xkcdPinkyRed}{10}, \pointer{1} \textbf{child}};
    \node[concept, below=0.7 of b2, xshift=-2cm] (p)  {\textcolor{xkcdPinkyRed}{8}, \pointer{3} \textbf{parent}};
    
    % Intermediate nodes
    \node[role, below=0.2 of b2, xshift=1cm] (r3) {\textcolor{xkcdPinkyRed}{9}, \role{:ARG1}};
    \node[role, below=0.2 of b2, xshift=-1cm]  (r4) {\textcolor{xkcdPinkyRed}{7}, \role{:ARG0}};
    
    % Stop nodes for bottom graph
    \node[stop, right=0.6 of b2, yshift=-0.4cm] (s2) {\textcolor{xkcdPinkyRed}{11}, \stopwordamr{<stop>}};
    
    % Edges
    \draw[->] (b2) -- (r3);
    \draw[->] (r3) -- (c2);
    \draw[->] (b2) -- (r4);
    \draw[->] (r4) -- (p);
    \draw[->] (b2) -- (s2);
\end{tikzpicture}

%% file: assets/tikz/amr_preprocessing_merging.tex
\begin{tikzpicture}[
    ->, 
    >=Stealth, 
    font=\scriptsize, 
    node distance=1cm and 1.5cm,
    every node/.style={font=\scriptsize},
    concept/.style={
        % draw, 
        shape=rectangle,
        minimum height=3.5mm,
        minimum width=6mm, 
        align=center,
        inner sep=2pt
    },
    role/.style={
        % draw,
        shape=rectangle,
        minimum height=3.5mm,
        minimum width=6mm,
        align=center,
        inner sep=2pt,
        % fill=xkcdPrussianBlue!15
    },
    stop/.style={
        % draw,
        shape=rectangle,
        minimum height=3.5mm,
        minimum width=6mm,
        inner sep=2pt,
        align=center,
        % fill=xkcdOrange!15
    }
  ]
    % Nodes
    \node[concept] (0) {\textcolor{xkcdPinkyRed}{0}, \pointer{0} \textbf{want-01}};
    \node[role, below=0.3 of 0] (3) {\textcolor{xkcdPinkyRed}{3}, \role{:ARG1}};
    \node[role, left=1.2 of 3]  (1) {\textcolor{xkcdPinkyRed}{1}, \role{:ARG0}};
    \node[stop, right=1.2 of 3] (5) {\textcolor{xkcdPinkyRed}{5}, \stopwordamr{<stop>}};
    
    \node[concept, fill=xkcdApricot!20, below=2.2 of 1, xshift=-1.3cm] (2) {\textcolor{xkcdPinkyRed}{2}, \pointer{1} \textbf{child}};
    \node[concept, fill=xkcdApricot!20, below=2.2 of 1, xshift=1.3cm] (10) {\textcolor{xkcdPinkyRed}{10}, \pointer{1} \textbf{child}};
    
    \node[concept, fill=xkcdDuskyPink!15, below=0.3 of 3] (4) {\textcolor{xkcdPinkyRed}{4}, \pointer{2} \textbf{believe-01}};
    \node[concept, fill=xkcdDuskyPink!15, right=1 of 4] (6) {\textcolor{xkcdPinkyRed}{6}, \pointer{2} \textbf{believe-01}};
    
    \node[role, below=0.3 of 4] (9) {\textcolor{xkcdPinkyRed}{9}, \role{:ARG1}};
    \node[role, right=0.4 of 9]  (7) {\textcolor{xkcdPinkyRed}{7}, \role{:ARG0}};
    \node[stop, below=0.3 of 6] (11) {\textcolor{xkcdPinkyRed}{11}, \stopwordamr{<stop>}};
    
    \node[concept, below=0.3 of 7] (8)  {\textcolor{xkcdPinkyRed}{8}, \pointer{3} \textbf{parent}};

    % Edges
    \draw[->] (0) -- (1);
    \draw[->] (0) -- (3);
    \draw[->] (0) -- (5);
    \draw[->] (1) -- (2);
    \draw[->] (1) -- (10);
    \draw[->] (3) -- (4);
    \draw[->] (3) -- (6);
    \draw[->] (4) -- (7);
    \draw[->] (4) -- (9);
    \draw[->] (4) -- (11);
    \draw[->] (6) -- (7);
    \draw[->] (6) -- (9);
    \draw[->] (6) -- (11);
    \draw[->] (7) -- (8);
    \draw[->] (9) -- (2);
    \draw[->] (9) -- (10);

    % Legend (one line, vertically aligned with proper spacing between groups)
    \matrix[below=1.4cm of 9.center, column sep=0mm] {
        \node[concept, fill=xkcdApricot!20, minimum size=4mm, inner sep=0pt, anchor=center] {\pointer{1}}; &
        \node[anchor=center, align=left] {Co-referring \pointer{1} nodes}; &
        % spacer columns
        \node[anchor=center] {}; &
        \node[anchor=center] {}; &
        \node[concept, fill=xkcdDuskyPink!15, minimum size=4mm, inner sep=0pt, anchor=center] {\pointer{2}}; &
        \node[anchor=center, align=left] {Co-referring \pointer{2} nodes}; \\
    };

\end{tikzpicture}

%% file: assets/algorithms/amr2text_full.tex
\begin{algorithm}[!ht]
    \small
    \captionsetup{font=small}
    \caption{AMR-to-Text Generation with \modelname{}}
    \label{alg:amr2text_streamlined}
    \begin{algorithmic}[1]
        \Require AMR graph \( \gA \), pretrained LLM \( \pi_\theta = \textsc{Decode} \circ \textsc{Embed} \)
        \Ensure Generated text sequence \( \gS \)

        \State \( \gL_\gA, \sigma_\gA = \textsc{BFS}(\gA) \) \Comment{Linearize AMR and align labels}
        \State \( \gG_\gA = \tau(\gL_\gA, \sigma_\gA) \) \Comment{Transform to semantic-preserving graph (SPG)}
        \State \( \bm{\Gamma} = \textsc{MagLapEVD}(\gG_\gA, k) \) \Comment{Compute magnetic Laplacian eigvecs}
        % \For{each \( (\ell_i, i) \in \gL_\gA \)}
        % \For{\( i = 1, \ell_i \in \gL_\gA \)}
        \For{each \( (i, \ell_i) \in \text{enumerate}(\gL_\gA) \)}
            \State \( v_i = \sigma_\gA(i) \)
            \State \( \vt^{(\ell_i)} = \textsc{Tokenize}(\ell_i) \) \Comment{Tokenize label \pcref{eq:node_tokenization}}
            \State \( \phi(v_i) = \bm{\Gamma}_{i,:}^\top \) \Comment{Select complex eigenvector}
            \State \( \textsc{PE}^{(v_i)} = [\Re(\phi(v_i)) \,\, \Im(\phi(v_i))] \) \Comment{Node-level PE \pcref{eq:node_pe}}
            \For{each token \( (j, t^{(\ell_i)}_j) \in \text{enumerate}(\vt^{(\ell_i)}) \)}
                \State \( \textsc{IntraPE}^{(\ell_i)}_j = \textsc{SinPE}(j) \) \Comment{Intra-node PE \pcref{eq:intra_pe}}
                \State \( \textsc{AmrPE}^{(\ell_i)}_j = f_\vartheta (\textsc{PE}^{(v_i)} \,\big\Vert\ \textsc{IntraPE}^{(\ell_i)}_j ) \) \Comment{Token-wise AMR PE \pcref{eq:amr_pe}}
            \EndFor
        \EndFor
        \State \( \smash{\textsc{AmrPE} = ( \textsc{AmrPE}^{(\ell_1)}_1 \, \dots \, \textsc{AmrPE}^{(\ell_L)}_{p_L} )^\top} \)  \Comment{AMR PE \pcref{eq:amr_pe_all}}
        \State \( \gT_{\gL} = \vt^{(\ell_1)} \mathbin{\|} \vt^{(\ell_2)} \mathbin{\|} \dots \mathbin{\|} \vt^{(\ell_L)} \) \Comment{Token sequence \pcref{eq:linearization_tokenization}}
        \State \( \mX = \textsc{Embed}(\gT_{\gL}) \) \Comment{Token sequence embedding}
        \State \( \mH = \mX + \textsc{AmrPE} \) \Comment{Inject structure-aware PE \pcref{eq:integrate_embedding}}
        \State \( \gS = \textsc{Decode}(\mH) \) \Comment{Generate output sequence}
    \end{algorithmic}
\end{algorithm}

%% file: assets/tables/HP_table.tex
\begin{table}[!ht]
\centering
\small
\caption{Hyperparameter configurations for each \modelname{} models}
\label{tab:hyperparams}
\resizebox{\textwidth}{!}{%
    \begin{tabular}{llcccccc}
    \toprule
    \textbf{Category} & \textbf{Hyperparameter} & \textbf{LLaMA 1B} & \textbf{LLaMA 3B} & \textbf{Qwen 0.5B} & \textbf{Qwen 1.5B} & \textbf{Qwen 3B} & \textbf{Gemma 2B} \\
    \midrule
    \multirow{4}{*}{LoRA}
     & Rank ($r$) & \texttt{32} & \texttt{32} & \texttt{32} & \texttt{32} & \texttt{16} & \texttt{32} \\
     & Scaling factor ($\alpha$) & \texttt{32} & \texttt{64} & \texttt{32} & \texttt{64} & \texttt{16} & \texttt{32} \\
     & Dropout & \texttt{0.05} & \texttt{0.05} & \texttt{0.05} & \texttt{0.05} & \texttt{0.05} & \texttt{0.05} \\
     & Head enabled & \texttt{True} & \texttt{True} & \texttt{True} & \texttt{True} & \texttt{True} & \texttt{True} \\
    \midrule
    \multirow{2}{*}{Training}
     & Epochs & \texttt{6} & \texttt{5} & \texttt{6} & \texttt{5} & \texttt{6} & \texttt{5} \\
     & Warmup steps & \texttt{100} & \texttt{100} & \texttt{100} & \texttt{100} & \texttt{100} & \texttt{100} \\
     & Effective batch size & \texttt{256} & \texttt{256} & \texttt{256} & \texttt{256} & \texttt{256} & \texttt{256} \\
    \midrule
    \multirow{5}{*}{Custom}
     & \# Eigenvectors ($k$) & \texttt{30} & \texttt{30} & \texttt{30} & \texttt{30} & \texttt{30} & \texttt{25} \\
     & MLP LR Multiplier ($\mu$) & \texttt{0.8} & \texttt{0.8} & \texttt{0.9} & \texttt{0.8} & \texttt{0.8} & \texttt{0.5} \\
     & Magnetic param ($q$) & \texttt{0.25} & \texttt{0.25} & \texttt{0.25} & \texttt{0.25} & \texttt{0.25} & \texttt{0.25} \\
     & Sinusoidal base ($q_{\text{sin}}$) & \texttt{1000} & \texttt{1000} & \texttt{1000} & \texttt{1000} & \texttt{1000} & \texttt{1000} \\
     & Sinusoidal dim ($d$) & \texttt{8} & \texttt{8} & \texttt{8} & \texttt{8} & \texttt{8} & \texttt{8} \\
    \bottomrule
    \end{tabular}
}
\end{table}

\begin{table}[!ht]
\centering
\small
\caption{Hyperparameter configurations for each conventionally fine-tuned model.}
\label{tab:hyperparams_FT}
\resizebox{\textwidth}{!}{%
\begin{tabular}{llcccccc}
\toprule
\textbf{Category} & \textbf{Hyperparameter} & \textbf{LLaMA 1B} & \textbf{LLaMA 3B} & \textbf{Qwen 0.5B} & \textbf{Qwen 1.5B} & \textbf{Qwen 3B} & \textbf{Gemma 2B} \\
\midrule
\multirow{4}{*}{LoRA}
 & Rank ($r$) & \texttt{16} & \texttt{8} & \texttt{16} & \texttt{32} & \texttt{16} & \texttt{32} \\
 & Scaling factor ($\alpha$) & \texttt{16} & \texttt{8} & \texttt{16} & \texttt{32} & \texttt{16} & \texttt{32} \\
 & Dropout & \texttt{0.05} & \texttt{0.05} & \texttt{0.05} & \texttt{0.05} & \texttt{0.05} & \texttt{0.05} \\
 & Head enabled & \texttt{True} & \texttt{True} & \texttt{True} & \texttt{True} & \texttt{True} & \texttt{True} \\
\midrule
\multirow{3}{*}{Training}
 & Epochs & \texttt{5} & \texttt{5} & \texttt{8} & \texttt{6} & \texttt{8} & \texttt{5} \\
 & Warmup steps & \texttt{100} & \texttt{100} & \texttt{100} & \texttt{100} & \texttt{100} & \texttt{100} \\
 & Effective batch size & \texttt{256} & \texttt{256} & \texttt{256} & \texttt{256} & \texttt{256} & \texttt{256} \\
\bottomrule
\end{tabular}
}
\end{table}

%% file: sections/appendix/C-additional_experiments.tex
\section{Additional Experiments}
\label[appendix]{sec:additional_experiments}

\subsection{Stratified evaluation over number of nodes}

Similarly to the evaluation in \cref{sec:exp.complexity_stratified}, we perform a stratified analysis based on the number of nodes in the original AMR graph $\mathcal{A}$ to examine how both graph size and structural complexity influence the performance of \modelname{}. As shown in \cref{fig:stratified_eval_numnodes}, \modelname{} exhibits consistent improvements over standard fine-tuning across most model sizes, particularly for larger models. However, the trend is less pronounced than in \cref{fig:stratified_eval}, where stratification was based on graph depth. This contrast highlights that the gains from \modelname{} are more strongly associated with structural complexity and long-range dependencies than with graph size alone, suggesting that structural information yields diminishing returns when applied to merely larger—but not necessarily deeper—graphs.
\input{figures/stratified_eval_numnodes_horiz}

\rebuttal{%
    \subsection{Effect of model scale on \modelname{}}%
}%
    \label[appendix]{app:effect_model_size}
\rebuttal{%
    A common hypothesis is that structural inductive biases become less relevant as model scale increases, under the assumption that sufficiently large language models can internalize structural reasoning through parametric capacity alone. Our results do not provide evidence in support of this hypothesis within the range of model sizes we evaluate.%
}% 

\rebuttal{%
    As shown in \cref{fig:effect_model_size}, the relative improvements of \modelname{} over standard fine-tuning (FT) do not exhibit a monotonic decreasing trend with increasing model size. Instead, gains fluctuate across both scale and model family. For example, the largest BLEU improvements are observed for mid-sized LLaMA models, while the strongest ChrF improvements occur for the 3B Qwen model. Moreover, these trends are not consistent across metrics: within a single model family, BLEU and ChrF gains follow different trajectories as scale increases, and model rankings vary depending on the evaluation measure.
}% 

\rebuttal{%
    We emphasize that this does not rule out the possibility that, at larger scales than those considered in this study, the benefits of \modelname{} may eventually diminish or disappear. However, within the regimes we test, no such pattern is observable. Instead, the effect of \modelname{} appears to depend on a combination of model scale, architecture, and generation dynamics, rather than scale alone. Therefore, based on current evidence, the claim that increasing model size renders \modelname{} unnecessary is not supported.%
}

\vspace{2em}

\rebuttal{%
    \subsection{Bootstrap paired significance test}%
}%
    \label[appendix]{app:bootstrap}
    \input{figures/model_size_and_winrate}
\rebuttal{%
    To evaluate whether the performance differences between \modelname{} and standard fine-tuning (FT) are robust to test-set variation, we conduct a bootstrap paired significance analysis. For each model, we repeatedly resample the AMR~3.0 test set with replacement and compute BLEU, and ChrF for both FT and \modelname{} on each bootstrap replicate. This yields paired score distributions without requiring multiple independent training runs, allowing us to isolate test-set variability as a source of uncertainty.
}% 

\rebuttal{%
    \cref{tab:bootstrap} reports the bootstrap mean and standard deviation for each metric and model. Across all settings, the variance induced by resampling is small (typically within \(\pm 0.3\)–\(0.5\) BLEU), indicating that the observed performance differences are not driven by unstable test-set fluctuations. 
}% 

\rebuttal{%
    To further characterize the consistency of these differences, \cref{fig:bootstrap_winrate} presents a win-rate matrix, showing for each model–metric pair the fraction of bootstrap samples in which \modelname{} outperforms FT. In most cases, \modelname{} wins on a clear majority of samples, including all LLaMA variants and two out of the three Qwen models. For settings where the average improvements are small, the bootstrap distributions reflect this appropriately through near-balanced win rates, rather than artificially inflating significance.
}% 

\rebuttal{%
    Overall, this analysis indicates that the gains reported in the main results are stable under reasonable test-set perturbations. At the same time, the bootstrap results remain appropriately conservative in cases where the FT–\modelname{} gap is minimal, avoiding overinterpretation of marginal differences.%
}

\subsection{Extra Evaluation Metrics}
\label[appendix]{app:extra_metrics}
Along with the metrics reported in \cref{sec:exp.main_results}, we evaluate \modelname{} vs. FT using METEOR \citep{lavie2007meteor} and BERTScore \citep{zhang2020bertscoreevaluatingtextgeneration}, two complementary metrics that are more sensitive to semantic adequacy and fluency. The results are shown in \cref{tab:extra_metrics}. We further stratified the gains by AMR graph depth, following the same setup as in \cref{fig:stratified_eval}. Both BERTScore and METEOR gains increase with graph depth as shown in \cref{tab:extra_metrics_strat}. These results support our claim: \modelname{} becomes increasingly beneficial for structurally complex inputs, and these improvements hold across multiple metrics.

\subsection{Runtime}
\label[appendix]{app:runtime}
We provide a formal breakdown of \modelname{}'s time and space complexity and clarify the scope of its computational overhead. All graph-related operations are performed once at preprocessing time and do not affect training or inference time. These include: graph preprocessing, magnetic Laplacian computation, partial eigendecomposition (EVD), and positional encoding computation. The main bottleneck in this process is the partial EVD. 

The dense partial EVD requires $\mathcal{O}(k n^2)$, where $k$ is the number of computed eigenvalues, $n$ the number of nodes. For most practical scenarios in the AMR-to-text generation task, this complexity is manageable, as AMRs are relatively small, with AMR 3.0 (\(\sim \)54 nodes) and DocAMR (\(\sim \)730 nodes).
Asymptotically, sparse solvers become advantageous when $n \gg k$. This translates in practice to $n \gtrapprox 2000$. In this case, the complexity is $\mathcal{O}(k m)$ where $m$ is the number of edges.

\input{assets/tables/bootstrap_50}
\input{assets/tables/extra_metrics}
\input{assets/tables/inference_time}
We first show in \cref{tab:sparse_dense_times} that, in our case, using dense solvers yields faster runtime compared to sparse solvers. Then, we report in \cref{tab:runtime_results} the average and maximum time (in seconds) required for inference and precomputation, showing the negligible impact of the latter. The computation of our structure-aware encodings is a preprocessing step that leaves the model architecture and runtime efficiency intact. Since structure-aware encodings are computed once per graph and reused throughout training and inference, the overall overhead remains minimal.

\rebuttal{At inference time, \modelname{} does not change the computation inside the Transformer itself. The token sequence length and the hidden dimensionality remain the same, so the attention and feed-forward layers have the same cost as in standard fine-tuning (FT). The only additional costs compared to FT come from two sources. First, we compute the graph positional encodings for the input AMR. This is done once per graph and does not depend on the decoding length. As shown in \cref{tab:runtime_results}, this step adds on average 0.01 seconds on AMR 3.0 and 0.28 seconds on DocAMR, which is small compared to the overall inference time. Second, we apply a lightweight two-layer MLP to project these encodings into the model's embedding space. This MLP maps \(\mathbb{R}^{2k+d} \rightarrow \mathbb{R}^{d_{\text{emb}}}\) and \(\mathbb{R}^{d_{\text{emb}}} \rightarrow \mathbb{R}^{d_{\text{emb}}}\), and introduces \(d_{\text{emb}}(2k + d + d_{\text{emb}} + 2)\) additional parameters. This overhead is negligible relative to the size of the underlying LLM.}

\rebuttal{%
    \subsection{Potential applicability beyond AMR}
    \label[appendix]{app:other_domains}
    \modelname{} does not assume properties unique to AMR and only requires a graph structure together with a node-to-token alignment. This suggests a potential extensions to other graph-structured data associated with text, such as semantic role graphs, knowledge graph triplets, or discourse trees mapped to sentences, just to name a few. A full empirical study of these settings is outside the scope of this work, but we view this as a promising direction for future research.%
}

%% file: figures/stratified_eval_numnodes_horiz.tex
\begin{figure}[!t]
    \centering
    \begin{subfigure}[t]{0.48\textwidth}
        \centering
        \includegraphics[width=\linewidth]{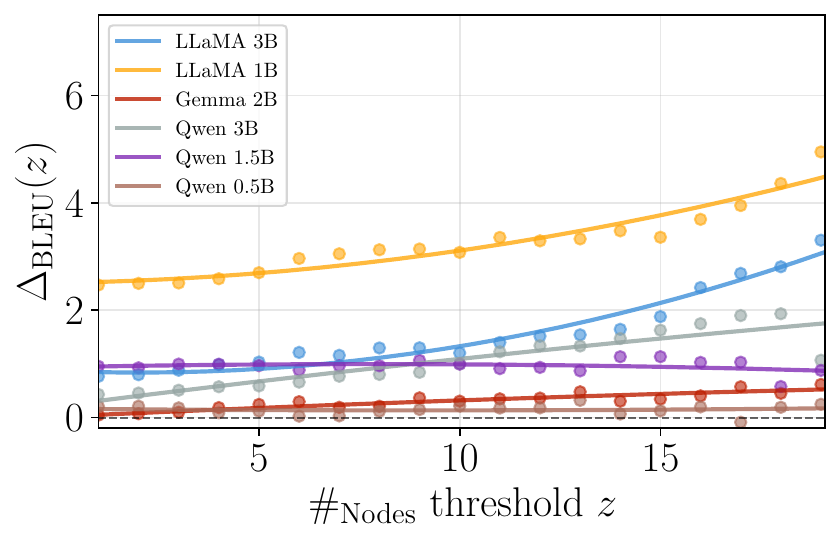}
        \caption{Absolute improvement (\ensuremath{\Delta_{\mathrm{BLEU}}(z)}).}
        \label{fig:stratified_eval_numnodes.absolute}
    \end{subfigure}
    \hfill
    \begin{subfigure}[t]{0.48\textwidth}
        \centering
        \includegraphics[width=\linewidth]{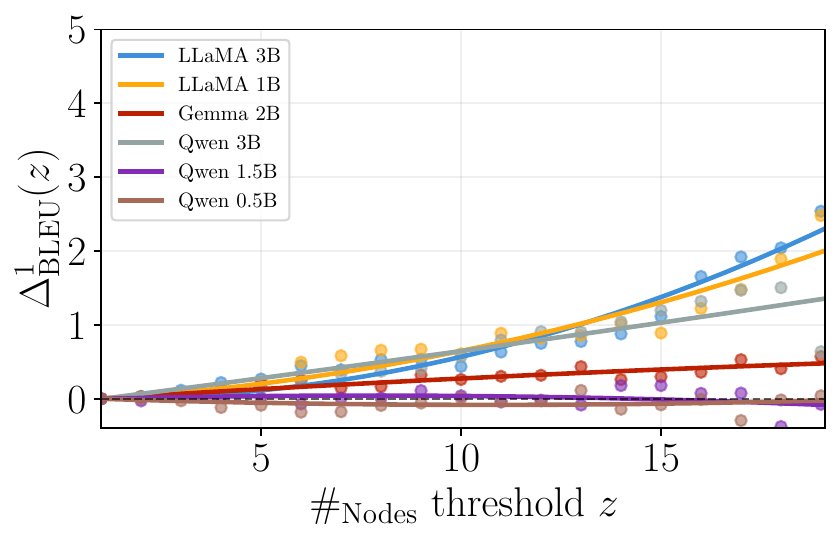}
        \caption{Relative improvement (\ensuremath{\Delta^1_{\mathrm{BLEU}}(z)}).}
        \label{fig:stratified_eval_numnodes.relative}
    \end{subfigure}
    
    \caption{
    BLEU score improvements of structurally-aware fine-tuned (\modelname{}) models over conventionally fine-tuned (FT) counterparts, on AMR instances with number of nodes \ensuremath{\#_\mathrm{Nodes}(\mathcal{A}) \geq z}. 
    (a) \textbf{Absolute improvement} (\ensuremath{\Delta_\text{BLEU}}): differences in BLEU score between \modelname{} and FT models across varying number of nodes and model families.
    (b) \textbf{Relative improvement} (\ensuremath{\Delta^{1}_\text{BLEU}}): differences in BLEU scores normalized by single-node graph performance. The results show consistent gains, though the magnitude of improvement is less pronounced compared to depth-based stratification \pcref{fig:stratified_eval}[see], indicating that structural complexity plays a more critical role than graph size alone. All lines are second-degree polynomial fits.
    }
    \label{fig:stratified_eval_numnodes}
\end{figure}

%% file: figures/model_size_and_winrate.tex
\begin{wrapfigure}[36]{r}{0.3\textwidth}
    \centering
    \vspace{-1.5em}
    \includegraphics[width=\linewidth]{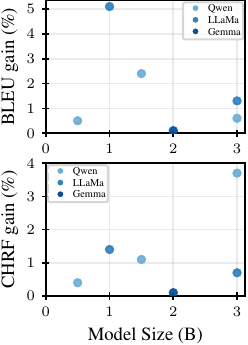}
    \caption{
        \rebuttal{\textbf{Relative BLEU and ChrF++ improvements as a function of model size.}}
    }
    \label{fig:effect_model_size}

    \vspace{1em}

    \includegraphics[width=\linewidth]{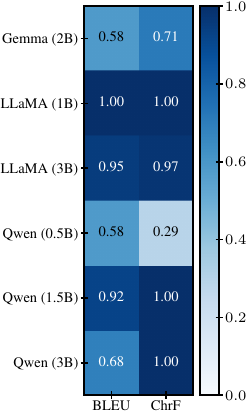}
    \caption{
        \rebuttal{\textbf{Win-rate heatmap of the bootstrap paired test.}}
    }
    \label{fig:bootstrap_winrate}

\end{wrapfigure}

%% file: assets/tables/bootstrap_50.tex
\begin{table}[!t]
    \centering
    \small
    \caption{\rebuttal{\textbf{Bootstrap means and standard deviations of BLEU, and ChrF over 50 resampled test sets.} The variability is small across models, and the relative behavior of SAFT vs.\ FT aligns with the trends reported in the main results. Bold results are statistically significant (\(p < 0.05\)).}}
    \label{tab:bootstrap}
    \begin{tabular}{llcc}
    \toprule
    \textbf{Model} & \textbf{Variant} & \textbf{BLEU} & \textbf{ChrF} \\
    \midrule
    \multirow{2}{*}{Gemma (2B)} & FT & 52.97 ± 0.49 & 69.74 ± 0.29 \\
    & \cellcolor{tumblue!30}SAFT & \cellcolor{tumblue!30}53.01 ± 0.48 & \cellcolor{tumblue!30}69.98 ± 0.30\\
    \midrule
    \multirow{2}{*}{LLaMA 3.2 (1B)} & FT & 45.64 ± 0.55 & 65.74 ± 0.37\\
    & \cellcolor{tumblue!30}SAFT & \cellcolor{tumblue!30}\textbf{47.75 ± 0.54} & \cellcolor{tumblue!30}\textbf{67.65 ± 0.33}\\
    \midrule
    \multirow{2}{*}{LLaMA 3.2 (3B)} & FT & 53.52 ± 0.40 & 71.70 ± 0.25\\
    & \cellcolor{tumblue!30}SAFT & \cellcolor{tumblue!30}54.45 ± 0.44 & \cellcolor{tumblue!30}72.40 ± 0.30 \\
    \midrule
    \multirow{2}{*}{Qwen 2.5 (0.5B)} & FT & 42.68 ± 0.40 & 63.14 ± 0.30\\
    & \cellcolor{tumblue!30}SAFT & \cellcolor{tumblue!30}42.94 ± 0.52 & \cellcolor{tumblue!30}62.94 ± 0.30\\
    \midrule
    \multirow{2}{*}{Qwen 2.5 (1.5B)} & FT & 50.66 ± 0.46 & 68.69 ± 0.30\\
    & \cellcolor{tumblue!30}SAFT & \cellcolor{tumblue!30}51.69 ± 0.54 & \cellcolor{tumblue!30}\textbf{70.70 ± 0.34}\\
    \midrule
    \multirow{2}{*}{Qwen 2.5 (3B)} & FT & 51.75 ± 0.50 & 69.33 ± 0.37\\
    & \cellcolor{tumblue!30}SAFT & \cellcolor{tumblue!30}52.00 ± 0.45 & \cellcolor{tumblue!30}\textbf{70.44 ± 0.29}\\
    \bottomrule
    \end{tabular}
\end{table}

%% file: assets/tables/extra_metrics.tex
\begin{table}[t]
    \centering
    % First table
    \begin{minipage}[t]{0.48\textwidth}
        \small
        \centering
        \caption{
        \textbf{Comparison of FT vs. \modelname{}.} 
        We report METEOR and BERTScore for Qwen 2.5 and LLaMA 3.2 (3B) on AMR 3.0.
        Best results per metric are highlighted in bold, with relative gain in parentheses.
        }
        \label{tab:extra_metrics}
        \resizebox{\linewidth}{!}{%
        \begin{tabular}{llll}
            \toprule
            \textbf{Model} & \textbf{Variant} & \textbf{METEOR} $\uparrow$ & \textbf{BERTScore} $\uparrow$ \\
            \midrule
            \multirow{2}{*}{Qwen 2.5 (3B)} 
            & FT & 59.4 & 82.85 \\
            & \cellcolor{tumblue!30}\modelname{} 
            & \cellcolor{tumblue!30}\textbf{60.1} \scriptsize{(+1.18\%)} 
            & \cellcolor{tumblue!30}\textbf{83.69} \scriptsize{(+1.01\%)} \\
            \midrule
            \multirow{2}{*}{LLaMA 3.2 (3B)} 
            & FT & 70.3 & 86.19 \\
            & \cellcolor{tumblue!30}\modelname{} 
            & \cellcolor{tumblue!30}\textbf{70.5} \scriptsize{(+0.28\%)} 
            & \cellcolor{tumblue!30}\textbf{86.22} \scriptsize{(+0.03\%)} \\
            \bottomrule
        \end{tabular}
        }
    \end{minipage}
    \hfill
    % Second table
    \begin{minipage}[t]{0.48\textwidth}
        \small
        \centering
        \caption{
        \textbf{Impact of graph depth on gains.} 
        We report relative improvements in BERTScore and METEOR across different depths.
        }
        \label{tab:extra_metrics_strat}
        \resizebox{\linewidth}{!}{%
        \begin{tabular}{lcc}
            \toprule
            \textbf{Graph Depth} & \textbf{BERTScore Gain} $\uparrow$ & \textbf{METEOR Gain} $\uparrow$ \\
            \midrule
            0  & 0.84 & 1.08 \\
            3  & 1.30 & 1.31 \\
            9  & 3.11 & 5.01 \\
            10 & 2.96 & 3.27 \\
            \bottomrule
        \end{tabular}
        }
    \end{minipage}
\end{table}

%% file: assets/tables/inference_time.tex
\begin{table}[t]
    \centering
    % First table
    \begin{minipage}[t]{0.49\textwidth}
        \small
        \centering
        \caption{
        \textbf{Sparse vs. dense preprocessing times.} 
        We report average and maximum runtimes (in seconds) for AMR 3.0 and DocAMR.
        }
        \label{tab:sparse_dense_times}
        \resizebox{\linewidth}{!}{%
        \begin{tabular}{l l cc}
            \toprule
            \textbf{Dataset} & \textbf{Method} & \textbf{Avg Time (s)} $\downarrow$ & \textbf{Max Time (s)} $\downarrow$ \\
            \midrule
            \multirow{2}{*}{AMR 3.0} 
            & Sparse & 0.009  & 0.12  \\
            & Dense  & 0.027  & 0.57 \\
            \midrule
            \multirow{2}{*}{DocAMR} 
            & Sparse & 1.31   & 12.51 \\
            & Dense  & 0.28   & 2.49  \\
            \bottomrule
        \end{tabular}
        }
    \end{minipage}
    \hfill
    % Second table
    \begin{minipage}[t]{0.49\textwidth}
        \small
        \centering
        \caption{
        \textbf{Inference and preprocessing times.} 
        Inference time and graph preprocessing time (including EVD, sinusoidal encoding, and projection via the MLP) for Qwen 3B.
        }
        \label{tab:runtime_results}
        \resizebox{\linewidth}{!}{%
        \begin{tabular}{lcc}
            \toprule
            \textbf{Dataset} & \textbf{Inference Time (s)} & \textbf{Preprocessing Time (s)} \\
            \midrule
            AMR 3.0 & Avg: 1.73 \quad Max: 6.81 & Avg: 0.01 \quad Max: 0.12 \\
            DocAMR  & Avg: 325.24 \quad Max: 582.22 & Avg: 0.28 \quad Max: 2.49 \\
            \bottomrule
        \end{tabular}
        }
    \end{minipage}
\end{table}

%% file: sections/appendix/D-limitations.tex
\section{Limitations}
\label[appendix]{app:limitations}

While our approach achieves consistent improvements, particularly on semantically complex graphs, several limitations remain. First, it introduces computational overhead from graph preprocessing and structural encoding, though this can be mitigated through caching. Second, gains are less pronounced on simpler inputs with limited structural information, suggesting the method's inductive bias is not universally beneficial. Third, effectiveness depends on hyperparameter choices such as positional encoding dimensionality, which may require tuning. Finally, extending this method to other graph-structured tasks requires task-specific node-to-token alignments, adding engineering complexity.

%% file: sections/appendix/E-assets_and_licences.tex
\section{Assets and Licences}
\label[appendix]{app:assets_and_licences}

\subsection{Datasets}
\label[appendix]{app:datasets}
We evaluate our approach on the AMR 3.0 dataset (LDC2020T02\footnote{\url{https://catalog.ldc.upenn.edu/LDC2020T02}}) \citep{knight2020amr30}, which consists of approximately 55k training instances, 1.3k for development, and 1.4k for testing. Compared to earlier versions \citep{knight2017amr20}, AMR 3.0 includes more diverse graph structures and broader linguistic coverage, providing a rigorous benchmark for AMR-to-text generation.

This release is a semantically annotated corpus of over 59k English sentences drawn from a diverse mix of domains, including broadcast conversation, discussion forums, weblogs, newswire, and fiction. Annotations cover PropBank-style frames, non-core semantic roles, coreference, named entities, modality, negation, quantities, and questions. Sentence-level annotations are represented as rooted, directed acyclic graphs designed to abstract away from surface syntax and emphasize predicate-argument structure.

For a subset of experiments, we also evaluate on the DocAMR dataset (part of AMR 3.0), which extends AMR to the document level by providing inter-sentence coreference and discourse-level annotations. This enables assessment of long-range semantic dependencies and coherence in multi-sentence generation. \rebuttal{DocAMR consists of 284 documents in the training split and 9 documents in the test split, covering a total of 8027 gold sentence-level AMRs.}
\input{figures/dataset_sizes}

We use the dataset as released by the Linguistic Data Consortium (LDC2020T02), without augmenting with any silver data (i.e., data labeled through heuristic or automated methods). AMR 3.0 is distributed under the LDC User Agreement and is not publicly available; access requires an institutional or individual LDC license. For reference, the release was published on January 15, 2020 and includes contributions from DARPA-funded programs (BOLT, DEFT, MRP, LORELEI) and NSF-supported research.

\rebuttal{To make sample availability across structural complexities explicit, we include histograms of graph depths for AMR~3.0 and DocAMR (\cref{fig:dataset_size}). AMR~3.0 exhibits a long tail with few very high-depth graphs, whereas DocAMR contains nine document-level graphs whose constituent sentence-level AMRs all originate from the AMR~3.0 test split, each covering many AMRs. To test the capabilities of our model across multiple complexities (number of AMRs in the case of DocAMR), we sample subgraphs from document-level AMRs of increasing complexities, with a resulting count distribution as shown in \cref{fig:dataset_size}(b).}

\begin{table}[!ht]
    \centering
    \small
    \caption{Datasets used for AMR-to-text generation.}
    \resizebox{\linewidth}{!}{%
        \begin{tabular}{@{}llll@{}}
        \toprule
        \textbf{Dataset}    & \textbf{Size (Train/Dev/Test)} & \textbf{Key Features} & \textbf{License} \\
        \midrule
        AMR 3.0             & 55k / 1.3k / 1.4k               & Sentence-level graphs, broad linguistic coverage   & LDC Non-Member License \\
        DocAMR              & \rebuttal{284 / - / 9}          & Document-level annotations, coreference, discourse & LDC Non-Member License \\
        \bottomrule
        \end{tabular}
    }
\end{table}

\subsection{Models}
We conduct experiments using a selection of publicly available pretrained language models with open or research-focused licenses. All models are used strictly for academic purposes, in compliance with their respective licenses.

\paragraph{LitGPT.}
We build on the \texttt{LitGPT} framework\footnote{\url{https://github.com/Lightning-AI/litgpt}}, an open-source project released under the Apache License 2.0\footnote{\url{http://www.apache.org/licenses/LICENSE-2.0}}. It provides modular components for efficient fine-tuning, inference, and reproducibility across large-scale models.

\paragraph{Qwen.}
Qwen models\footnote{\url{https://github.com/QwenLM/Qwen}}, developed by Alibaba Cloud, are released under the Apache License 2.0. This permissive open-source license permits modification, distribution, and commercial use, provided appropriate attribution is maintained.

\paragraph{Gemma.}
Gemma\footnote{\url{https://github.com/google-deepmind/gemma}}, developed by Google DeepMind, is also licensed under the Apache License 2.0. This allows for both academic and commercial applications and emphasizes interoperability with a wide range of open-source software.

\paragraph{LLaMA 2.}
LLaMA 2 models\footnote{\url{https://ai.meta.com/resources/models-and-libraries/llama-downloads/}}, released by Meta, are governed by the LLAMA 2 Community License Agreement. The license permits use, modification, and redistribution, but restricts:
\begin{itemize}
    \item Commercial use by entities exceeding 700 million monthly active users without explicit permission from Meta
    \item Use of LLaMA outputs to train competing large language models
\end{itemize}
Redistributions must include a notice file, and use is subject to Meta's Acceptable Use Policy\footnote{\url{https://llama.com/use-policy}}.

\begin{table}[!ht]
    \centering
    % \small
    \caption{Pretrained models and licensing details.}
    \resizebox{\linewidth}{!}{%
    \begin{tabular}{@{}llll@{}}
    \toprule
    \textbf{Model} & \textbf{Provider} & \textbf{License} & \textbf{Notes} \\
    \midrule
    LitGPT   & Lightning AI      & Apache 2.0 & Permissive, for training and inference \\
    Qwen     & Alibaba Cloud     & Apache 2.0 & Open-source, commercial use permitted \\
    Gemma    & Google DeepMind   & Apache 2.0 & Open-source, commercial use permitted \\
    LLaMA 2  & Meta              & LLAMA 2 Community License & Requires license for large-scale commercial use \\
    \bottomrule
    \end{tabular}
    }
\end{table}

%% file: figures/dataset_sizes.tex
\begin{figure}[!t]
    \centering
    \includegraphics[width=\linewidth]{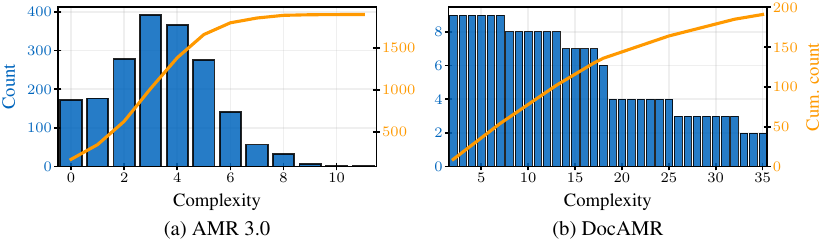}
    \caption{\rebuttal{Count of AMR graphs per complexity on both AMR 3.0 and DocAMR (test sets).}} 
    \label{fig:dataset_size}
\end{figure}

%% file: sections/appendix/F-related_work.tex
\section{Additional information on related work}
\label[appendix]{app:related_work}

Prior work on AMR-to-text generation—and more broadly, text generation from graph-structured data—has been explored through three main paradigms: \textbf{linearization-based approaches}, which serialize graphs into sequences; \textbf{adapter-based approaches}, which introduce graph-native modules into pretrained architectures; and \textbf{graph-tuning-based approaches}, which couple LLMs with trainable graph encoders or embeddings. To the best of our knowledge, no prior work has investigated graph positional encodings as a lightweight, architecture-agnostic means of enabling structure-aware fine-tuning.

\subsection{Linearization-based approaches}

These methods convert the input graph into a linear sequence and fine-tune a pre-trained encoder-decoder transformer (e.g., BART, T5) in a standard seq-to-seq setup.

\citet{bevilacqua2021spring} introduced a symmetric framework for AMR parsing and generation by fine-tuning BART on linearized AMR graphs using both DFS and BFS traversals (SPRING). AMR-BART \citep{bai2022graph} builds on SPRING by incorporating self-supervised graph denoising tasks during pretraining, which improves robustness to structural noise. BiBL \citep{cheng2022bibl} further extends this line of work by jointly modeling AMR-to-text and text-to-AMR transitions through single-stage multitask learning with auxiliary losses.
These models share a common foundation: they linearize the AMR graph and fine-tune a standard transformer. This strategy has also been applied to large language models (LLMs) via fine-tuning \citep{raut2025can, mager2020gpt} or prompting \citep{yao2024semantic, jin2024analyzing} using the linearized AMR graph as input.

More generally, the practice of aligning LLMs with structured data through linearization has found success across domains such as molecular generation \citep{zheng2024llmsdrug}, network traffic analysis \citep{cui2025trafficllm}, tabular reasoning \citep{fang2024llmstabular}, and 3D mesh processing \citep{wang2024llamamesh}.

\subsection{Adapter-based approaches}

Adapter-based methods directly model the structure of the input graph using graph neural networks (GNNs) or related components, which are then integrated into transformer architectures.

StructAdapt \citep{ribeiro2021structadapt} \rebuttal{is built around the encoder–decoder architecture (e.g., T5) in which a dedicated graph encoder (which they call StructAdapt) produces structure-enriched representations that the decoder then consumes via cross-attention and MLP adapters. Its core design relies on this asymmetry: graph processing happens entirely in the encoder, while the decoder only receives those enriched encoder states}.
Other methods take a similar direction by modifying the attention mechanism to incorporate structural biases from the input graph \citep{zhu2019modeling}. Another line of work avoids transformer pretraining altogether, instead training graph-to-sequence models from scratch that can natively process graph inputs \citep{song2018graph2seq, wang2020amr}.

\subsection{LLMs for graph-structured data}

The rise of large language models (LLMs) \citep{vaswani2017attention, devlin2019bert, brown2020language, touvron2023llama} has reshaped NLP. Recently, there has been growing interest in extending LLMs to handle graph-structured inputs \citep{jin2024llmgraphsurvey}, particularly in domains like molecules, knowledge graphs, and social networks. Existing methods typically fall into one of three strategies: (i) flattening graphs into linear sequences \citep{jiang2023structgpt, fatemi2024talk, yao2024exploring}; (ii) modifying the LLM architecture to incorporate graph encoders \citep{zhang2022greaselm}; or (iii) generating structure-aware token embeddings that align with LLM representations \citep{tian2024graph, tang2024graphgpt}. The latter direction shows promise but introduces additional training complexity due to the need for separate graph encoders and alignment mechanisms.

\subsection{Graph-tuning-based approaches}

A more recent line of work integrates graph modules directly into LLMs, aiming to couple structural encoding with pretrained language models. 
GraphAdapter \citep{huang2024graphadapter} employs a trainable GNN as an adapter, aligned with the LLM during fine-tuning, requiring both pretraining and parameter-intensive alignment. 
GraphLLM \citep{chai2023graphllm} inserts graph transformers at every layer of the LLM by introducing learned graph-based prefix tokens into the key/value projections. 
GraphGPT \citep{tang2024graphgpt} prepends graph embeddings produced by a trainable GNN into the prompt via a learned projector, keeping the LLM frozen. 
While these methods inject structural information effectively, they introduce substantial complexity in the form of additional trainable modules, architectural modifications, or expensive pretraining requirements.

\subsection{Comparison and positioning of \modelname{}}

Across these paradigms, a common trade-off emerges: linearization-based methods retain compatibility with standard architectures but discard explicit structure; adapter-based methods preserve structure but sacrifice full LLM compatibility; graph-tuning-based methods tightly integrate graphs with LLMs but at the cost of additional parameters and architectural changes. 
In contrast, \modelname{} is architecture-agnostic and parameter-efficient. It introduces relational inductive bias through precomputed, parameter-free positional encodings derived from the magnetic Laplacian spectrum. These encodings are aligned with the LLM’s embedding space via a lightweight projection layer, requiring no graph-specific training, no architectural changes, and no costly pretraining. 
This design makes \modelname{} a simple and efficient way to enable structure-aware fine-tuning of LLMs, while remaining general to tasks involving graph-structured inputs.

%% file: sections/appendix/G-examples.tex
\rebuttal{%
\section{Generated outputs and error analysis}
We present a qualitative error analysis comparing baseline fine-tuning (FT) and our structure-aware fine-tuning method (SAFT). To avoid cherry-picking and to ensure that examples are sampled in a principled way, we compute smoothed sentence-level BLEU scores for both models on every test instance. We then partition the dataset into performance buckets based on percentile thresholds: cases where SAFT is considerably better than FT, cases where FT is considerably better than SAFT, and cases where both models either perform well or fail. From each bucket, we draw a small random sample of five examples. For each selected instance, we report the reference output, the two model predictions, and their respective BLEU scores. We additionally highlight token-level differences using a color-coded character diff to make divergences visually salient. The goal is not to claim statistical significance from individual examples, but to give the reader concrete insight into the characteristic strengths and failure modes of each system across distinct performance regimes.
\input{sections/appendix/G1-sentences}
}
\newpage

%adding the figures to be viewed at the end of the appendix
\input{figures/preprocessing_examples/amr_example1}
\input{figures/preprocessing_examples/preprocessing_example1}
\input{figures/preprocessing_examples/amr_example2}
\input{figures/preprocessing_examples/preprocessing_example2}

\input{figures/preprocessing_examples/amr_example3}
\input{figures/preprocessing_examples/preprocessing_example3}

\input{figures/preprocessing_examples/amr_example4}
\input{figures/preprocessing_examples/preprocessing_example4}

%% file: sections/appendix/G1-sentences.tex
\subsection*{SAFT better}

\textbf{Ground Truth:} International; weapons; proliferation; Government; energy\\\textbf{FT Prediction:} International; weapons; proliferation; \textcolor{red}{G}\textcolor{red}{o}\textcolor{red}{v}\textcolor{red}{e}\textcolor{red}{r}\textcolor{red}{n}\textcolor{red}{m}en\textcolor{red}{t}\textcolor{green}{e}\textcolor{green}{r}\textcolor{green}{g}\textcolor{green}{y}; energy [BLEU: 28.6]\\\textbf{SAFT Prediction:} International; weapons; proliferation; Government; energy [BLEU: 100.0]\\[1mm]
\vspace{1em}\hrule\vspace{1em}

\textbf{Ground Truth:} International; crime; Government; narcotics\\\textbf{FT Prediction:} International; crime; \textcolor{green}{g}\textcolor{red}{G}overnment; narcotics [BLEU: 18.8]\\\textbf{SAFT Prediction:} International; crime; Government; narcotics [BLEU: 100.0]\\[1mm]
\vspace{1em}\hrule\vspace{1em}

\textbf{Ground Truth:} The issues have been unresolved for 4 years.\\\textbf{FT Prediction:} The issue\textcolor{red}{s} \textcolor{green}{w}\textcolor{green}{e}\textcolor{green}{r}\textcolor{red}{h}\textcolor{red}{a}\textcolor{red}{v}e \textcolor{red}{b}\textcolor{red}{e}\textcolor{red}{e}n\textcolor{green}{o}\textcolor{green}{t} unresolved for \textcolor{green}{f}\textcolor{green}{o}\textcolor{green}{u}\textcolor{green}{r}4 years. [BLEU: 7.0]\\\textbf{SAFT Prediction:} The issue\textcolor{red}{s} have been unresolved for 4 years. [BLEU: 70.7]\\[1mm]
\vspace{1em}\hrule\vspace{1em}

\textbf{Ground Truth:} You can't get her sectioned for that.\\\textbf{FT Prediction:} You can't get \textcolor{green}{a}\textcolor{red}{h}\textcolor{red}{e}\textcolor{red}{r} section\textcolor{red}{e}\textcolor{red}{d} \textcolor{red}{f}\textcolor{red}{o}\textcolor{red}{r}\textcolor{green}{b}\textcolor{green}{e}\textcolor{green}{c}\textcolor{green}{a}\textcolor{green}{u}\textcolor{green}{s}\textcolor{green}{e}\textcolor{green}{ }\textcolor{green}{b}\textcolor{green}{e}\textcolor{green}{c}\textcolor{green}{a}\textcolor{green}{u}\textcolor{green}{s}\textcolor{green}{e} that. [BLEU: 14.8]\\\textbf{SAFT Prediction:} You can't get her sectioned for that. [BLEU: 100.0]\\[1mm]
\vspace{1em}\hrule\vspace{1em}

\textbf{Ground Truth:} 2. Create a few nuclear-powered aircraft carrier battle groups.\\\textbf{FT Prediction:} 2. Creat\textcolor{red}{e}\textcolor{green}{i}\textcolor{green}{n}\textcolor{green}{g} a few nuclear\textcolor{green}{ }\textcolor{red}{-}powered aircraft carrier\textcolor{green}{s} \textcolor{green}{g}\textcolor{green}{r}\textcolor{green}{o}\textcolor{green}{u}\textcolor{green}{p}\textcolor{green}{s}\textcolor{red}{b}\textcolor{red}{a}\textcolor{red}{t}\textcolor{red}{t}\textcolor{red}{l}\textcolor{red}{e} groups. [BLEU: 5.6]\\\textbf{SAFT Prediction:} 2. Create a few \textcolor{red}{n}\textcolor{red}{u}\textcolor{red}{c}\textcolor{green}{b}\textcolor{green}{a}\textcolor{green}{t}\textcolor{green}{t}le\textcolor{red}{a}\textcolor{red}{r}-powered aircraft carrier battle groups. [BLEU: 59.7]\\[1mm]
\vspace{1em}\hrule\vspace{1em}

\subsection*{FT better}

\textbf{Ground Truth:} Ukraine does not supply or have plans to supply any armaments to the Government of South Sudan.\\\textbf{FT Prediction:} Ukraine \textcolor{green}{h}\textcolor{green}{a}\textcolor{red}{d}\textcolor{red}{o}\textcolor{red}{e}s not supply \textcolor{red}{o}\textcolor{red}{r}\textcolor{green}{a}\textcolor{green}{n}\textcolor{green}{y} have plans to supply any arm\textcolor{green}{s}aments to the \textcolor{green}{S}\textcolor{red}{G}o\textcolor{green}{u}\textcolor{red}{v}\textcolor{red}{e}\textcolor{red}{r}\textcolor{red}{n}\textcolor{red}{m}\textcolor{red}{e}\textcolor{red}{n}t\textcolor{green}{h} of South Sudan. [BLEU: 34.7]\\\textbf{SAFT Prediction:} Ukraine \textcolor{green}{h}\textcolor{green}{a}\textcolor{red}{d}\textcolor{red}{o}\textcolor{red}{e}s not supply \textcolor{red}{o}\textcolor{red}{r}\textcolor{green}{a}\textcolor{green}{n}\textcolor{green}{y} have \textcolor{red}{p}\textcolor{red}{l}an\textcolor{green}{y}\textcolor{red}{s} to supply any arm\textcolor{green}{s}aments to the \textcolor{green}{S}\textcolor{red}{G}o\textcolor{green}{u}\textcolor{red}{v}\textcolor{red}{e}\textcolor{red}{r}\textcolor{red}{n}\textcolor{red}{m}\textcolor{red}{e}\textcolor{red}{n}t\textcolor{green}{h} of South Sudan. [BLEU: 12.6]\\[1mm]
\vspace{1em}\hrule\vspace{1em}

\textbf{Ground Truth:} The proposal may complicate the Bush administration's efforts to win an exemption for India to engage in nuclear trade.\\\textbf{FT Prediction:} The propos\textcolor{green}{e}\textcolor{green}{d}\textcolor{red}{a}\textcolor{red}{l} \textcolor{red}{m}\textcolor{red}{a}\textcolor{red}{y}\textcolor{green}{c}\textcolor{green}{o}\textcolor{green}{u}\textcolor{green}{l}\textcolor{green}{d} complicate the Bush administration's efforts to win \textcolor{green}{I}\textcolor{red}{a}n\textcolor{green}{d}\textcolor{green}{i}\textcolor{green}{a} exemption for India to engage in nuclear trade. [BLEU: 68.6]\\\textbf{SAFT Prediction:} The propos\textcolor{green}{e}\textcolor{green}{d}\textcolor{red}{a}\textcolor{red}{l} \textcolor{red}{m}\textcolor{red}{a}\textcolor{red}{y}\textcolor{green}{c}\textcolor{green}{o}\textcolor{green}{u}\textcolor{green}{l}\textcolor{green}{d} complicate the \textcolor{red}{B}\textcolor{red}{u}\textcolor{green}{a}\textcolor{green}{d}\textcolor{green}{m}\textcolor{green}{i}\textcolor{green}{n}\textcolor{green}{i}s\textcolor{red}{h}\textcolor{green}{t}\textcolor{green}{r}\textcolor{green}{a}\textcolor{green}{t}\textcolor{green}{i}\textcolor{green}{o}\textcolor{green}{n} administration's efforts to win \textcolor{green}{I}\textcolor{green}{n}\textcolor{green}{d}\textcolor{green}{i}an exemption f\textcolor{red}{o}r\textcolor{green}{o}\textcolor{green}{m} India \textcolor{red}{t}\textcolor{green}{f}\textcolor{green}{r}o\textcolor{green}{m} engage in nuclear trade. [BLEU: 29.7]\\[1mm]
\vspace{1em}\hrule\vspace{1em}

\textbf{Ground Truth:} In Virginia , it 's to benefit private business plans and not to serve the public interest .\\\textbf{FT Prediction:} In Virginia \textcolor{red}{,}\textcolor{green}{i}\textcolor{green}{t} it \textcolor{red}{'}\textcolor{green}{h}\textcolor{green}{a}\textcolor{green}{s}s \textcolor{green}{f}\textcolor{red}{t}o\textcolor{green}{r} benefit \textcolor{red}{p}\textcolor{red}{r}\textcolor{red}{i}\textcolor{red}{v}\textcolor{red}{a}t\textcolor{green}{h}e business plans \textcolor{green}{,}\textcolor{red}{a}\textcolor{red}{n}\textcolor{red}{d} not to serve the public \textcolor{green}{'}\textcolor{red}{i}\textcolor{red}{n}\textcolor{red}{t}\textcolor{red}{e}\textcolor{red}{r}\textcolor{red}{e}\textcolor{red}{s}\textcolor{red}{t} . [BLEU: 28.3]\\\textbf{SAFT Prediction:} I\textcolor{green}{t}\textcolor{red}{n} Virginia \textcolor{red}{,}\textcolor{green}{i}\textcolor{green}{t} it \textcolor{red}{'}\textcolor{green}{i}\textcolor{green}{s}s \textcolor{green}{f}\textcolor{red}{t}o\textcolor{green}{r} benefit \textcolor{red}{p}\textcolor{red}{r}\textcolor{red}{i}\textcolor{red}{v}\textcolor{red}{a}t\textcolor{green}{h}e business plans \textcolor{green}{,}\textcolor{red}{a}\textcolor{red}{n}\textcolor{red}{d} not to serve \textcolor{red}{t}\textcolor{red}{h}\textcolor{red}{e}\textcolor{green}{p}\textcolor{green}{u}\textcolor{green}{b}\textcolor{green}{l}\textcolor{green}{i}\textcolor{green}{c} public \textcolor{green}{'}\textcolor{red}{i}\textcolor{red}{n}\textcolor{red}{t}\textcolor{red}{e}\textcolor{red}{r}\textcolor{red}{e}\textcolor{red}{s}\textcolor{red}{t} . [BLEU: 8.4]\\[1mm]
\vspace{1em}\hrule\vspace{1em}

\textbf{Ground Truth:} 26/02/2010 14:32\\\textbf{FT Prediction:} 26/02/2010 14:32 [BLEU: 31.6]\\\textbf{SAFT Prediction:} \textcolor{red}{2}\textcolor{red}{6}\textcolor{green}{F}\textcolor{green}{e}\textcolor{green}{b}\textcolor{green}{r}\textcolor{green}{u}\textcolor{green}{a}\textcolor{green}{r}\textcolor{green}{y}/02/2010 14:32 [BLEU: 15.0]\\[1mm]
\vspace{1em}\hrule\vspace{1em}

\textbf{Ground Truth:} Nepal (NP)\\\textbf{FT Prediction:} Nepal (NP) [BLEU: 31.6]\\\textbf{SAFT Prediction:} Nepal (N\textcolor{green}{E}\textcolor{red}{P}) [BLEU: 15.0]\\[1mm]
\vspace{1em}\hrule\vspace{1em}

\subsection*{Both good}

\textbf{Ground Truth:} proliferation; technology; international; politics\\\textbf{FT Prediction:} proliferation; technology; international; politics [BLEU: 100.0]\\\textbf{SAFT Prediction:} proliferation; technology; international; politics [BLEU: 100.0]\\[1mm]
\vspace{1em}\hrule\vspace{1em}

\textbf{Ground Truth:} They deserve it. They asked for that.\\\textbf{FT Prediction:} They deserve it. They asked for that. [BLEU: 100.0]\\\textbf{SAFT Prediction:} They deserve it. They asked for that. [BLEU: 100.0]\\[1mm]
\vspace{1em}\hrule\vspace{1em}

\textbf{Ground Truth:} Tell your ex that all communication needs to go through the lawyer.\\\textbf{FT Prediction:} Tell your ex that all communication needs to go through \textcolor{green}{a}\textcolor{red}{t}\textcolor{red}{h}\textcolor{red}{e} lawyer. [BLEU: 82.7]\\\textbf{SAFT Prediction:} Tell your ex that all communication needs to go through \textcolor{green}{a}\textcolor{red}{t}\textcolor{red}{h}\textcolor{red}{e} lawyer. [BLEU: 82.7]\\[1mm]
\vspace{1em}\hrule\vspace{1em}

\textbf{Ground Truth:} Xinhua News Agency , Rome , September 1st , by reporters Aiguo Yang and Changrui Huang\\\textbf{FT Prediction:} Xinhua News Agency , Rome , September 1st , by reporter\textcolor{red}{s} Aiguo Yang and Changrui Huang [BLEU: 81.5]\\\textbf{SAFT Prediction:} Xinhua News Agency , Rome , September 1st , by reporter\textcolor{red}{s} Aiguo Yang and Changrui Huang [BLEU: 81.5]\\[1mm]
\vspace{1em}\hrule\vspace{1em}

\textbf{Ground Truth:} International; Government; technology; politics; economy\\\textbf{FT Prediction:} International; Government; technology; politics; economy [BLEU: 100.0]\\\textbf{SAFT Prediction:} International; Government; technology; politics; economy [BLEU: 100.0]\\[1mm]
\vspace{1em}\hrule\vspace{1em}

\subsection*{Both bad}

\textbf{Ground Truth:} Haha\\\textbf{FT Prediction:} Haha\textcolor{green}{.} [BLEU: 0.0]\\\textbf{SAFT Prediction:} Haha\textcolor{green}{.} [BLEU: 0.0]\\[1mm]
\vspace{1em}\hrule\vspace{1em}

\textbf{Ground Truth:} Good Evening Digicel.\\\textbf{FT Prediction:} Good \textcolor{green}{e}\textcolor{red}{E}vening\textcolor{green}{,}\textcolor{red}{ }\textcolor{red}{D}\textcolor{red}{i}\textcolor{red}{g}icel\textcolor{red}{.} [BLEU: 9.1]\\\textbf{SAFT Prediction:} Good \textcolor{green}{e}\textcolor{red}{E}vening\textcolor{green}{,}\textcolor{red}{ }\textcolor{red}{D}\textcolor{red}{i}\textcolor{red}{g}icel\textcolor{green}{,}\textcolor{red}{.} [BLEU: 9.1]\\[1mm]
\vspace{1em}\hrule\vspace{1em}

\textbf{Ground Truth:} To help the survivors of the Gulf.\\\textbf{FT Prediction:} \textcolor{red}{T}\textcolor{red}{o}\textcolor{green}{H}\textcolor{green}{e}\textcolor{green}{l}\textcolor{green}{p} help th\textcolor{green}{o}\textcolor{green}{s}e survivors \textcolor{green}{i}\textcolor{green}{n}\textcolor{red}{o}\textcolor{red}{f} the Gulf\textcolor{red}{.}\textcolor{green}{ }\textcolor{green}{r}\textcolor{green}{e}\textcolor{green}{g}\textcolor{green}{i}\textcolor{green}{o}\textcolor{green}{n} [BLEU: 3.7]\\\textbf{SAFT Prediction:} \textcolor{red}{T}\textcolor{red}{o}\textcolor{green}{S}\textcolor{green}{u}\textcolor{green}{r}\textcolor{green}{v} \textcolor{red}{h}\textcolor{green}{s}\textcolor{green}{u}\textcolor{green}{r}\textcolor{green}{v}\textcolor{green}{i}\textcolor{green}{v}e\textcolor{red}{l}\textcolor{red}{p} th\textcolor{green}{o}\textcolor{green}{s}e survivors \textcolor{green}{i}\textcolor{green}{n}\textcolor{red}{o}\textcolor{red}{f} the Gulf. [BLEU: 7.7]\\[1mm]
\vspace{1em}\hrule\vspace{1em}

\textbf{Ground Truth:} Please tell us our to pray because almost not believe in god and our prayer never arrived to god.\\\textbf{FT Prediction:} Please tell us \textcolor{green}{h}o\textcolor{green}{w}\textcolor{red}{u}\textcolor{red}{r} \textcolor{red}{t}\textcolor{red}{o}\textcolor{green}{p}\textcolor{green}{r}\textcolor{green}{a}\textcolor{green}{y}\textcolor{green}{e}\textcolor{green}{r}\textcolor{green}{s} pray \textcolor{green}{i}\textcolor{green}{n}\textcolor{green}{ }\textcolor{green}{w}\textcolor{red}{b}\textcolor{red}{e}\textcolor{red}{c}\textcolor{red}{a}\textcolor{red}{u}\textcolor{red}{s}e \textcolor{green}{w}\textcolor{green}{i}\textcolor{red}{a}\textcolor{red}{l}\textcolor{red}{m}\textcolor{red}{o}\textcolor{red}{s}t\textcolor{green}{h}\textcolor{red}{ }\textcolor{red}{n}o\textcolor{green}{u}t believ\textcolor{red}{e}\textcolor{green}{i}\textcolor{green}{n}\textcolor{green}{g} in \textcolor{green}{G}\textcolor{red}{g}od and \textcolor{green}{h}\textcolor{green}{e}\textcolor{red}{o}\textcolor{red}{u}\textcolor{red}{r} prayer\textcolor{green}{s} \textcolor{green}{w}\textcolor{green}{i}\textcolor{green}{l}\textcolor{green}{l}\textcolor{red}{n}\textcolor{red}{e}\textcolor{red}{v}\textcolor{red}{e}\textcolor{red}{r} arrived \textcolor{green}{a}t\textcolor{red}{o} \textcolor{green}{h}\textcolor{green}{i}\textcolor{green}{m}\textcolor{red}{g}\textcolor{red}{o}\textcolor{red}{d}. [BLEU: 6.2]\\\textbf{SAFT Prediction:} Please\textcolor{green}{,}\textcolor{red}{ }\textcolor{red}{t}\textcolor{red}{e}\textcolor{red}{l}\textcolor{red}{l} us \textcolor{green}{h}o\textcolor{green}{w}\textcolor{red}{u}\textcolor{red}{r}\textcolor{red}{ }\textcolor{red}{t}\textcolor{red}{o} pray\textcolor{red}{ }\textcolor{red}{b}e\textcolor{green}{r}\textcolor{red}{c}\textcolor{red}{a}\textcolor{red}{u}s\textcolor{green}{ }\textcolor{green}{p}\textcolor{green}{r}\textcolor{green}{a}\textcolor{green}{y}\textcolor{green}{,}\textcolor{green}{ }\textcolor{green}{w}e \textcolor{green}{w}\textcolor{green}{e}\textcolor{red}{a}\textcolor{red}{l}\textcolor{red}{m}\textcolor{red}{o}\textcolor{red}{s}\textcolor{red}{t}\textcolor{red}{ }\textcolor{red}{n}\textcolor{red}{o}\textcolor{red}{t} believ\textcolor{red}{e}\textcolor{green}{i}\textcolor{green}{n}\textcolor{green}{g} in \textcolor{green}{G}\textcolor{red}{g}od and our prayer\textcolor{green}{s} \textcolor{green}{w}\textcolor{green}{i}\textcolor{green}{l}\textcolor{green}{l}\textcolor{red}{n}\textcolor{red}{e}\textcolor{red}{v}\textcolor{red}{e}\textcolor{red}{r} arrive\textcolor{red}{d} to god. [BLEU: 3.4]\\[1mm]
\vspace{1em}\hrule\vspace{1em}

\textbf{Ground Truth:} You may think that 's not rational land use .\\\textbf{FT Prediction:} You \textcolor{green}{c}\textcolor{red}{m}a\textcolor{green}{n}\textcolor{red}{y} think that \textcolor{green}{i}\textcolor{red}{'}s\textcolor{red}{ }\textcolor{red}{n}\textcolor{red}{o}t\textcolor{green}{h}\textcolor{red}{ }\textcolor{red}{r}at\textcolor{red}{i}\textcolor{red}{o}\textcolor{red}{n}\textcolor{red}{a}\textcolor{red}{l} \textcolor{red}{l}\textcolor{green}{t}\textcolor{green}{h}\textcolor{green}{e}\textcolor{green}{ }a\textcolor{red}{n}\textcolor{red}{d} use \textcolor{green}{u}\textcolor{green}{s}\textcolor{green}{e}. [BLEU: 5.4]\\\textbf{SAFT Prediction:} You \textcolor{green}{c}\textcolor{red}{m}a\textcolor{green}{n}\textcolor{red}{y} think that \textcolor{red}{'}\textcolor{red}{s}\textcolor{red}{ }\textcolor{red}{n}\textcolor{red}{o}\textcolor{red}{t}\textcolor{red}{ }\textcolor{red}{r}\textcolor{red}{a}\textcolor{red}{t}i\textcolor{red}{o}\textcolor{green}{s}\textcolor{green}{u}\textcolor{green}{s}\textcolor{green}{i}n\textcolor{green}{g}\textcolor{red}{a}\textcolor{red}{l} land us\textcolor{green}{i}\textcolor{green}{n}\textcolor{green}{g}\textcolor{green}{ }\textcolor{green}{u}\textcolor{green}{s}e \textcolor{green}{u}\textcolor{green}{s}\textcolor{green}{e}. [BLEU: 5.7]\\[1mm]
\vspace{1em}\hrule\vspace{1em}

%% file: figures/preprocessing_examples/amr_example1.tex
\begin{figure}[ht]
    \centering
    \includegraphics[width=0.80\linewidth]{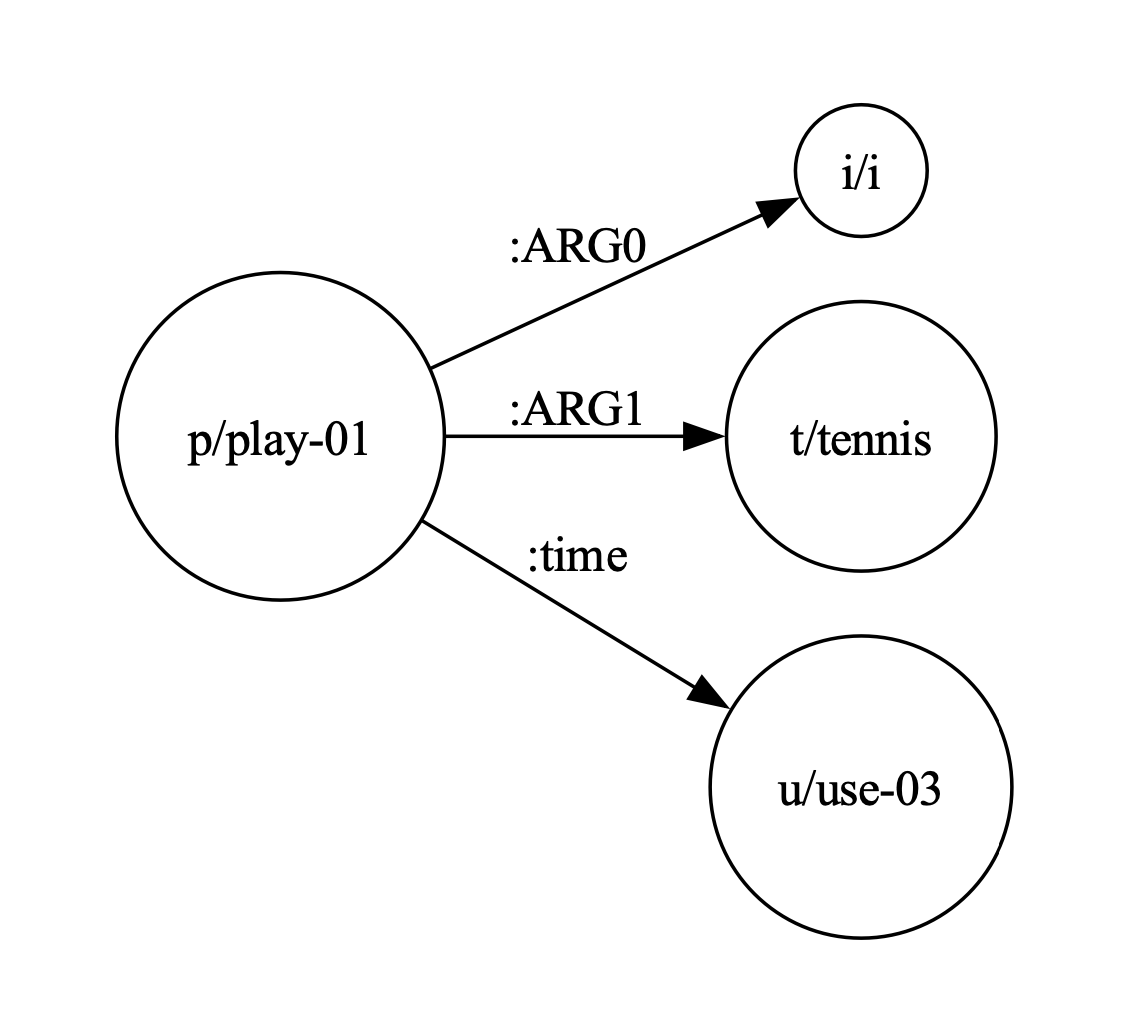}
    \caption{Original AMR graph for the sentence ``I used to play tennis''.}
    \label{fig:example1_amr}
\end{figure}

%% file: figures/preprocessing_examples/preprocessing_example1.tex
\begin{figure}[!ht]
\centering

\begin{subfigure}{0.96\linewidth}
    \centering
    \input{assets/listings/preprocessing_examples/bfs_example1}
    \caption{BFS Linearization}
\end{subfigure}

\begin{subfigure}{0.96\linewidth}
    \centering
    \includegraphics[width=\linewidth]{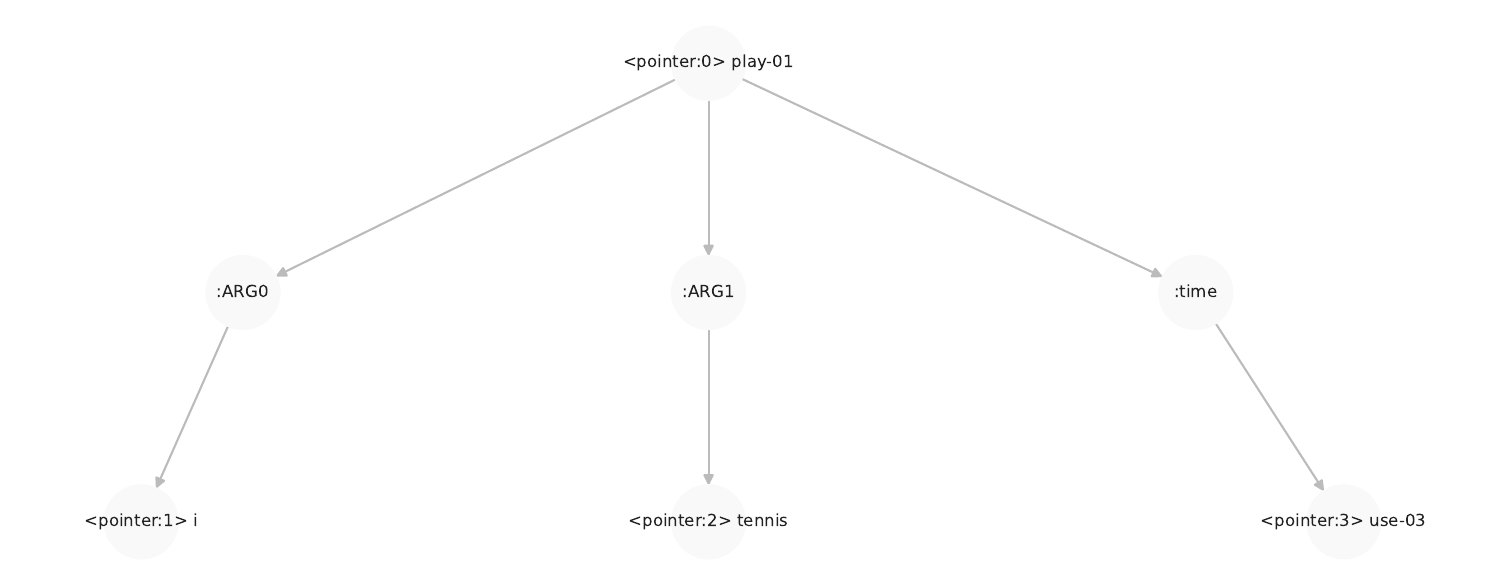}
    \caption{\textsc{ToSubgraph}}
\end{subfigure}

\begin{subfigure}{0.96\linewidth}
    \centering
    \includegraphics[width=\linewidth]{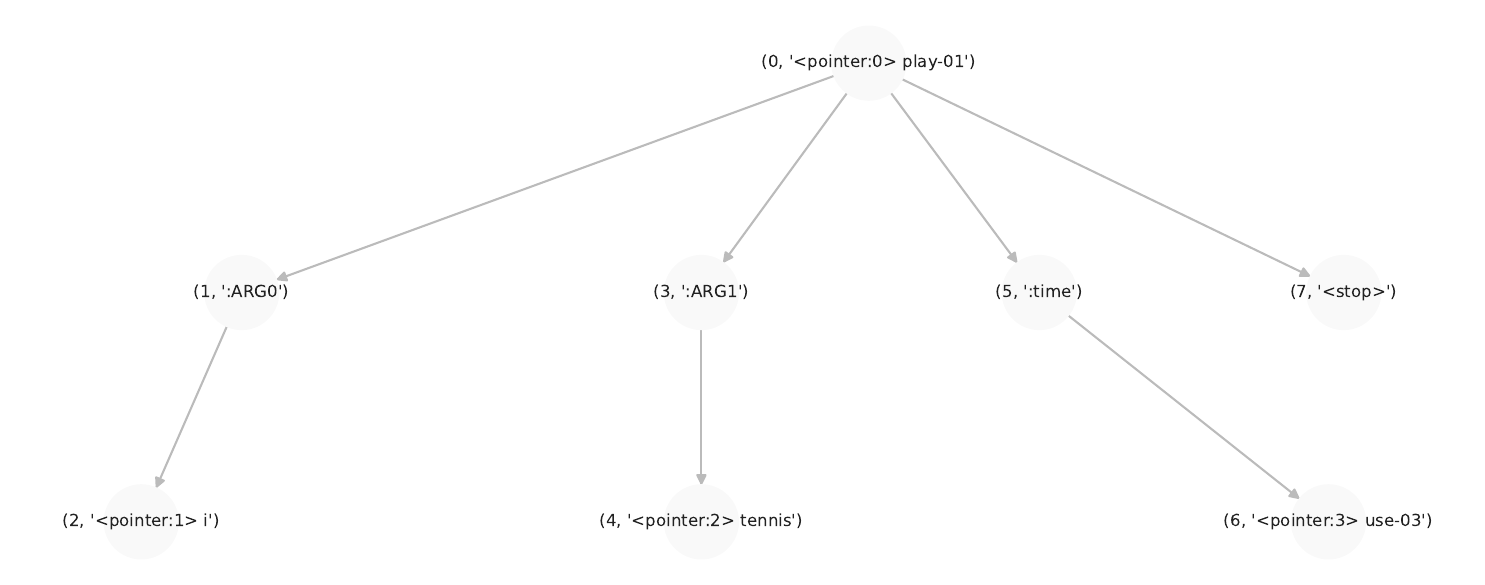}
    \caption{\textsc{RoleExpand}, \textsc{AddStopNodes}, and \(\sigma_\mathcal{A}^{-1}\)}
\end{subfigure}

\begin{subfigure}{0.96\linewidth}
    \centering
    \includegraphics[width=\linewidth]{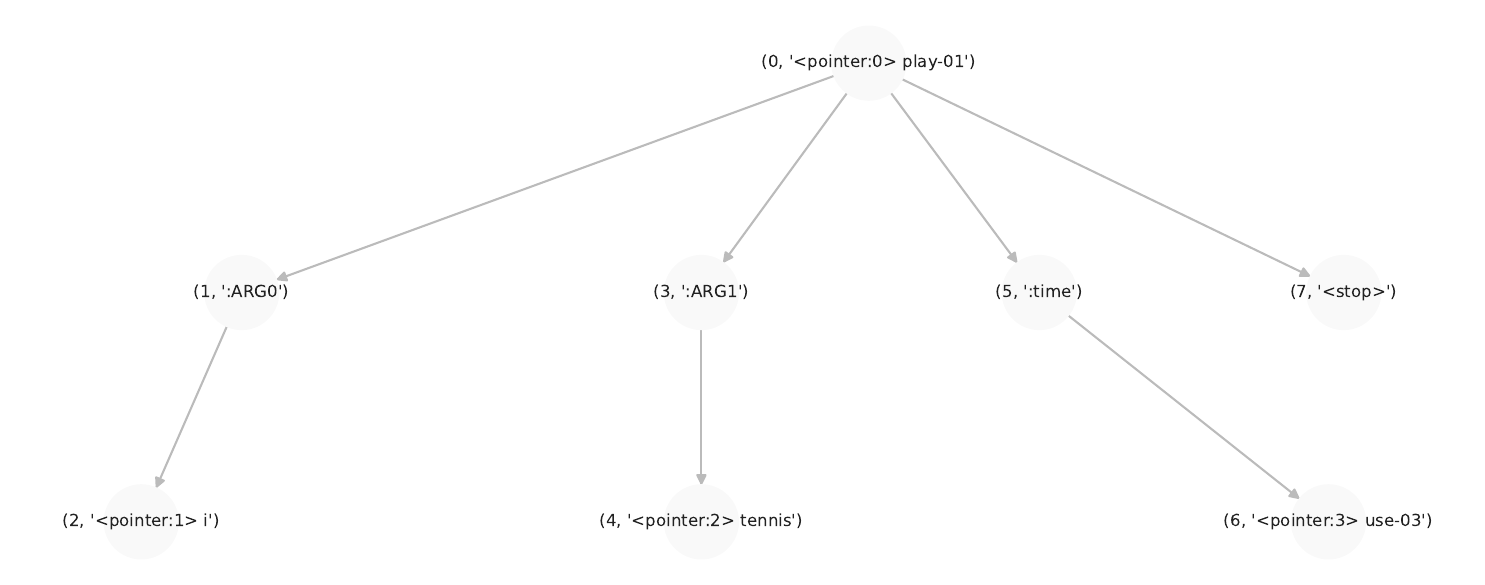}
    \caption{\textsc{Merge}}
\end{subfigure}

\vspace{1.5em}

\caption{Overview of preprocessing steps for the AMR corresponding to the sentence: ``I used to play tennis''.}
\label{fig:preprocessing_example_1}

\end{figure}

%% file: assets/listings/preprocessing_examples/bfs_example1.tex
\begin{tcolorbox}[
  blanker,
  fontupper=\ttfamily,
  before upper={\footnotesize},
  breakable,
  enhanced jigsaw,
  width=\linewidth,
  top=0pt,
  bottom=0pt,
  left=1pt,
  right=1pt,
  boxsep=1pt,
  colback=white,
  sharp corners,
  nobeforeafter,
]
\pointer{0} \concept{play-01} \role{:ARG0} \pointer{1} \concept{i} \role{:ARG1} \pointer{2} \concept{tennis} \role{:time} \pointer{3} \concept{use-03} \stopwordamr{<stop>}
\end{tcolorbox}

%% file: figures/preprocessing_examples/amr_example2.tex
\begin{figure}[ht]
    \centering
    \includegraphics[width=0.70\linewidth]{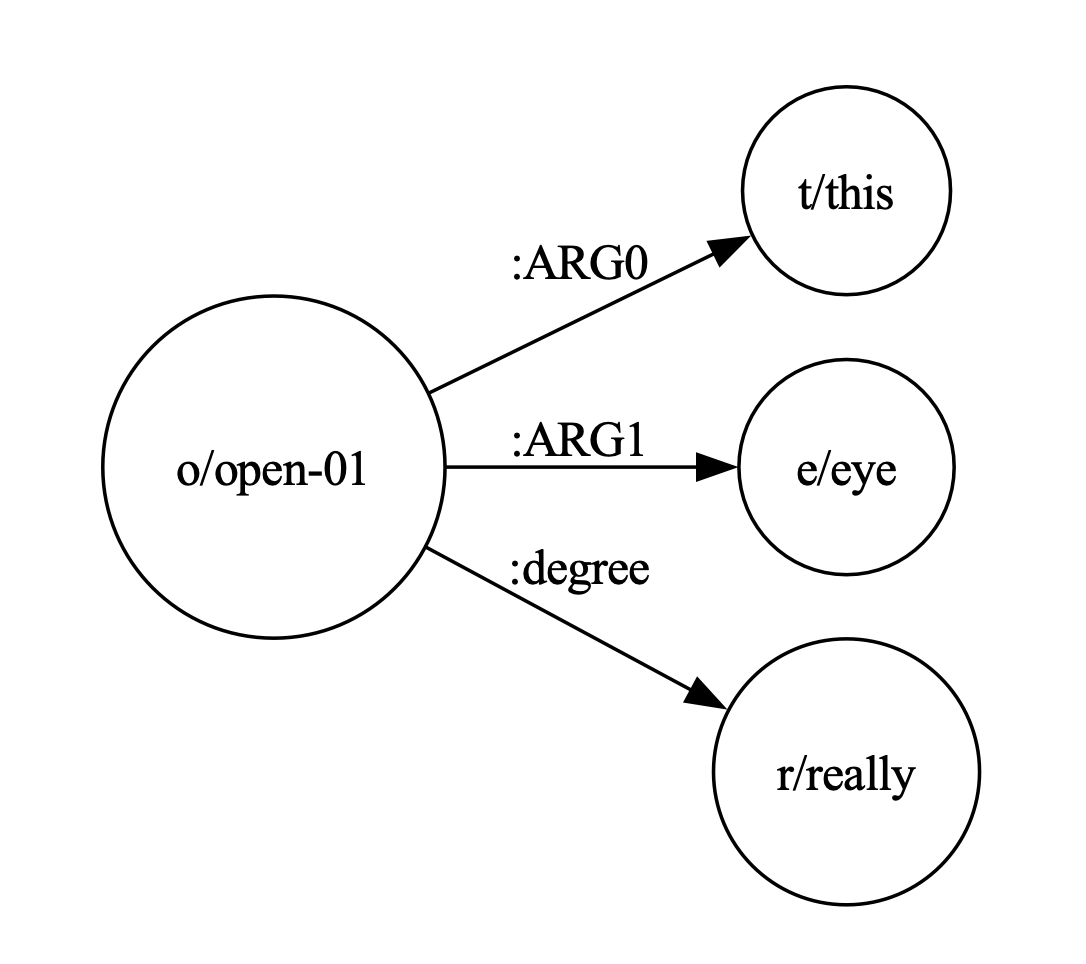}
    \caption{Original AMR graph for the sentence ``This is really eye-opening''.}
    \label{fig:example2_amr}
\end{figure}

%% file: figures/preprocessing_examples/preprocessing_example2.tex
\begin{figure}[ht]
\centering

\begin{subfigure}{0.96\linewidth}
    \centering
    \input{assets/listings/preprocessing_examples/bfs_example2}
    \caption{BFS Linearization}
\end{subfigure}

\begin{subfigure}{0.96\linewidth}
    \centering
    \includegraphics[width=\linewidth]{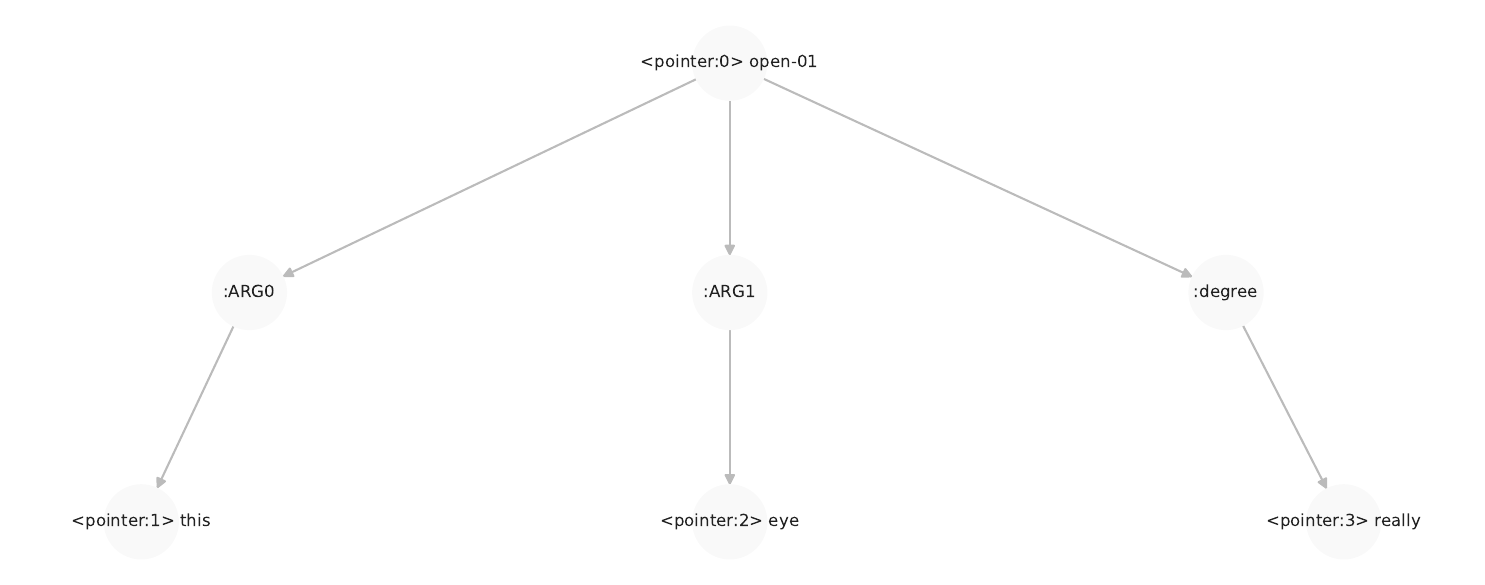}
    \caption{\textsc{ToSubgraph}}
\end{subfigure}

\begin{subfigure}{0.96\linewidth}
    \centering
    \includegraphics[width=\linewidth]{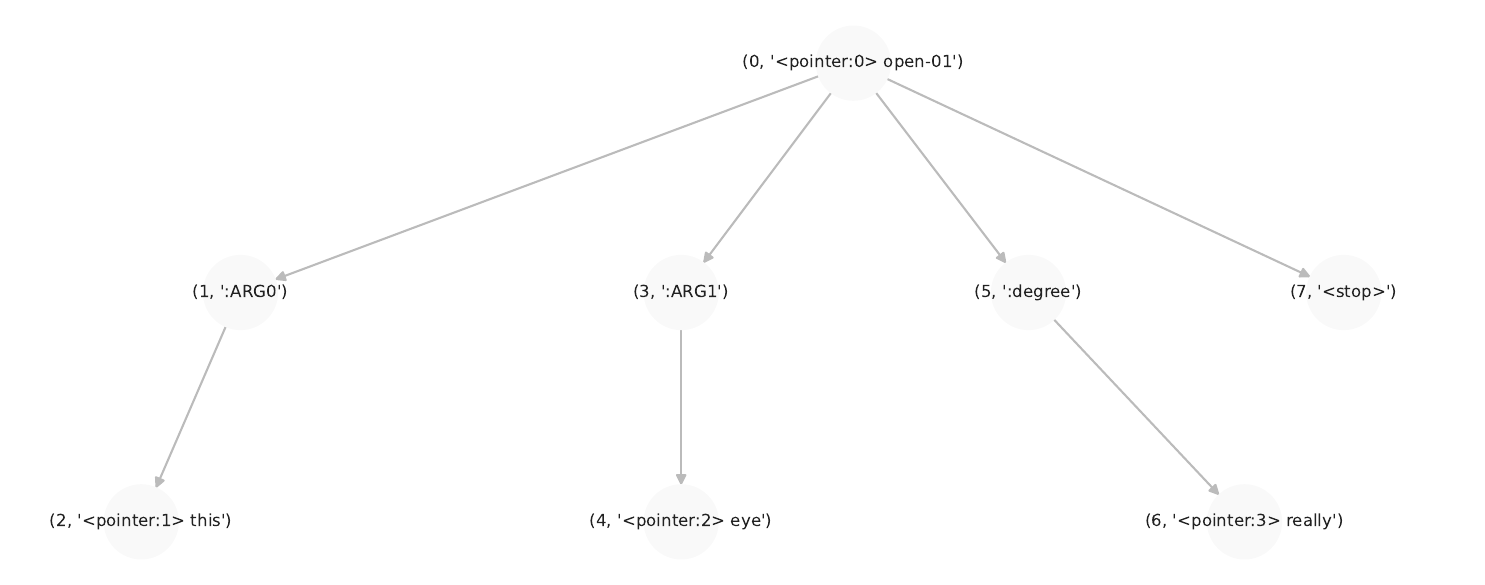}
    \caption{\textsc{RoleExpand}, \textsc{AddStopNodes}, and \(\sigma_\mathcal{A}^{-1}\)}
\end{subfigure}

\begin{subfigure}{0.96\linewidth}
    \centering
    \includegraphics[width=\linewidth]{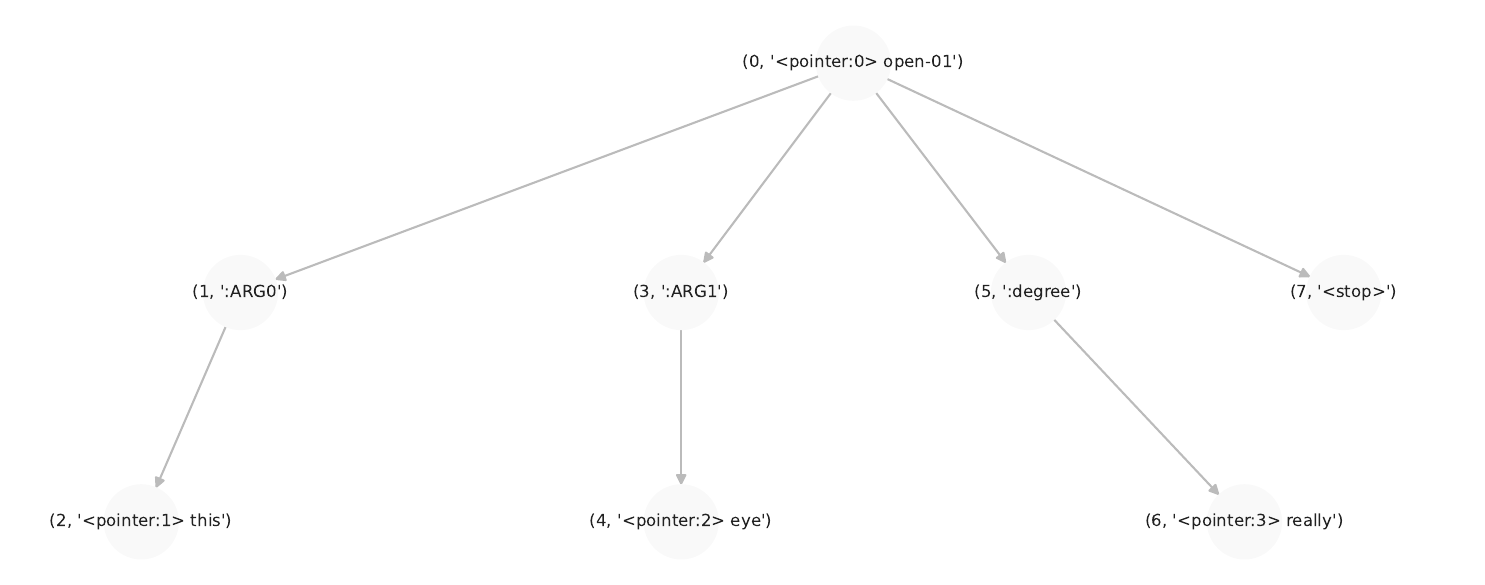}
    \caption{\textsc{Merge}}
\end{subfigure}

\caption{Overview of preprocessing steps for the AMR corresponding to the sentence: ``This is really eye-opening''.}
\label{fig:preprocessing_example_2}

\end{figure}

%% file: assets/listings/preprocessing_examples/bfs_example2.tex
\begin{tcolorbox}[
  blanker,
  fontupper=\ttfamily,
  before upper={\footnotesize},
  breakable,
  enhanced jigsaw,
  width=\linewidth,
  top=0pt,
  bottom=0pt,
  left=1pt,
  right=1pt,
  boxsep=1pt,
  colback=white,
  sharp corners,
  nobeforeafter,
]
\pointer{0} \concept{open-01} \role{:ARG0} \pointer{1} \concept{this} \role{:ARG1} \pointer{2} \concept{eye} \role{:degree} \pointer{3} \concept{really} \stopwordamr{<stop>}
\end{tcolorbox}

%% file: figures/preprocessing_examples/amr_example3.tex
\begin{figure}[ht]
    \centering
    \includegraphics[width=0.80\linewidth]{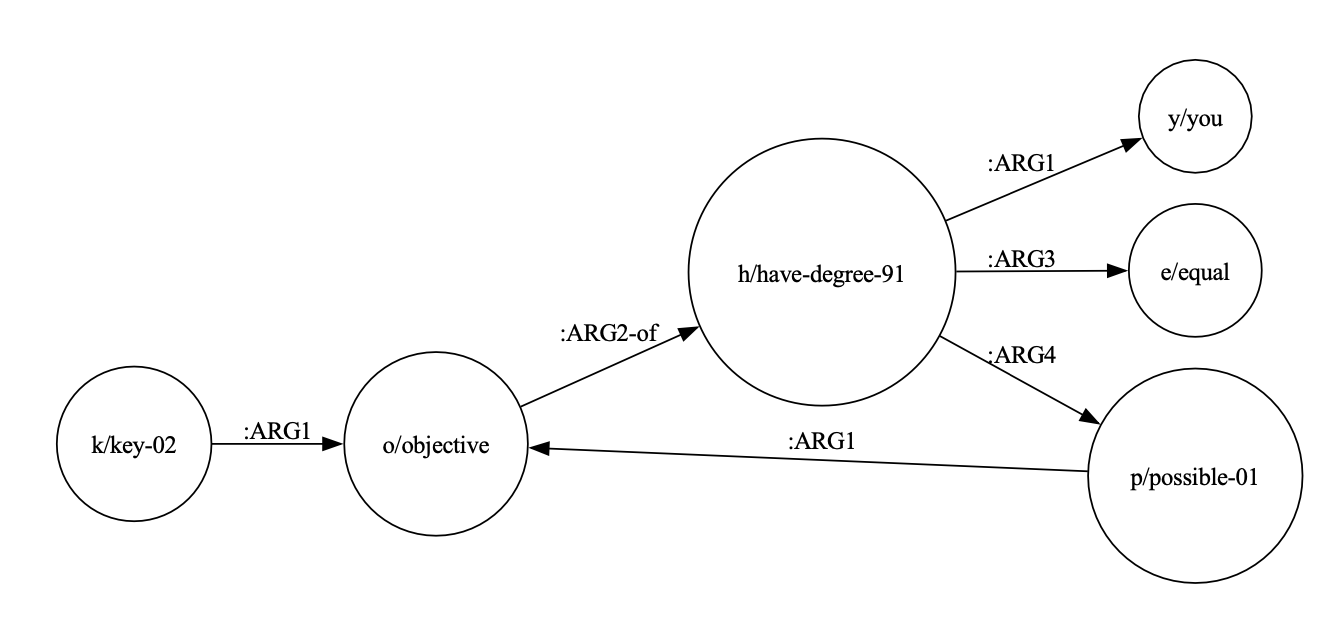}
    \caption{Original AMR graph for the sentence ``The key is to be as objective as possible''.}
    \label{fig:example3_amr}
\end{figure}

%% file: figures/preprocessing_examples/preprocessing_example3.tex
\begin{figure}[!ht]
\centering
\begin{subfigure}{0.96\linewidth}
    \centering
    \input{assets/listings/preprocessing_examples/bfs_example3}
    \caption{BFS Linearization}
\end{subfigure}
\begin{subfigure}{0.96\linewidth}
    \centering
    \includegraphics[width=\linewidth]{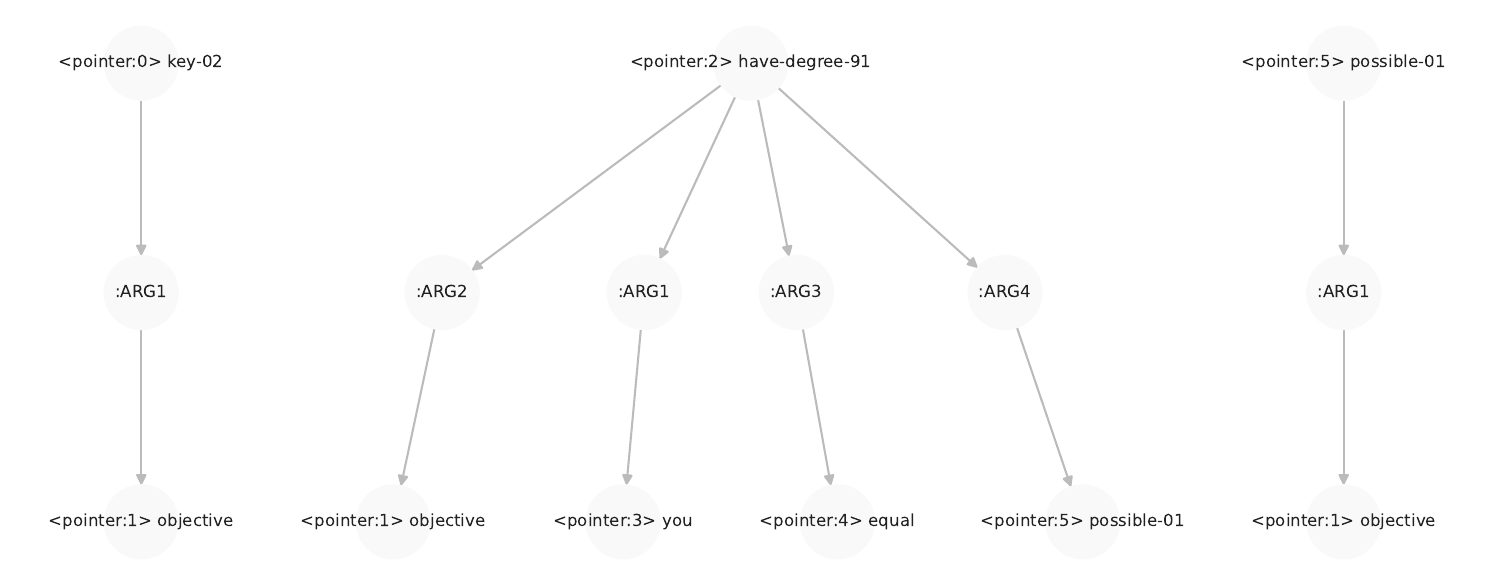}
    \caption{\textsc{ToSubgraph}}
\end{subfigure}
\begin{subfigure}{0.96\linewidth}
    \centering
    \includegraphics[width=\linewidth]{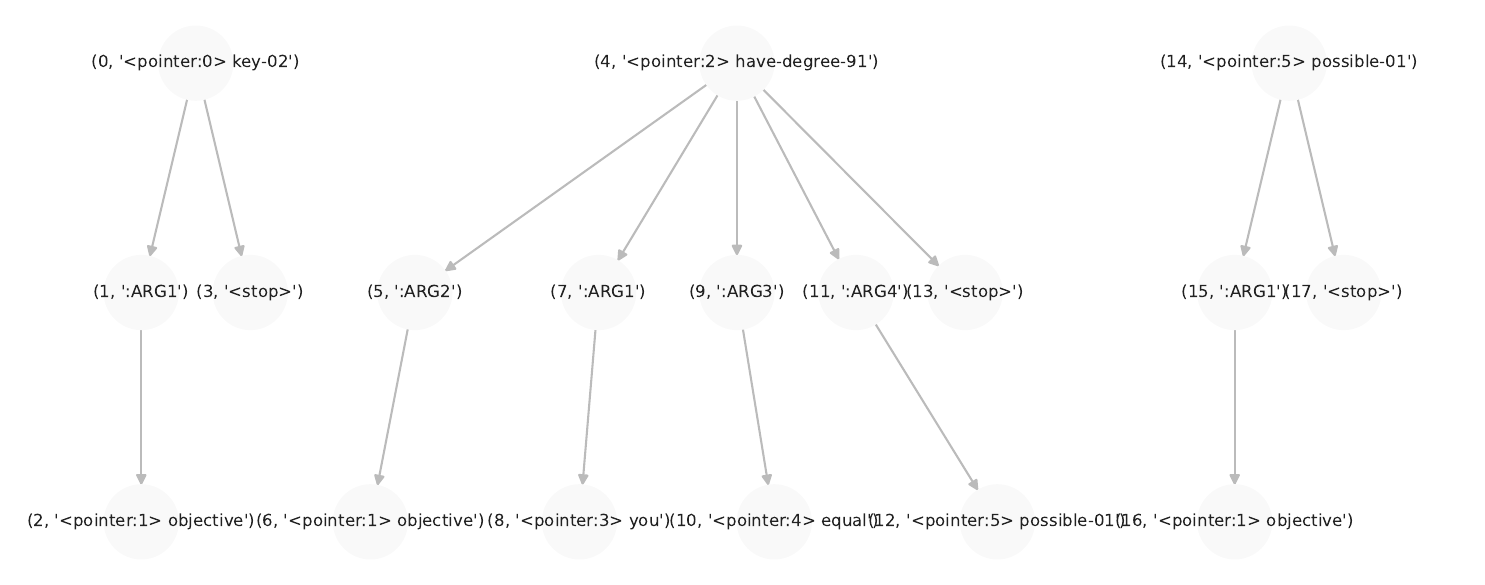}
    \caption{\textsc{RoleExpand}, \textsc{AddStopNodes}, and \(\sigma_\mathcal{A}^{-1}\)}
\end{subfigure}
\begin{subfigure}{0.96\linewidth}
    \centering
    \includegraphics[width=\linewidth]{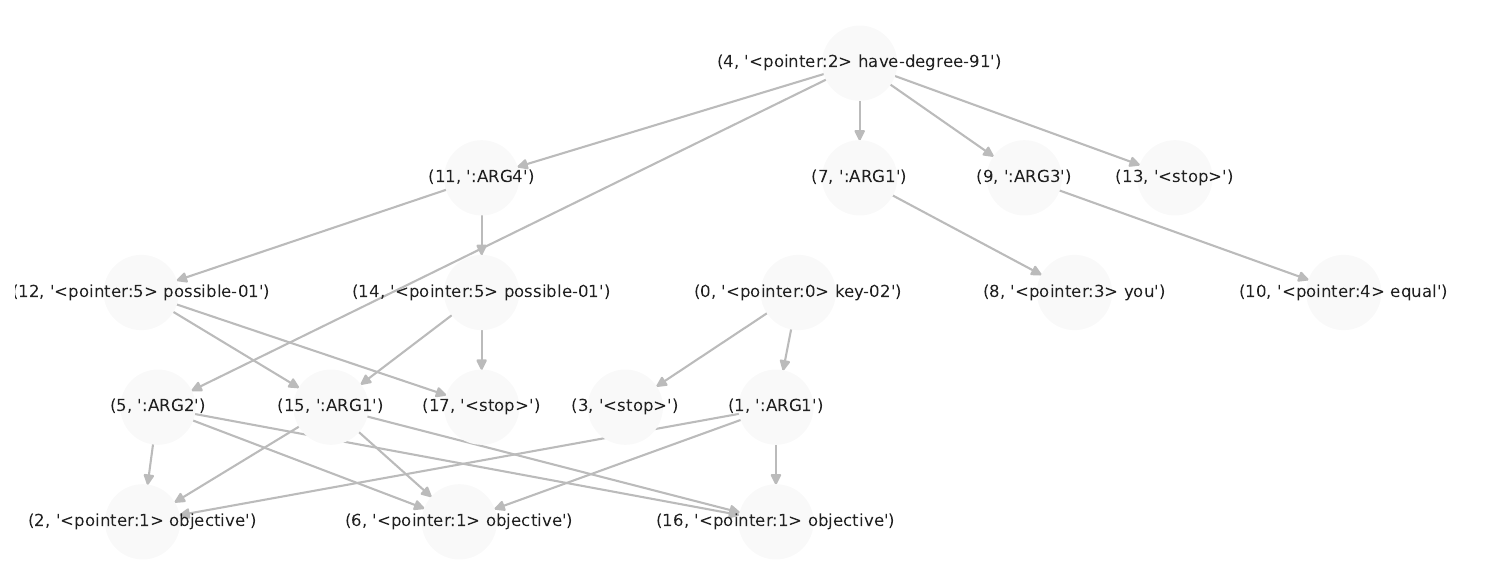}
    \caption{\textsc{Merge}}
\end{subfigure}
\caption{Overview of preprocessing steps for the AMR corresponding to the sentence: ``The key is to be as objective as possible''.}
\label{fig:preprocessing_example_3}
\end{figure}

%% file: assets/listings/preprocessing_examples/bfs_example3.tex
\begin{tcolorbox}[
  blanker,
  fontupper=\ttfamily,
  before upper={\footnotesize},
  breakable,
  enhanced jigsaw,
  width=\linewidth,
  top=0pt,
  bottom=0pt,
  left=1pt,
  right=1pt,
  boxsep=1pt,
  colback=white,
  sharp corners,
  nobeforeafter,
]
\pointer{0} \concept{key-02} \role{:ARG1} \pointer{1} \concept{objective} \stopwordamr{<stop>} 
\pointer{2} \concept{have-degree-91} \role{:ARG2} \pointer{1} \role{:ARG1} \pointer{3} \concept{you} \role{:ARG3} \pointer{4} \concept{equal} \role{:ARG4} \pointer{5} \concept{possible-01} \stopwordamr{<stop>} 
\pointer{5} \role{:ARG1} \pointer{1} \stopwordamr{<stop>}
\end{tcolorbox}

%% file: figures/preprocessing_examples/amr_example4.tex
\begin{figure}[ht]
    \centering
    \includegraphics[width=0.80\linewidth]{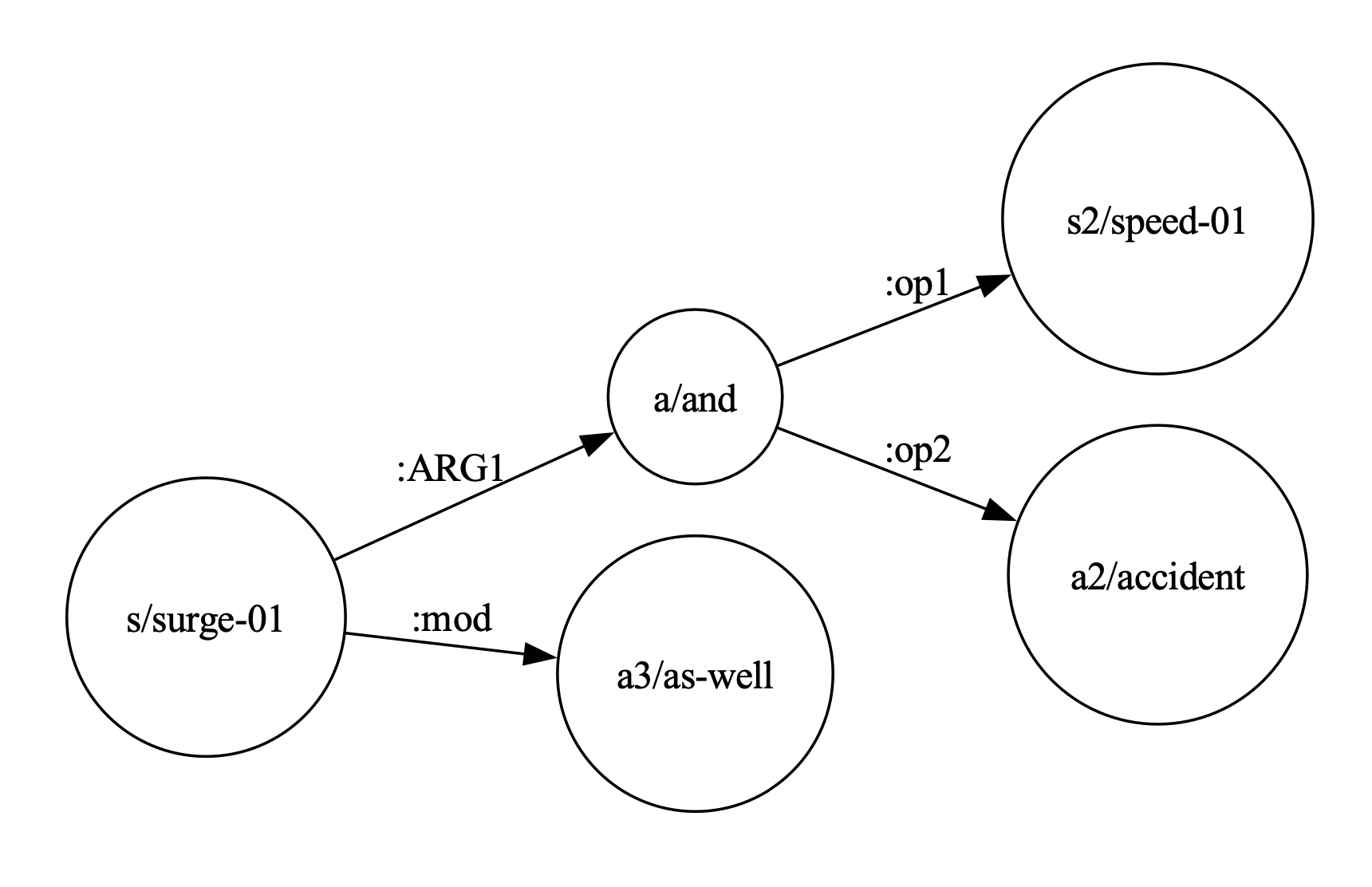}
    \caption{Original AMR graph for the sentence ``Speeding and accidents have surged as well''.}
    \label{fig:example4_amr}
\end{figure}

%% file: figures/preprocessing_examples/preprocessing_example4.tex
\begin{figure}[!ht]
    \centering
    \begin{subfigure}{0.96\linewidth}
        \centering
        \input{assets/listings/preprocessing_examples/bfs_example4}
        \caption{BFS Linearization}
    \end{subfigure}
    \begin{subfigure}{0.96\linewidth}
        \centering
        \includegraphics[width=\linewidth]{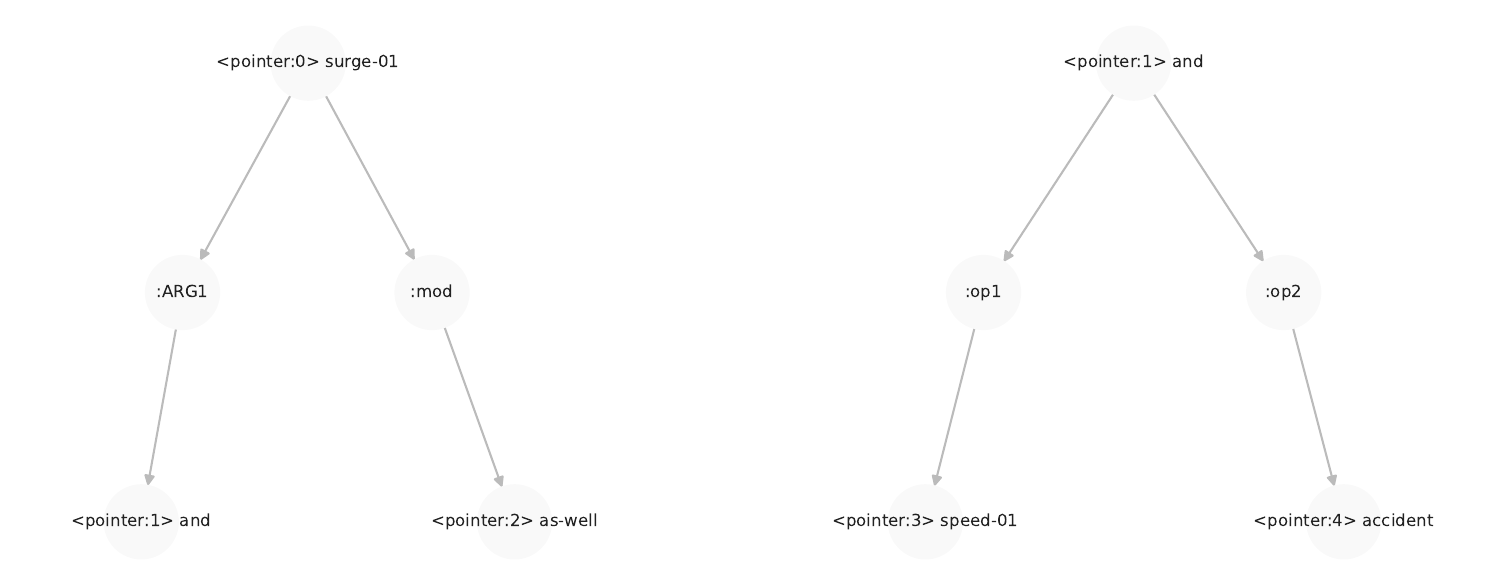}
        \caption{\textsc{ToSubgraph}}
    \end{subfigure}
    \begin{subfigure}{0.96\linewidth}
        \centering
        \includegraphics[width=\linewidth]{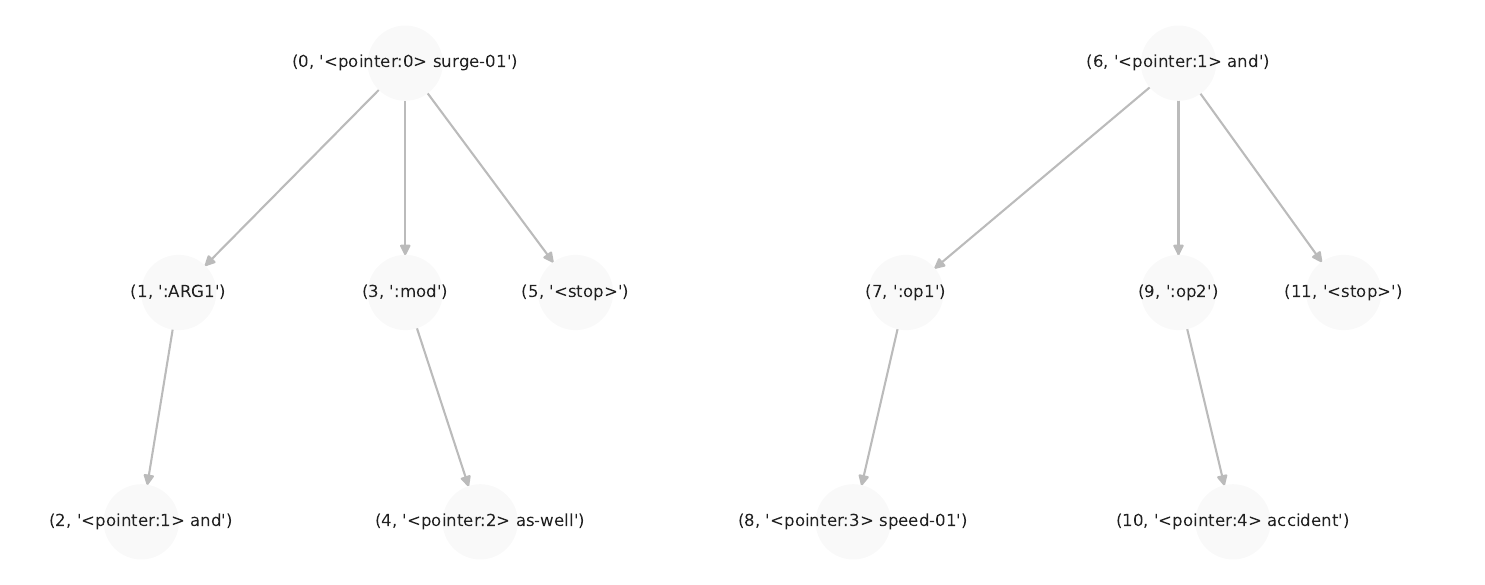}
        \caption{\textsc{RoleExpand}, \textsc{AddStopNodes}, and \(\sigma_\mathcal{A}^{-1}\)}
    \end{subfigure}    
    \begin{subfigure}{0.96\linewidth}
        \centering
        \includegraphics[width=\linewidth]{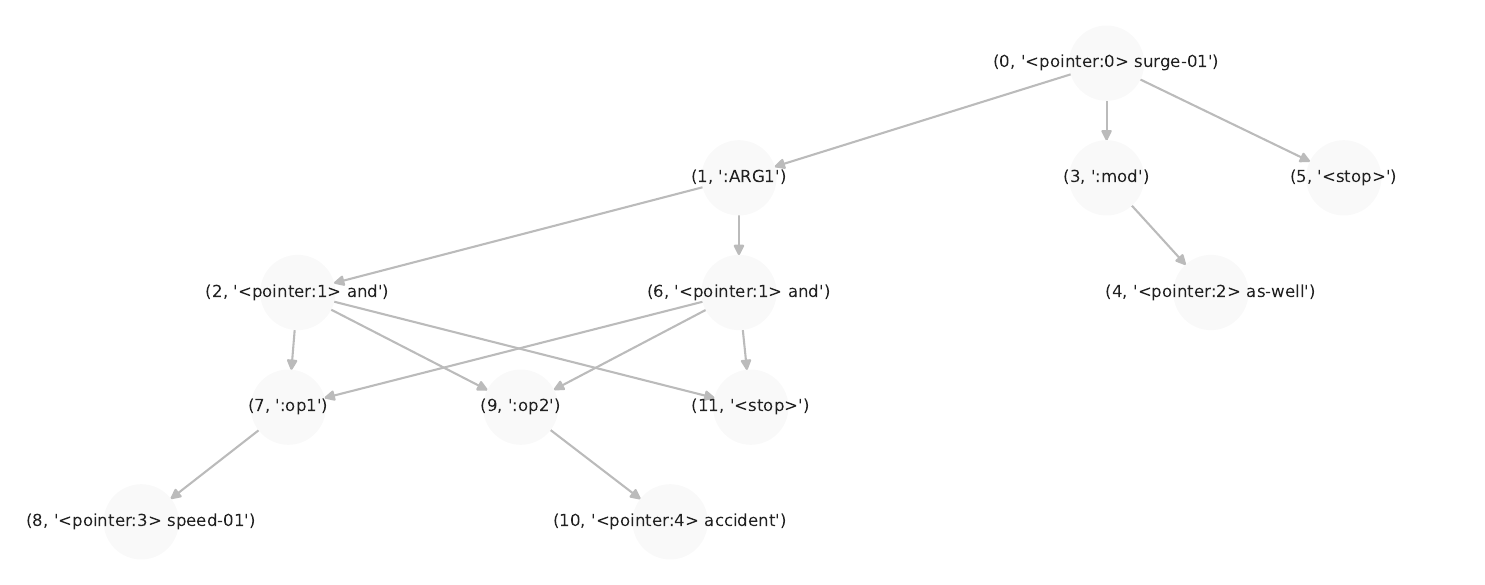}
        \caption{\textsc{Merge}}
    \end{subfigure}

    \caption{Overview of preprocessing steps for the AMR corresponding to the sentence: ``Speeding and accidents have surged as well''.}
    \label{fig:preprocessing_example_4}
    
\end{figure}

%% file: assets/listings/preprocessing_examples/bfs_example4.tex
\begin{tcolorbox}[
  blanker,
  fontupper=\ttfamily,
  before upper={\footnotesize},
  breakable,
  enhanced jigsaw,
  width=\linewidth,
  top=0pt,
  bottom=0pt,
  left=1pt,
  right=1pt,
  boxsep=1pt,
  colback=white,
  sharp corners,
  nobeforeafter,
]
\pointer{0} \concept{surge-01} \role{:ARG1} \pointer{1} \concept{and} \role{:mod} \pointer{2} \concept{as-well} \stopwordamr{<stop>} 
\pointer{1} \role{:op1} \pointer{3} \concept{speed-01} \role{:op2} \pointer{4} \concept{accident} \stopwordamr{<stop>}
\end{tcolorbox}